%% file: main.tex
\documentclass[10pt,twocolumn,letterpaper]{article}

\usepackage[pagenumbers]{cvpr} 
\usepackage{makecell}

\input{preamble}

\newcommand*\samethanks[1][\value{footnote}]{\footnotemark[#1]}

\definecolor{cvprblue}{rgb}{0.21,0.49,0.74}
\usepackage[pagebackref,breaklinks,colorlinks,allcolors=cvprblue]{hyperref}

\title{Generating an Image From 1,000 Words: Enhancing Text-to-Image With Structured Captions}

\author{
Eyal Gutflaish \footnote{}~ \ \ \
Eliran Kachlon \samethanks~ \ \ \
Hezi Zisman \ \ \
Tal Hacham \ \ \
Nimrod Sarid \ \ \
Alexander Visheratin \\ 
Saar Huberman \ \ \ 
Gal Davidi \ \ \ 
Guy Bukchin \ \ \ 
Kfir Goldberg \ \ \
Ron Mokady \footnote{}~~ \ \ \ 
\\[2mm]
\vspace{1em}
BRIA AI 
\\[0.5mm]
\vspace{-1mm}
\centering
}

\begin{document}

\twocolumn[{
\renewcommand\twocolumn[1][]{#1}

\vspace{-1.5cm}
\maketitle
\begin{center}
  \centering
  \vspace{-1.5cm}
  \includegraphics[width=\textwidth]{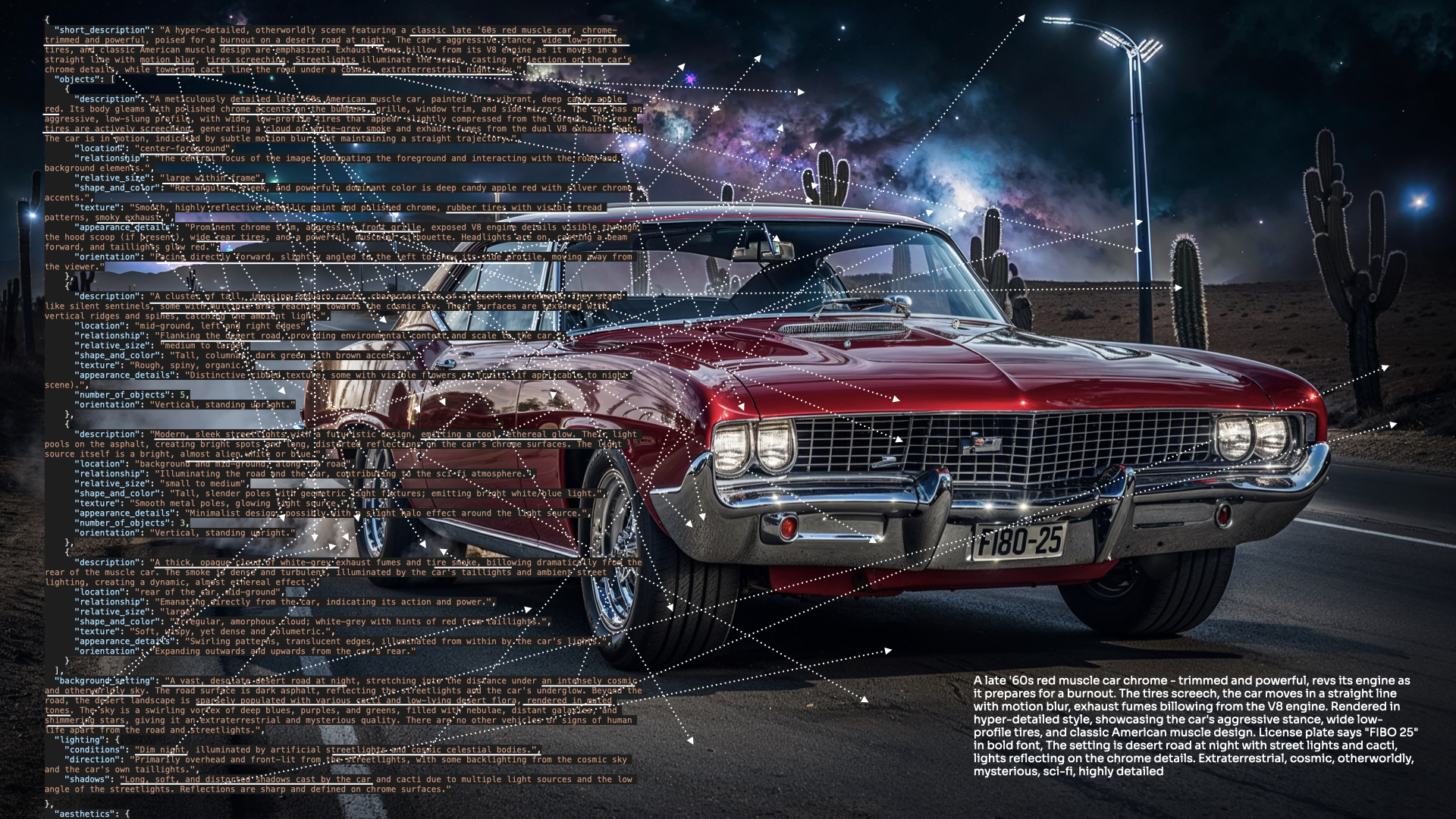}
\captionof{figure}{\textbf{Generating an image from a long structured caption.} Training with long structured captions substantially improves controllability and expressiveness. Unlike short prompts, each scene element is specified with rich attributes (e.g., position, size, texture, and relations), while global properties (e.g., background, lighting, composition, and photographic characteristics) are explicitly defined. The result is a complete, machine-readable description that enables faithful, controllable generation for professional use.}
  \label{fig:teaser}
\end{center}
}]

\input{sec/0_abstract}

\vspace{-1cm}

{
\renewcommand{\thefootnote}{\fnsymbol{footnote}}
\footnotetext[1]{Equal contribution.}
\footnotetext[2]{Project lead.}
}

\input{sec/1_intro}

\input{sec/2_related}

\input{sec/3_method}

\input{sec/4_experiments}
\input{sec/5_conclusion}
{
    \small
    \bibliographystyle{ieeenat_fullname}
    \bibliography{main}
}

\input{sec/X_suppl}

\end{document}

%% file: preamble.tex
\usepackage{xcolor}
\definecolor{darkgreen}{RGB}{0,100,0}
\definecolor{llmtokens}{HTML}{BAEFB5}
\definecolor{textencoding}{HTML}{8B88F8}
\definecolor{imagetokens}{HTML}{8AB9F1}
\usepackage{tikz}
\usepackage{circuitikz}
\usepackage{adjustbox}
\usepackage{multirow}
\usepackage{booktabs,tabularx}
\usepackage{tcolorbox}
\usepackage{float}
\usepackage{graphicx}
\usepackage{subcaption}
\tcbuselibrary{listings,breakable}
\usetikzlibrary{arrows.meta,calc,positioning,fit,shapes,backgrounds}
\tikzset{every node/.style={font=\footnotesize}}
\usepackage[page]{appendix}
\usepackage{listings}

\newcommand{\modelname}{FIBO}



%% file: sec/0_abstract.tex
\begin{abstract}

Text-to-image models have rapidly evolved from casual creative tools to professional-grade systems, achieving unprecedented levels of image quality and realism. Yet, most models are trained to map short prompts into detailed images, creating a gap between sparse textual input and rich visual outputs. This mismatch reduces controllability, as models often fill in missing details arbitrarily, biasing toward average user preferences and limiting precision for professional use. We address this limitation by training the first open-source text-to-image model on \emph{long structured captions}, where every training sample is annotated with the same set of fine-grained attributes. This design maximizes expressive coverage and enables disentangled control over visual factors. To process long captions efficiently, we propose \emph{DimFusion}, a fusion mechanism that integrates intermediate tokens from a lightweight LLM without increasing token length. We also introduce the \emph{Text-as-a-Bottleneck Reconstruction (TaBR)} evaluation protocol. By assessing how well real images can be reconstructed through a captioning–generation loop, TaBR directly measures controllability and expressiveness—even for very long captions where existing evaluation methods fail. 
Finally, we demonstrate our contributions by training the large-scale model \emph{\modelname{}}, achieving state-of-the-art prompt alignment among open-source models. Model weights are publicly available at \url{https://huggingface.co/briaai/FIBO}
 to foster future research.
\end{abstract}

%% file: sec/1_intro.tex
\section{Introduction}
\label{sec:intro}

Text-to-image models have transformed the way visual assets are created \cite{rombach2022high, saharia2022photorealistic, ramesh2022hierarchical}. What began as casual tools for generating playful pictures has rapidly evolved into professional-grade systems delivering substantial creative and commercial value. Within only a few years, these models have achieved unprecedented levels of realism and visual fidelity \cite{cai2025hidream, flux2024, esser2024scaling, wu2025qwenimagetechnicalreport}.

However, most existing models are trained to map short natural language prompts into highly detailed images. This asymmetric training creates a fundamental gap between the sparse representation of user input and the rich details required to produce a high-quality image. As a result, the model often fills in missing information arbitrarily, “guessing” the user’s intent ~\cite{orgad2023editing,wu2024stable,rassin2024grade, huberman2025image}. To mitigate this, many models are tuned toward “average” human preferences, optimizing for majority choices in user studies \cite{wallace2024diffusion, kirstain2023pick, liu2025improving}. While this yields pleasing results for casual users, it limits both precision and creativity for professionals who require fine-grained control over composition, lighting, depth of field, and other visual factors. Indeed, the adage “an image is worth a thousand words” highlights the mismatch: short prompts cannot fully specify an image.

To address this limitation, we propose training text-to-image models with \textit{long structured captions}. These captions encode significantly more visual detail than any prior open-source dataset, and their structured form ensures that each training image is paired with a precise textual description. 
Unlike existing approaches that bias toward human preference, our captions are designed to maximize expressive coverage, enabling the model to generate every plausible visual configuration. 
We show that models trained with long structured captions exhibit not only improved prompt adherence but also superior controllability: their disentangled representations allow modification of specific visual factors while leaving others unchanged. Since users may struggle to author such lengthy prompts, we employ a vision–language model (VLM) to bridge natural human intent and structured prompts, making detailed prompting practical in real-world use.

Yet, training with long captions introduces a new efficiency challenge. To address this, we propose \textit{DimFusion}, a more efficient text–image fusion mechanism. Like prior methods, DimFusion leverages intermediate tokens from an LLM and integrates them directly with image tokens. Unlike them, it combines LLM layers without increasing the number of tokens. Remarkably, we find that even a relatively small LLM suffices for encoding structured prompts.

Beyond modeling, we propose a new evaluation paradigm, \emph{Text-as-a-Bottleneck Reconstruction (TaBR)}, specifically designed to assess alignment with extremely long captions. Since humans struggle to reliably evaluate prompts containing thousands of words, we anchor the evaluation in images. TaBR integrates an image captioner into the pipeline and measures expressive power via a reconstruction protocol: given a real image, we caption it, regenerate it from the caption using the text-to-image model, and evaluate similarity to the original. While similarity can be measured automatically, we find human judgments to perform better. This image-grounded evaluation is more objective than standard user-preference scoring, reflects the expressive power of the caption–generation loop, and enables systematic optimization toward controllability rather than preference-driven popularity.

Finally, we introduce \emph{\modelname{}}, a large-scale text-to-image model capable of generating an image from structured long captions. \modelname{} demonstrates the effectiveness of our approach, achieving state-of-the-art prompt adherence even for captions exceeding 1,000 words. Beyond fidelity, \modelname{} exhibits novel disentanglement capabilities, enabling intuitive and fine-grained control over individual image factors such as color, expression, or composition without unintended changes elsewhere. To foster community progress, we release \modelname{}’s model weights, code, and full implementation details.
Our contributions are as follows:
\begin{itemize}
    \item  We release \emph{\modelname{}}, the first open-source text-to-image model trained entirely on long structured captions.
    \item We introduce \emph{DimFusion}, a novel mechanism to efficiently fuse intermediate LLM tokens into image generation models.
    \item We propose \emph{TaBR}, a new evaluation protocol that measures expressive power via caption–generation–reconstruction, shifting focus from user preference optimization to controllability.
\end{itemize}

%% file: sec/2_related.tex
\section{Related Work}

\input{Figures/workflow/workflow}

\paragraph{Text-to-image models.}
Diffusion models have rapidly become the workhorse of text-to-image synthesis. Early systems, like GLIDE~\cite{nichol2021glide}, Imagen~\cite{saharia2022photorealistic}, and DALL·E~2~\cite{ramesh2022hierarchical}, established the power of conditioning diffusion on strong language encoders. Latent diffusion~\cite{rombach2022high} made large-scale training practical, and SDXL~\cite{podell2023sdxl} pushed UNet-based scaling. A newer wave pivots to transformer backbones and flow-matching objectives: Stable Diffusion 3~\cite{esser2024scaling} adopts DiT-style architectures with bidirectional text–image mixing, while FLUX~\cite{flux2024}, HiDream-I1~\cite{cai2025hidream}, and Qwen-Image~\cite{wu2025qwenimagetechnicalreport} further advance this paradigm. Together, these advances mark a shift toward models that must understand and accurately follow long prompts---the challenge our work is designed to tackle.

\paragraph{Synthetic captions.}

Early text-to-image models were trained on large-scale web datasets such as LAION~\cite{schuhmann2022laion}, where captions derived from HTML alt text provided only coarse and noisy supervision. 
DALL·E~3~\cite{betker2023improving} demonstrated that augmenting training data with synthetically generated, descriptive captions improves both human and automated evaluations, allowing models to capture visual concepts that human annotators often omit. 
Subsequent works~\cite{esser2024scaling, liu2024playground} explored hybrid datasets that mix human-written and synthetic captions to balance linguistic diversity and descriptive precision. 
We extend this idea by using modern vision-language models to produce structured JSON captions that explicitly describe object attributes, spatial relations, composition, and photographic style—capturing fine-grained detail beyond prior open-source efforts. 
The same structured schema is used consistently during training and inference, enabling human-written prompts to be automatically converted into machine-readable structured descriptions.

\paragraph{LLM Fusion.} 

Earlier text-to-image models relied on CLIP~\cite{radford2021learning} or T5~\cite{raffel2020exploring} as text encoders~\cite{rombach2022high,podell2023sdxl,saharia2022photorealistic}, mapping raw text to a compact, semantically meaningful representation. Recent work explores leveraging LLMs for text encoding to exploit their richer semantics and compositional reasoning. Because different intermediate LLM layers encode complementary linguistic information~\cite{durrani2020analyzing,dar2022analyzing,rombach2022high}, several methods, e.g., Playground v3~\cite{liu2024playground} and Tang et al.~\cite{tang2025exploring}, fuse multiple layers into diffusion transformers via cross-attention. However, these approaches typically lack bidirectional text–image mixing (i.e., joint attention), which has been shown to improve prompt adherence~\cite{esser2024scaling}. HiDream-I1~\cite{cai2025hidream} use a dual-stream, decoder-only LLM but at high computational cost, since each attention block consumes tokens from both final and intermediate layers. Our proposed DimFusion improves efficiency by leveraging both intermediate and final LLM representations while keeping token length constant.

\paragraph{Cycle-consistency-based evaluation.} 
Cycle consistency is the principle that translating an input from one domain to another and back should approximately reconstruct the original (e.g., image$\rightarrow$text$\rightarrow$image). A prominent example is the self-supervised objective in CycleGAN~\cite{zhu2017unpaired}. In text-to-image generation, recent works use cycle consistency as a reward in reinforcement settings to improve prompt adherence~\cite{meng2025image,bahng2025cycle}, and as an automatic metric for text–image alignment~\cite{huang2025image2text2image}. Meng et~al.~\cite{meng2025image} further propose an automated evaluation based on image regeneration via captions, but without a direct pixel-level or perceptual comparison to the original image. In contrast, we introduce \emph{Text-as-a-Bottleneck Reconstruction (TaBR)}, which given real image, produce a detailed caption, regenerates the image from this caption, and then explicitly compares the reconstruction to the original. This image-grounded protocol provides a more direct measure of expressive power and controllability.

%% file: Figures/workflow/workflow.tex
\begin{figure}[t]
  \centering
  \includegraphics[width=\linewidth]{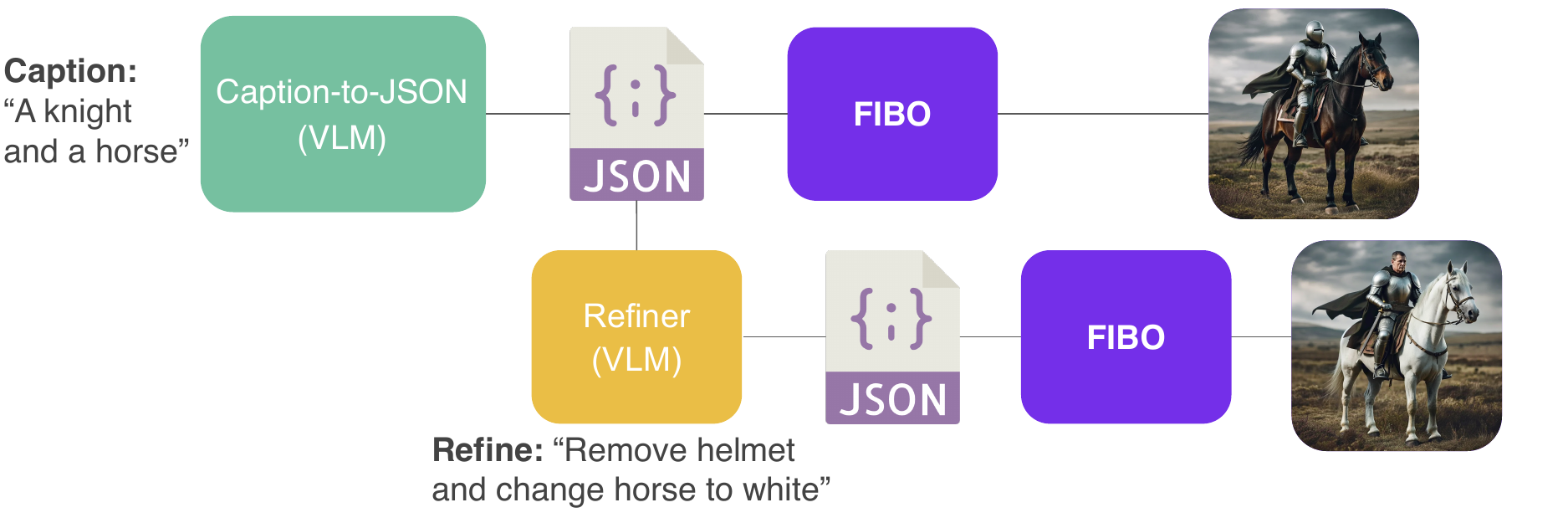}
  \caption{\textbf{Workflow.} A short caption is expanded by a VLM into a detailed JSON used by \modelname{} to generate an image. The user can then refine the JSON, and \modelname{} produces a new image that reflects only the requested changes, demonstrating strong disentanglement between modified and preserved elements. 
  }
  \label{fig:workflow}
\end{figure}

%% file: sec/3_method.tex
 \input{Figures/contextual_contradictions/contextual}
\input{Figures/control/control_examples}

\section{Method}
\label{sec:method}

In this section, we present the main contributions of the paper. Section~\ref{sec:captions} introduces the motivation for training with long structured prompts and analyzes their effect on controllability and disentanglement. We also present our workflow, as demonstrated in Figure~\ref{fig:workflow}.
Section~\ref{sec:dimfusion} introduces \emph{DimFusion}, our novel architecture for integrating an LLM with a text-to-image model under reduced computational requirements. 
Section~\ref{sec:eval} presents a novel evaluation protocol tailored for long captions, addressing the limitations of existing methods when handling inputs exceeding 1{,}000 words. Finally, Section~\ref{sec:training} describes our large-scale training setup, resulting in our model, \textit{\modelname{}}. Additional implementation details are provided in the Appendix.

\subsection{Training with Long Structured Captions}
\label{sec:captions}

While the prompt adherence of text-to-image models has steadily improved, even state-of-the-art models often struggle with complex scenes. For example, current models fail to generate contextual contradictions, reliably capture distinct facial expressions for multiple individuals, or to respect subtle specifications such as lighting conditions and camera parameters, as can be seen in Figures~\ref{fig:contextual} and \ref{fig:control}.



We attribute this to a fundamental limitation in training data: most captions lack detailed descriptions of fine-grained visual factors. Recent works have demonstrated that training with synthetic captions outperforms the use of default human-written captions~\cite{esser2024scaling, liu2024playground}.
However, synthetic captions are not exhaustive: some details are omitted depending on the judgment of the captioning model. As a result, crucial information is often absent during training, introducing ambiguity for the generator. Consequently, the model learns to ignore prompt instructions often, as it has become accustomed to incomplete conditioning.

To address this, we train with \textit{long structured captions}, in which crucial details are always represented. By ensuring that essential aspects such as colors, lighting, or composition are consistently present, we remove ambiguity and establish a stronger alignment between text and image. This explicit conditioning encourages the generator to ground its outputs in the caption rather than relying on priors.

We observe that training with long structured captions results in faster convergence and improved image quality compared to short captions. To validate this, we train a low resolution small model ($1$B parameters) for $100$K steps using the same images, comparing between long structured captions and short captions. Both qualitative and quantitative comparisons are in Figure~\ref{fig:caption-length-merged} demonstrate that training with long structured captions converges faster and achieves superior results compared to short captions. Complete experimental details are provided in the Appendix (Section \ref{sec:appendix_ablation}.

\input{tables/caption-length-fid-comparison}

Surprisingly, disentanglement emerges natively from our structured representation in an unsupervised manner—the model never observes paired images during training. In standard text-to-image models, modifying a single prompt detail often induces unintended changes in unrelated factors. In contrast, our model enables precise editing of generated images: altering one attribute in the structured caption typically affects only the corresponding visual factor, without requiring additional editing techniques~\cite{hertz2022prompt,mokady2023null}. This disentanglement arises naturally from conditioning on consistently detailed prompts and is illustrated in Figure~\ref{fig:disentanglement}. Notably, this is not an image-editing setup, as the model does not take an input image. We modify only the structured prompt while holding the random seed fixed.

\input{Figures/disentanglement/dis_fig}

The full schema of our structured captions is provided in the Appendix (Section \ref{app:structured-json}). Each caption begins with the most salient details of the image, including objects, background, and text. Objects are described with rich attributes such as size, position, shape, and color. Humans are also annotated with pose, expression, ethnicity, etc. We then incorporate information that is often missing in captions, including lighting, depth of field, and composition. To generate these descriptions, we employ a state-of-the-art VLM~\cite{comanici2025gemini} to produce the structured captions. We find that current VLMs excel at providing objective descriptions of visual content, though they are less reliable in subjective judgments, where they tend to produce flattering assessments. To address this, we complement our captions with scores from two dedicated aesthetic predictors~\cite{kirstain2023pick,laion2022_aesthetics_v1}, which quantify the aesthetic quality of each image. Caption-length statistics are reported in Table~\ref{tab:caption_stats}, demonstrating that the vast majority are long and detailed.

\input{tables/captions_stats}


While structured captions substantially enhance expressive power, we cannot expect users to author 1{,}000\,word prompts in a strict schema. To make it practical, we integrate a dedicated vision–language model (VLM) that supports three complementary operations: $(1)$ \emph{Generate}: expand a short prompt into a structured prompt. $(2)$ \emph{Refine}: modify an existing structured caption given editing instructions (see Figure~\ref{fig:workflow}). $(3)$\emph{Inspire}: extract a structured caption from an input image and optionally edit it; the image serves as creative guidance. Although the VLM adds computational overhead, it makes the pipeline far more capable: users can start from either a brief prompt or an existing image, iteratively refine via simple text instructions, and benefit from the VLM’s world knowledge, reasoning ability, and multilingual support.

For the VLM, a state-of-the-art model such as Gemini~2.5~\cite{comanici2025gemini} is effective but costly. To provide an efficient alternative suitable for community use, we fine-tune an open-source model, \mbox{Qwen-3~VL} 4B~\cite{qwen2.5-vl}. We train on synthetically generated short prompts and editing instructions, using the same structured schema employed for our image model to ensure tight alignment. Training is performed on $8\times$H100 with a total of $2$B tokens. To improve robustness, we decouple image-conditioned and text-only tasks during training and repeat each with different seeds, then final weights are produced via model merging~\cite{yang2024model}.

\subsection{DimFusion Architecture}
\label{sec:dimfusion}


To fully benefit from our structured captions, the model must process long sequences of text tokens and fuse them with its internal image tokens. Recent works highlight the advantage of using LLMs as text encoders, leveraging the strong semantic and compositional understanding encoded in their intermediate layers~\cite{durrani2020analyzing,dar2022analyzing,rombach2022high}. To unlock this potential, text tokens can be fused with image tokens through bi-directional mixing~\cite{esser2024scaling}, and information from both last and intermediate layers can be combined to maximize expressivity~\cite{cai2025hidream}. A key challenge in leveraging long captions is the prohibitive cost of processing thousands of text tokens. Naively combining intermediate layers by concatenating them along the sequence dimension increases token length, leading to substantial growth in attention computation. This challenge is especially pronounced in early pre-training stages, where models are trained on low-resolution images that produce far fewer image tokens than text tokens, causing the attention cost to scale with caption length.

We present \emph{DimFusion}, an efficient mechanism for fusing intermediate LLM representations into a text-to-image model. Rather than extending the text sequence length, DimFusion concatenates intermediate layers along the \emph{embedding dimension}, keeping the number of tokens fixed. This significantly reduces computational cost, particularly when handling long captions.

In more details, the \textbf{embedding dimension} $d_{\text{model}}$ is the size of each token’s feature vector—the last axis in tensors of shape $(B, L, d_{\text{model}})$. A text encoding is formed by concatenating the last two layers of the LLM along the embedding dimension and projecting the result to fit $D/2$, where $D$ is the transformer’s hidden dimension. At each subsequent block, hidden states from the corresponding LLM layer are projected to $D/2$ and concatenated with the current text encoding, restoring the dimension to $D$. After processing the block, the extra $D/2$ components are discarded, returning the text encoding to $D/2$. This design enables progressive fusion: each block integrates complementary information from intermediate LLM states while keeping token length fixed. Because half of each token is always preserved, DimFusion naturally supports dual-stream blocks followed by single-stream blocks, as in~\cite{esser2024scaling}. 
The overall architecture is illustrated in Figure~\ref{fig:arch}.

\input{Figures/dimfusion_arch}

We conducted an ablation study using a $1$B-parameter transformer with SmolLM3-3B as the LLM text encoder. 
As a baseline, we trained with the final layer of T5-XXL as the text encoding and no LLM fusion. 
Following HiDream~\cite{cai2025hidream}, we then evaluated \emph{TokenFusion}, where the text encoding is concatenated with the hidden states of the corresponding LLM block along the \emph{sequence} dimension. 
Finally, we assessed our proposed \emph{DimFusion}. 
All models were trained for 150k steps on 128 H200 GPUs with the same batch size (full training details appear in the appendix).
Figure~\ref{fig:dimfusion-merged} reports the training loss, FID and average time per step for these configurations. Both TokenFusion and DimFusion substantially outperform the T5-XXL baseline, confirming the value of LLM representations. 
Notably, DimFusion achieves a slightly better FID than TokenFusion while reducing the average step time from $0.8$ seconds to $0.5$ seconds, which corresponds to a $\sim$1.6× reduction in wall-clock time, demonstrating that DimFusion retains the advantages of deep LLM fusion while being more compute-efficient.

\input{Figures/dimfusion_ablation}

\subsection{Evaluating Long Captions}
\label{sec:eval}

Most text-to-image models are evaluated through simple user-preference protocols, where participants are asked questions such as \textit{``which image do you prefer?''} given outputs from different models, e.g., ~\cite{esser2024scaling}. While this setting is effective for short prompts, it fails for \textbf{long structured captions} that often exceeding 1{,}000 words, since humans struggle to parse such detailed textual descriptions. Moreover, prior work has shown that preference-based evaluation tends to reward a characteristic \textit{``AI aesthetic''}, marked by over-saturated colors and excessive focus on central subjects, a phenomenon recently termed \textit{bakeyness}~\cite{batifol2025flux}. We argue that this evaluation paradigm hinders progress, as it incentivizes models to optimize for short prompts and average aesthetics, rather than expressive power and controllability.

To overcome this, we propose \textbf{Text-as-a-Bottleneck Reconstruction (TaBR)}, a new evaluation protocol designed to suit long captions and generalize across domains and styles. Our key observation is that while humans find it difficult to digest long textual prompts, they can reliably compare complex images. TaBR leverages this asymmetry by introducing the image itself as the anchor for evaluation.

Inspired by cycle-consistency principles~\cite{bahng2025cycle}, TaBR begins with a real image, which is first captioned. This caption is then provided as the sole input to the generator, producing a reconstructed image. Finally, we compare the reconstructed outputs to the original. In practice, we generate two candidate reconstructions and ask annotators: \textit{``which image is more similar to the original?''} Unlike standard preference scoring, user judgments in TaBR are guided by the original image, which conveys far richer detail than a short textual description. By controlling the set of real images used in evaluation, we prevent judgments from collapsing into personal taste and instead obtain a direct measure of the model’s expressive power. An example appears in Figure~\ref{fig:tabr}: our model is the only one that preserves the pose in the second and bottom rows, demonstrating superior controllability using text alone. Quantitative results are reported in Section~\ref{sec:Quantitative_Results}.



\input{Figures/tabr/tabr}
\input{Figures/arch/arch_table}

\subsection{Large Scale Training}\label{sec:training}

\input{tables/prism_licensed}
\input{tables/geneval}

We demonstrate the effectiveness of our contributions through \emph{\modelname{}}, a large-scale text-to-image model with $8$B parameters trained entirely on long structured captions. 
\modelname{} adopts the DimFusion architecture introduced in Section~\ref{sec:dimfusion}, using \emph{SmolLM3-3B}~\cite{bakouch2025smollm3} as the fused LLM backbone. The model operates in the latent space~\cite{rombach2022high} with the Wan~2.2 VAE~\cite{wan2025wan} and a patch size of $1$. 
Additional hyperparameters are summarized in Table~\ref{tab:model_arch}.

Training is performed on $120$M licensed image–caption pairs, where captions are long, structured JSONs generated by \emph{Gemini~2.5}~\cite{comanici2025gemini}, as described in Section~\ref{sec:captions}. We employ progressive training, starting at low resolutions and gradually increasing throughout. We initialize with REPA~\cite{yu2024representation} using a coefficient of $0.1$ for the first $300$K low-resolution steps, followed by mixed-resolution training. After pretraining, we apply aesthetic finetuning with $3{,}000$ hand-picked images, then DPO training~\cite{wallace2024diffusion} with dynamic beta~\cite{liu2025improving} to improve text rendering. Additional implementation details, including optimization and data distribution, are provided in the Appendix.

%% file: Figures/contextual_contradictions/contextual.tex
\setlength{\tabcolsep}{0.5pt}
\renewcommand{\arraystretch}{1.0}

\begin{figure}[t]
\centering
\newcommand{\imgw}{0.24\linewidth} 
\begin{tabular}{@{}cccc@{}}


      \multicolumn{4}{c}{\textit{``A bear is performing a
handstand in the park.''}} \\
  \includegraphics[width=\imgw]{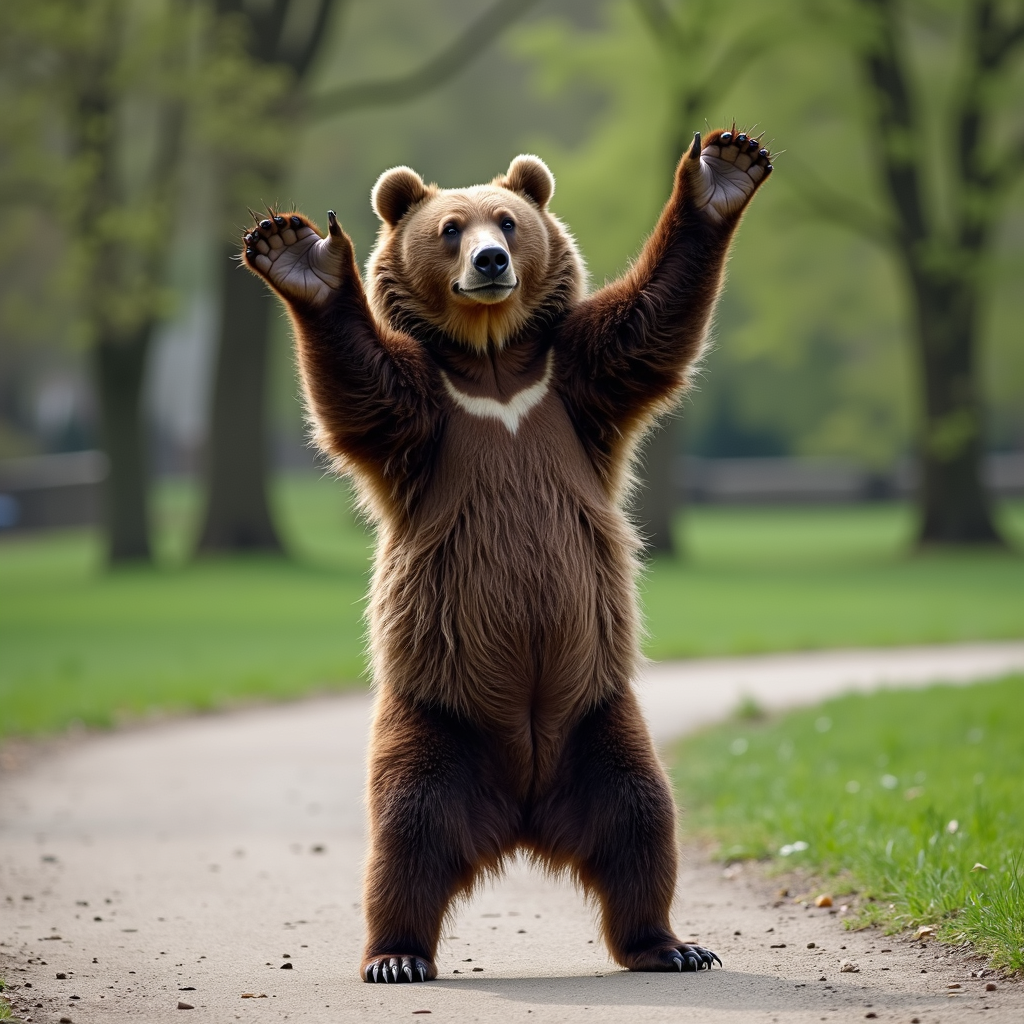} &
  \includegraphics[width=\imgw]{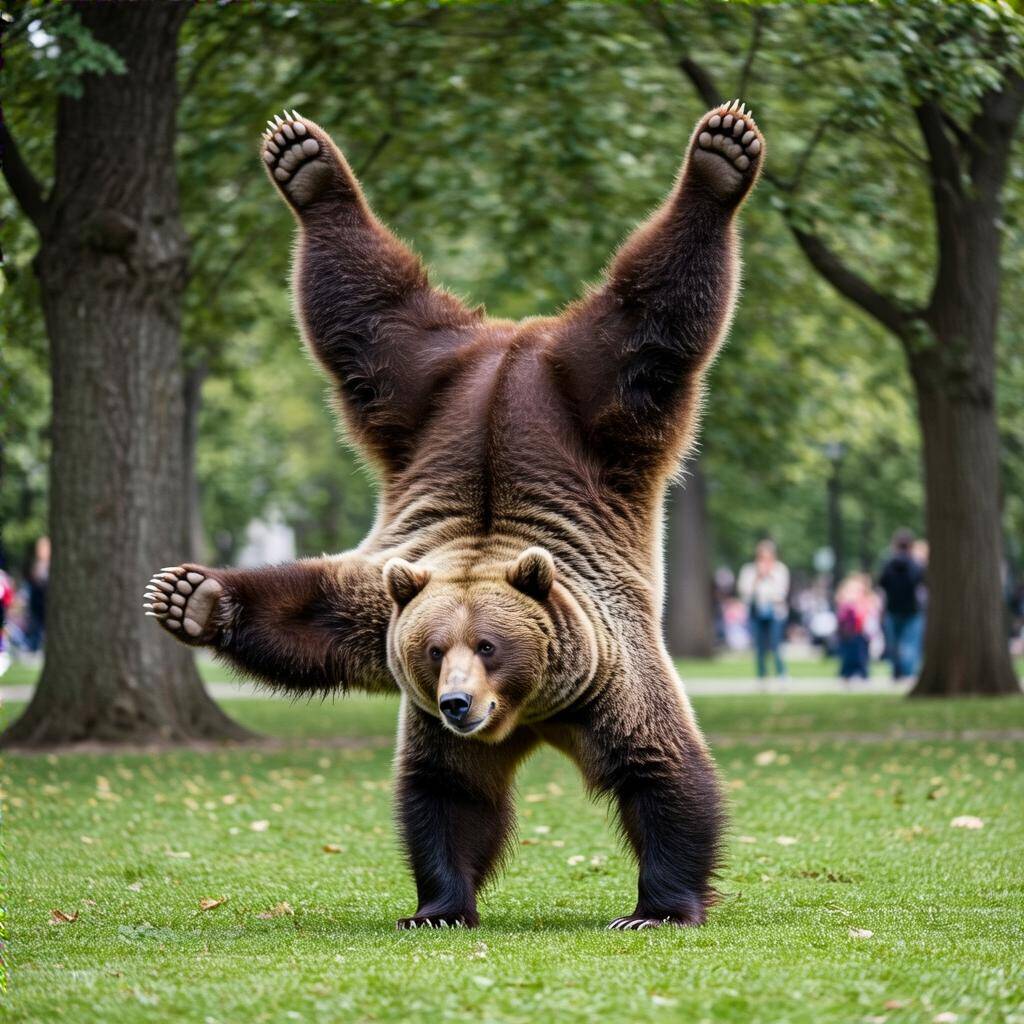} &
  \includegraphics[width=\imgw]{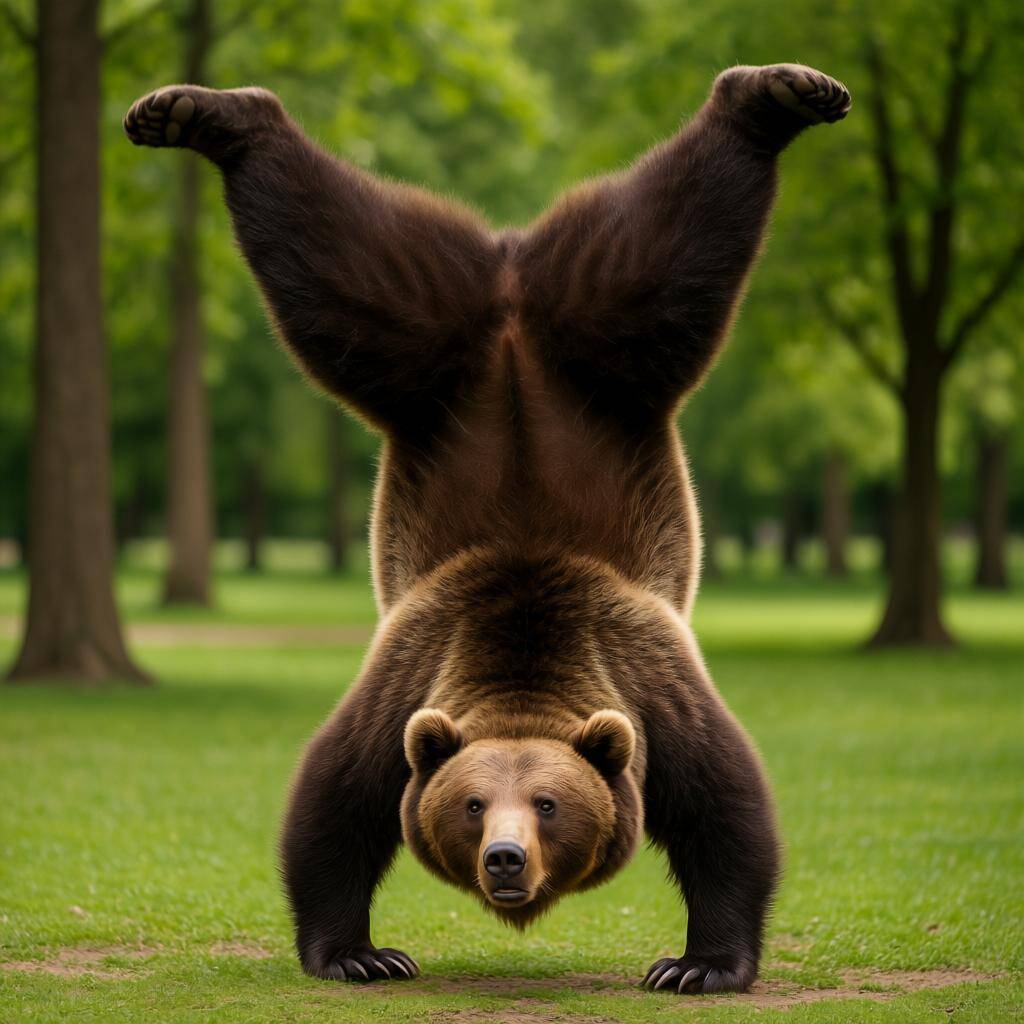} &
  \includegraphics[width=\imgw]{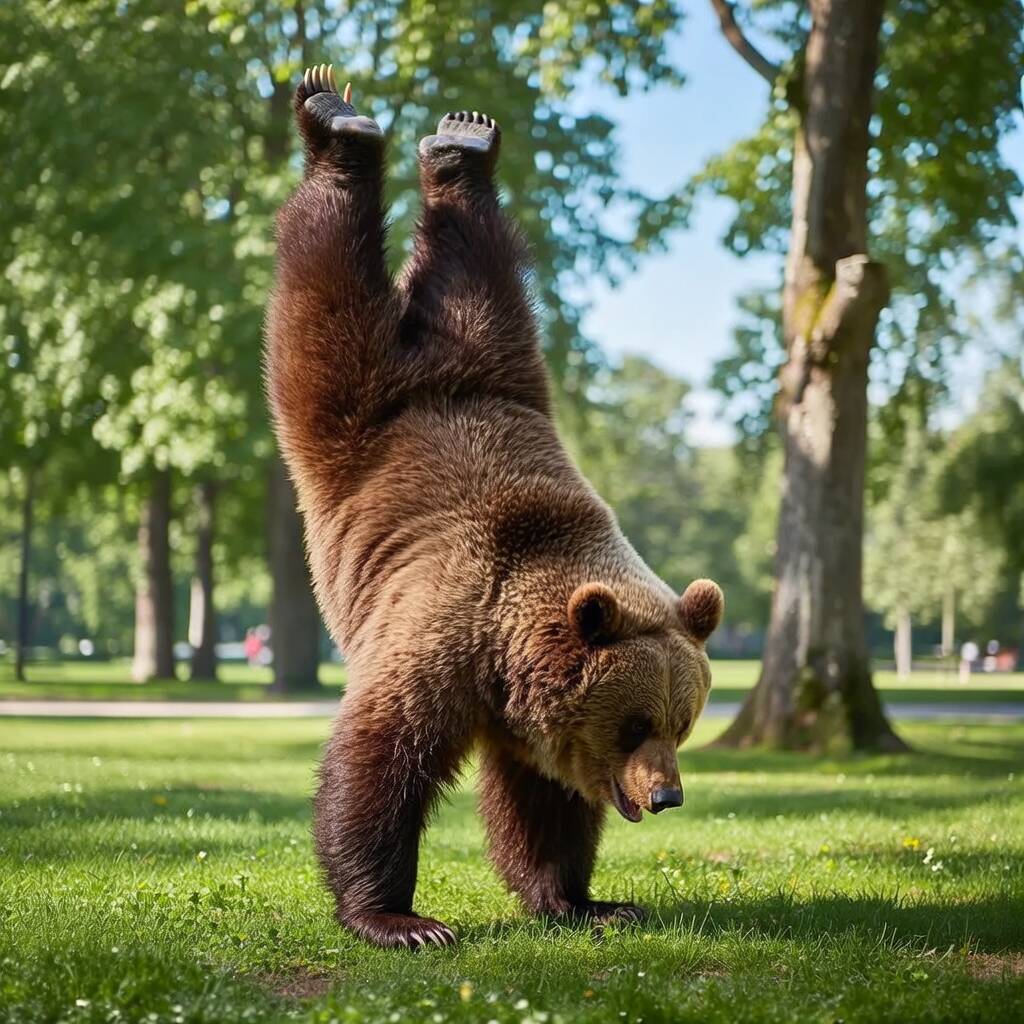}
  \\


    \multicolumn{4}{c}{\textit{``A woman writing with a dart.''}} \\
  \includegraphics[width=\imgw]{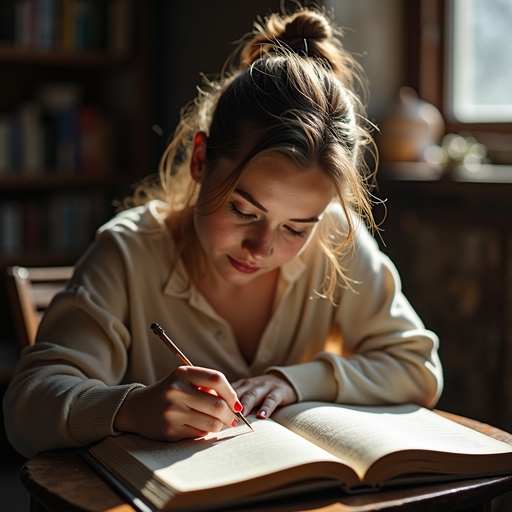} &
  \includegraphics[width=\imgw]{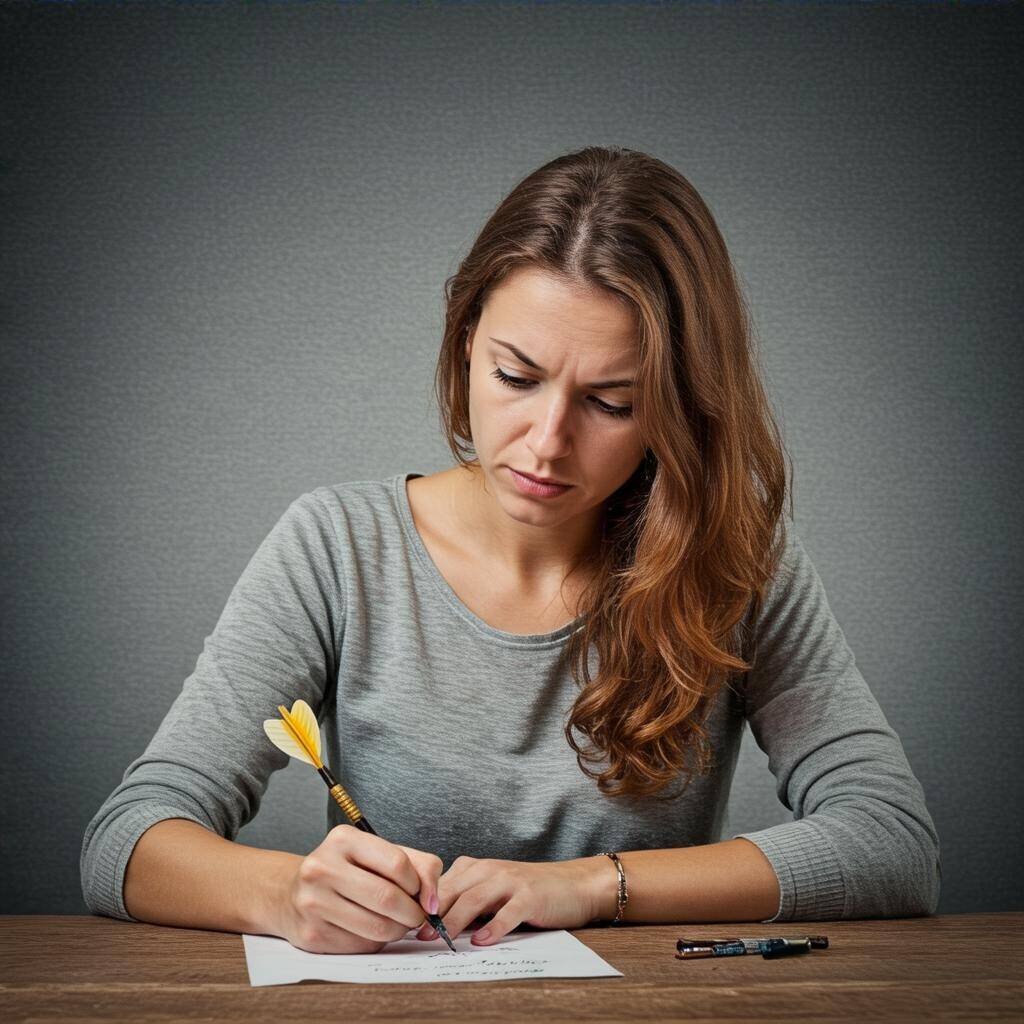} &
  \includegraphics[width=\imgw]{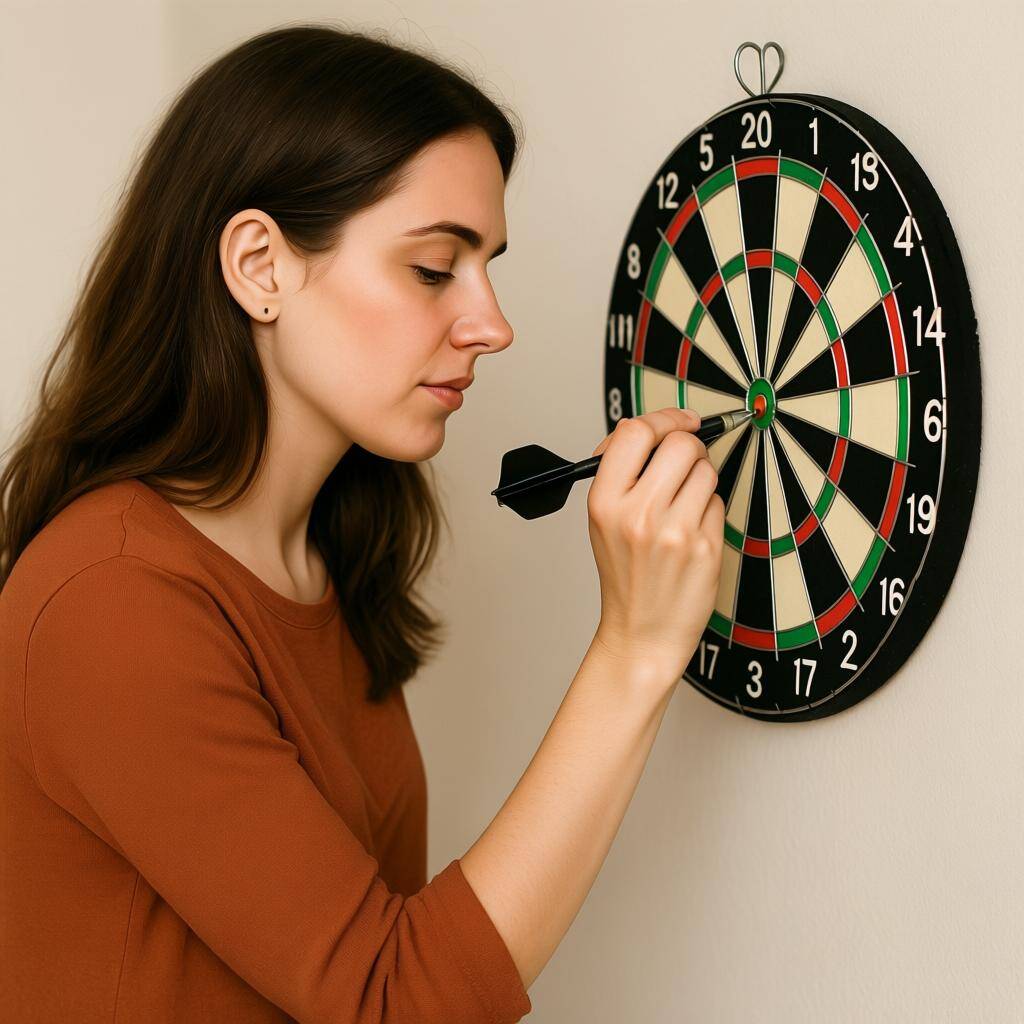} &
  \includegraphics[width=\imgw]{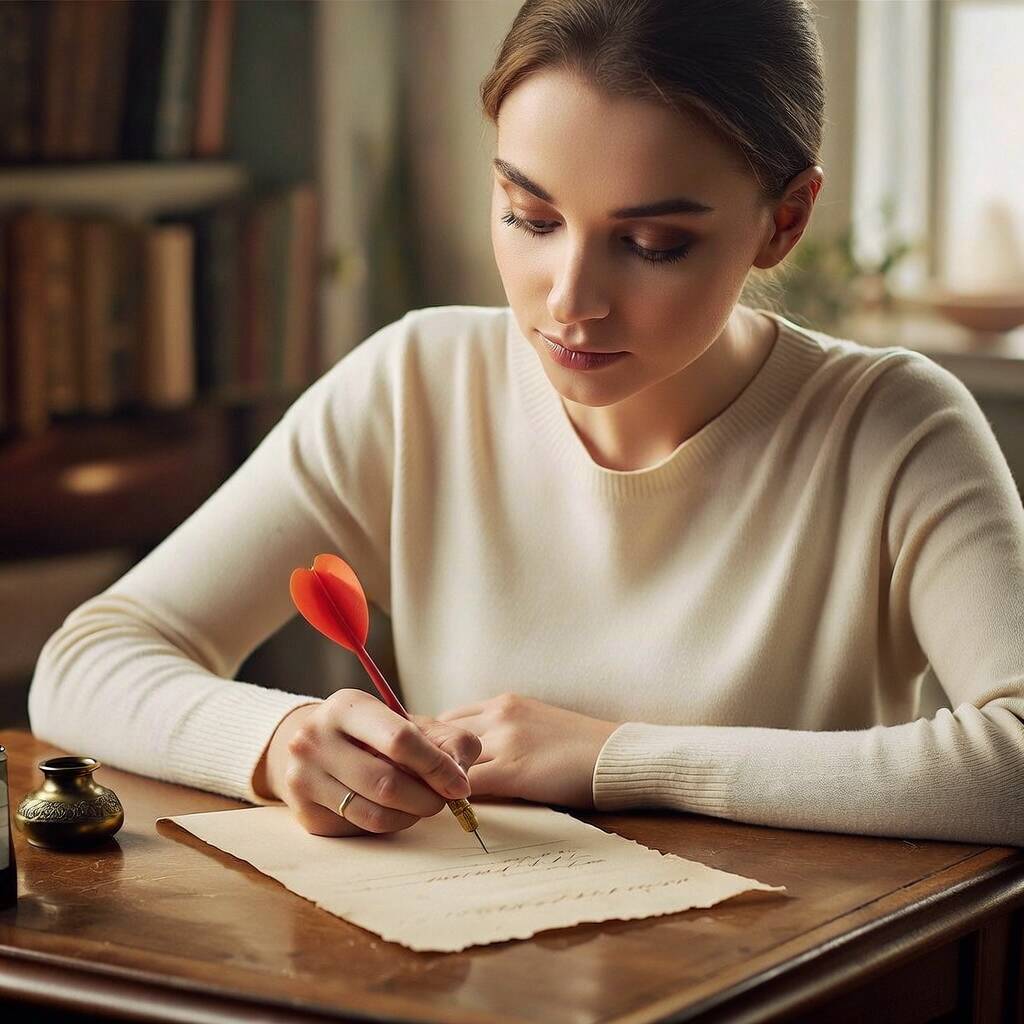}
  \\
  




\multicolumn{4}{c}{\textit{``A professional boxer does a split.''}} \\
  \includegraphics[width=\imgw]{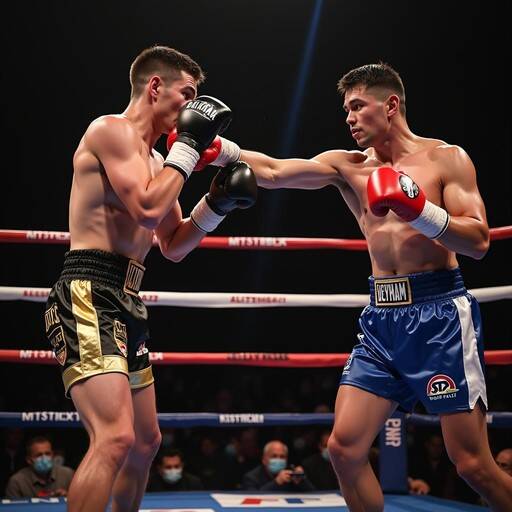} &
  \includegraphics[width=\imgw]{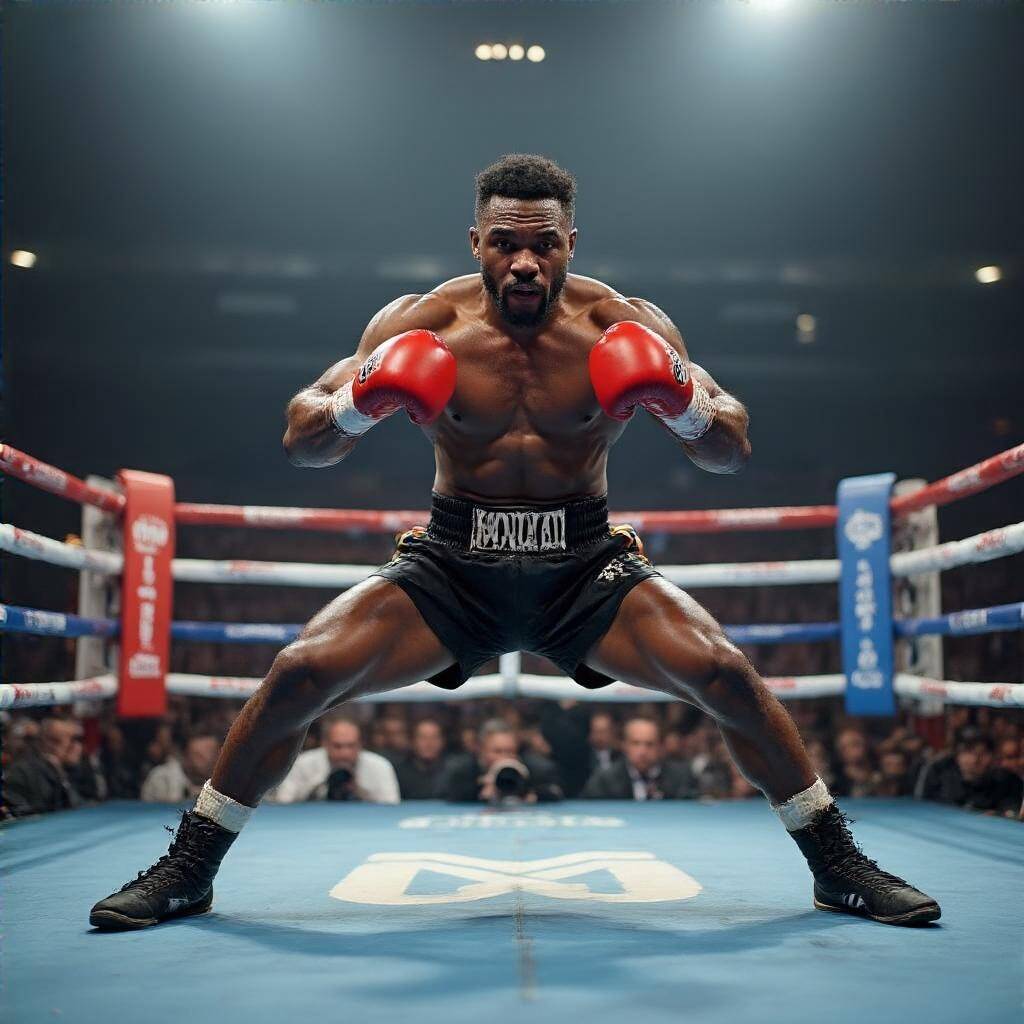} &
  \includegraphics[width=\imgw]{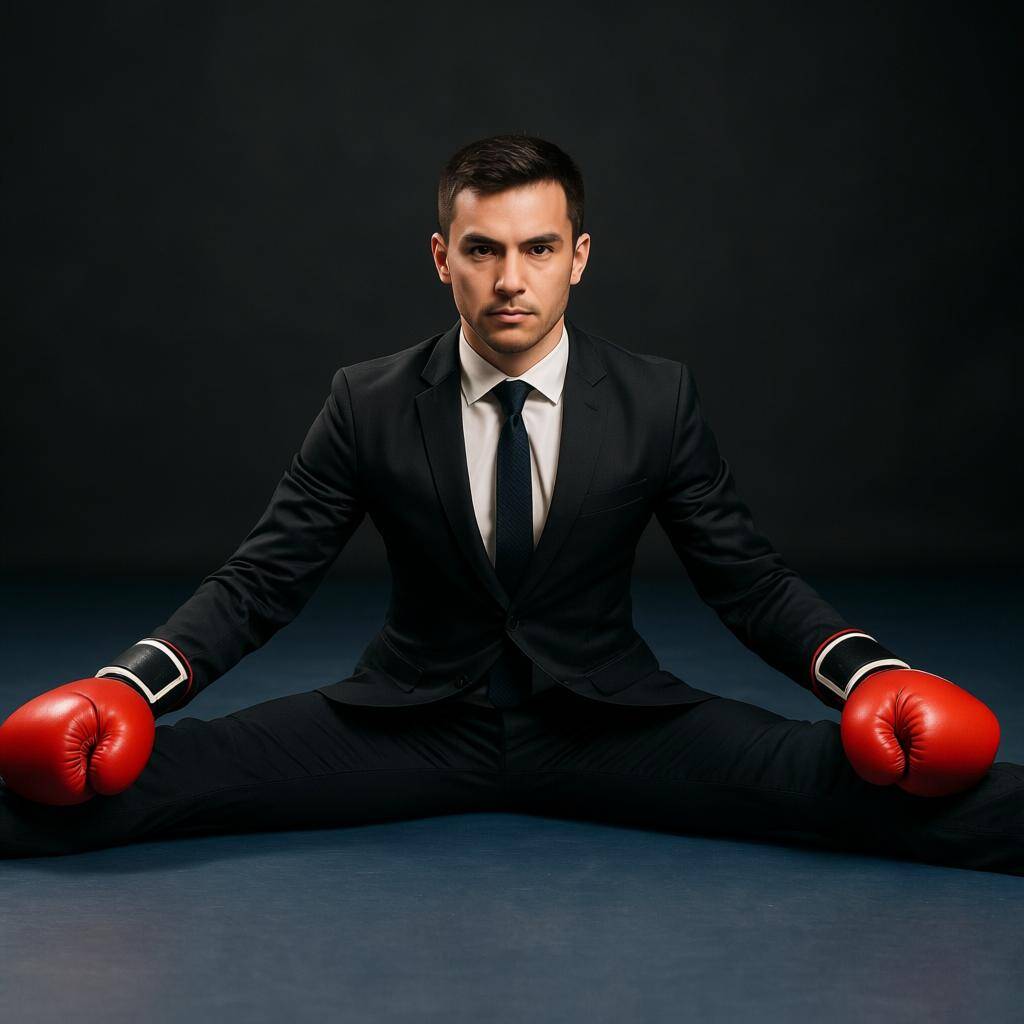} &
  \includegraphics[width=\imgw]{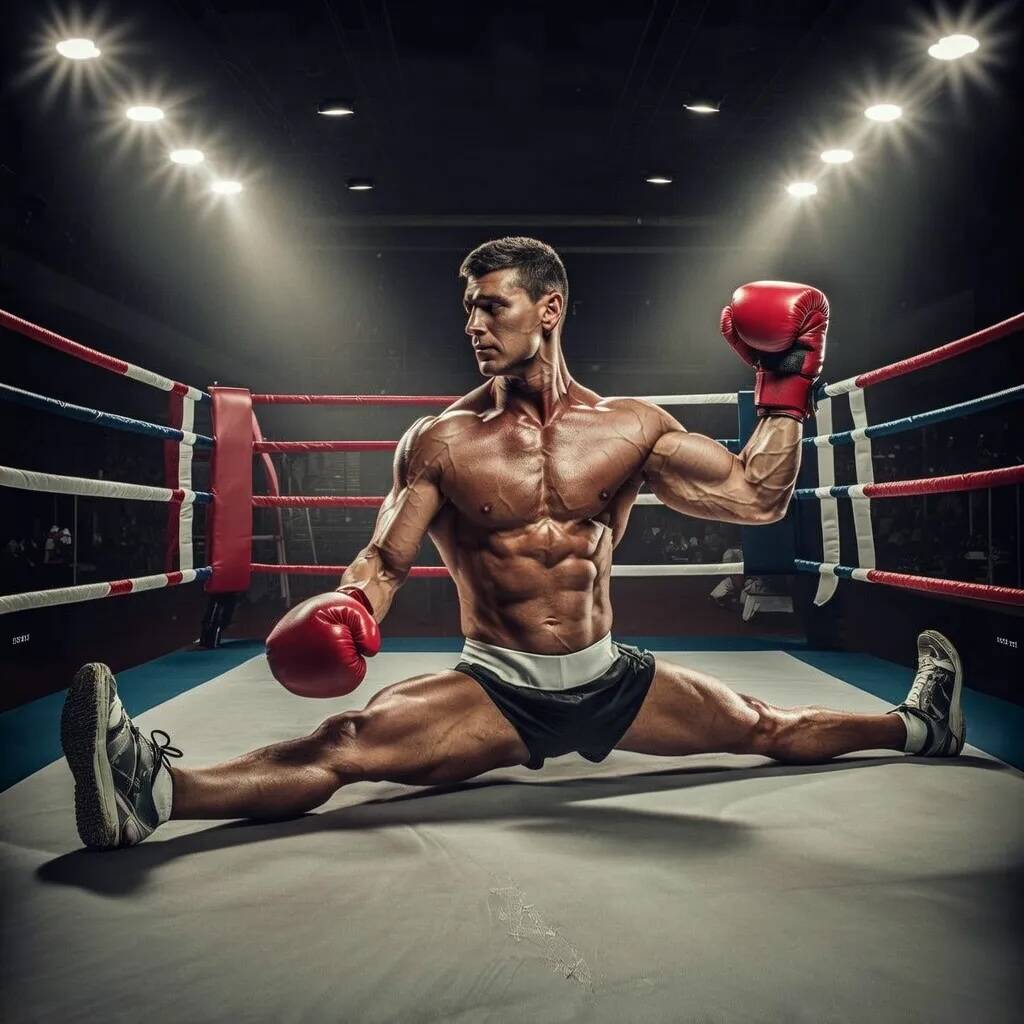}

  \\

  \parbox{\imgw}{\scriptsize\centering \textbf{FLUX}} &
  \parbox{\imgw}{\scriptsize\centering \textbf{HiDream}} &
  \parbox{\imgw}{\scriptsize\centering \textbf{Qwen-Image}} &
  \parbox{\imgw}{\scriptsize\centering \textbf{Fibo (Ours)}}
  \\
\end{tabular}

\vspace{-3pt}
\caption{\textbf{Contextual contradiction.} We use prompts from ContraBench~\cite{huberman2025image} and Whoops~\cite{Bitton_Guetta_2023}. \modelname{} generates semantically consistent images that faithfully represent the correct relations, unlike other models that often default to typical co-occurrences. }
\label{fig:contextual}
\end{figure}

%% file: Figures/control/control_examples.tex
\setlength{\tabcolsep}{0.5pt}
\renewcommand{\arraystretch}{1.0}

\begin{figure}[t]
\centering
\newcommand{\imgw}{0.24\linewidth} 
\begin{tabular}{@{}cccc@{}}

\multicolumn{2}{c}{\footnotesize\textit{``Shallow depth of field''}} &
\multicolumn{2}{c}{\footnotesize\textit{``Deep depth of field''}} \\
  \includegraphics[width=\imgw]{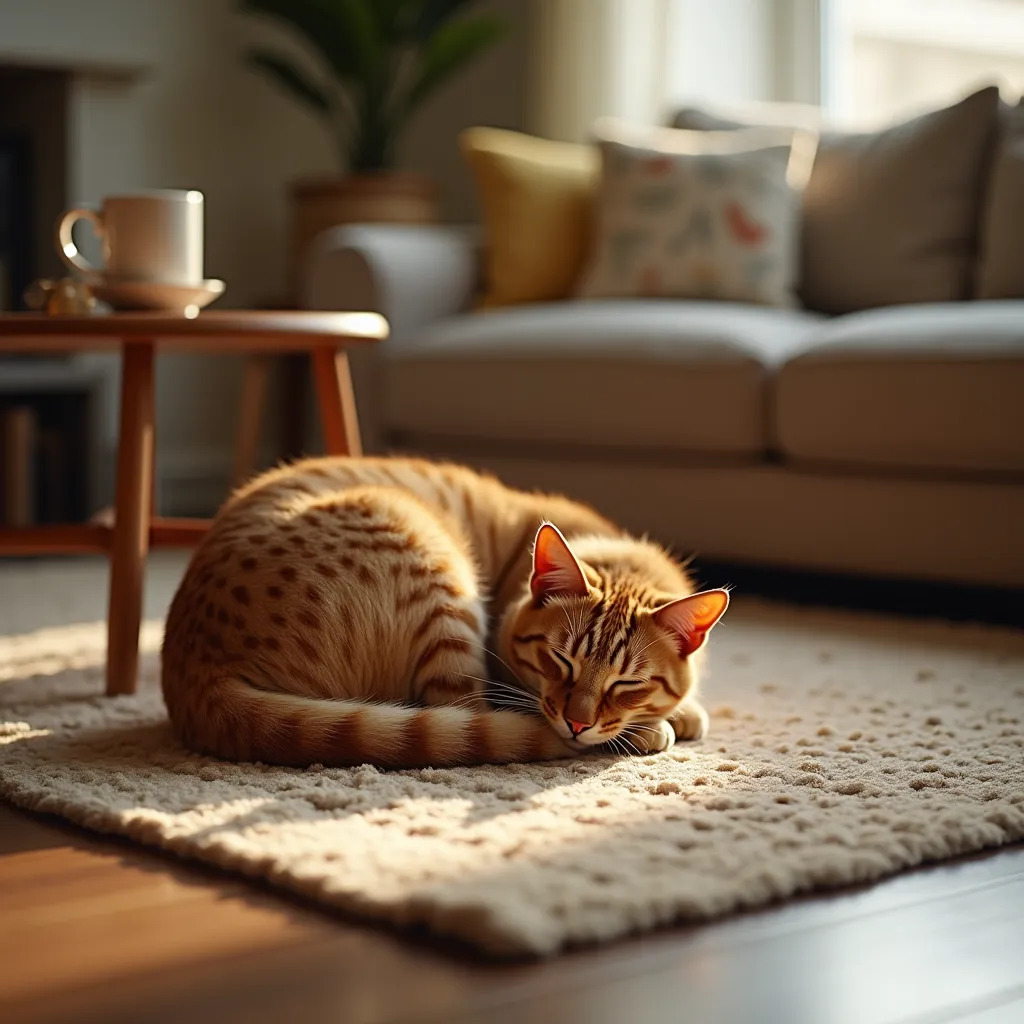} &
  \includegraphics[width=\imgw]{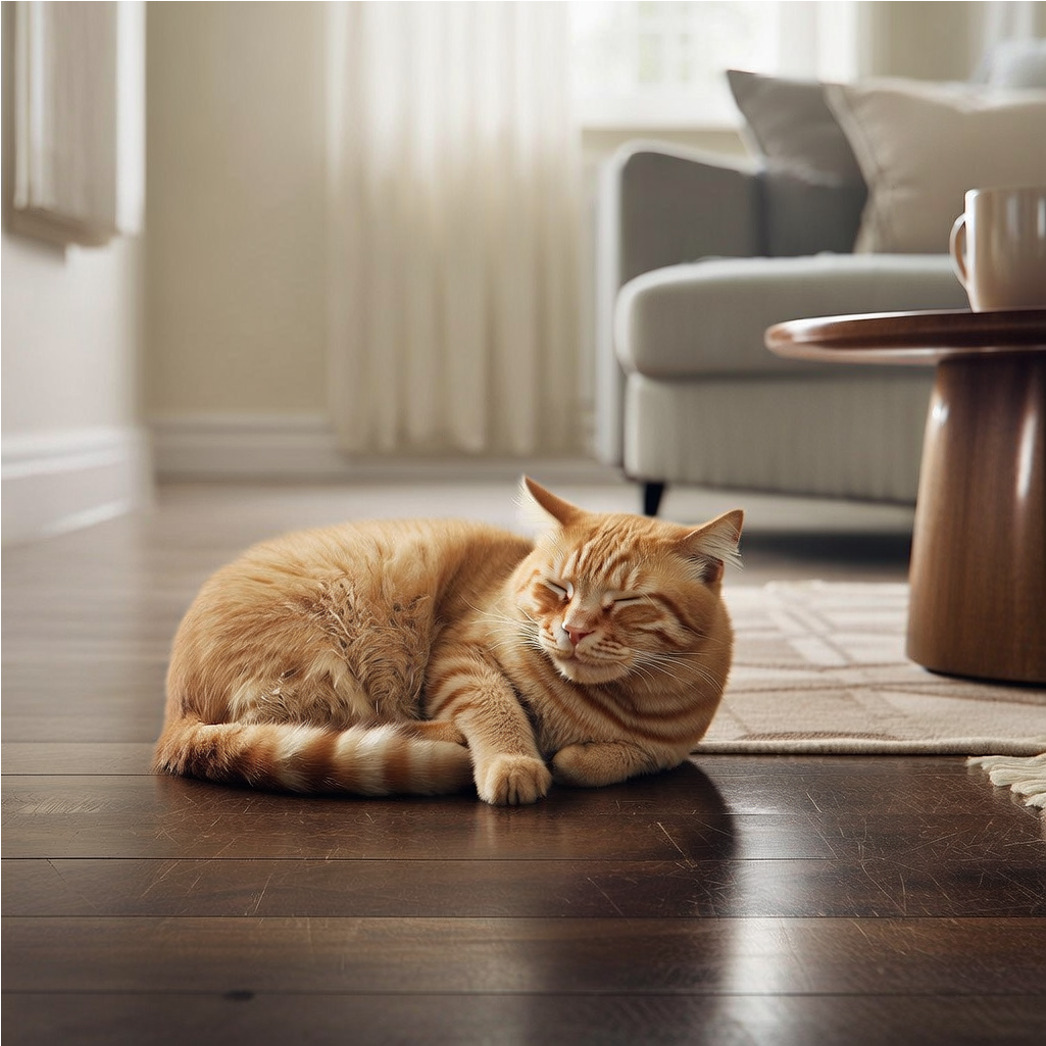} &
  \includegraphics[width=\imgw]{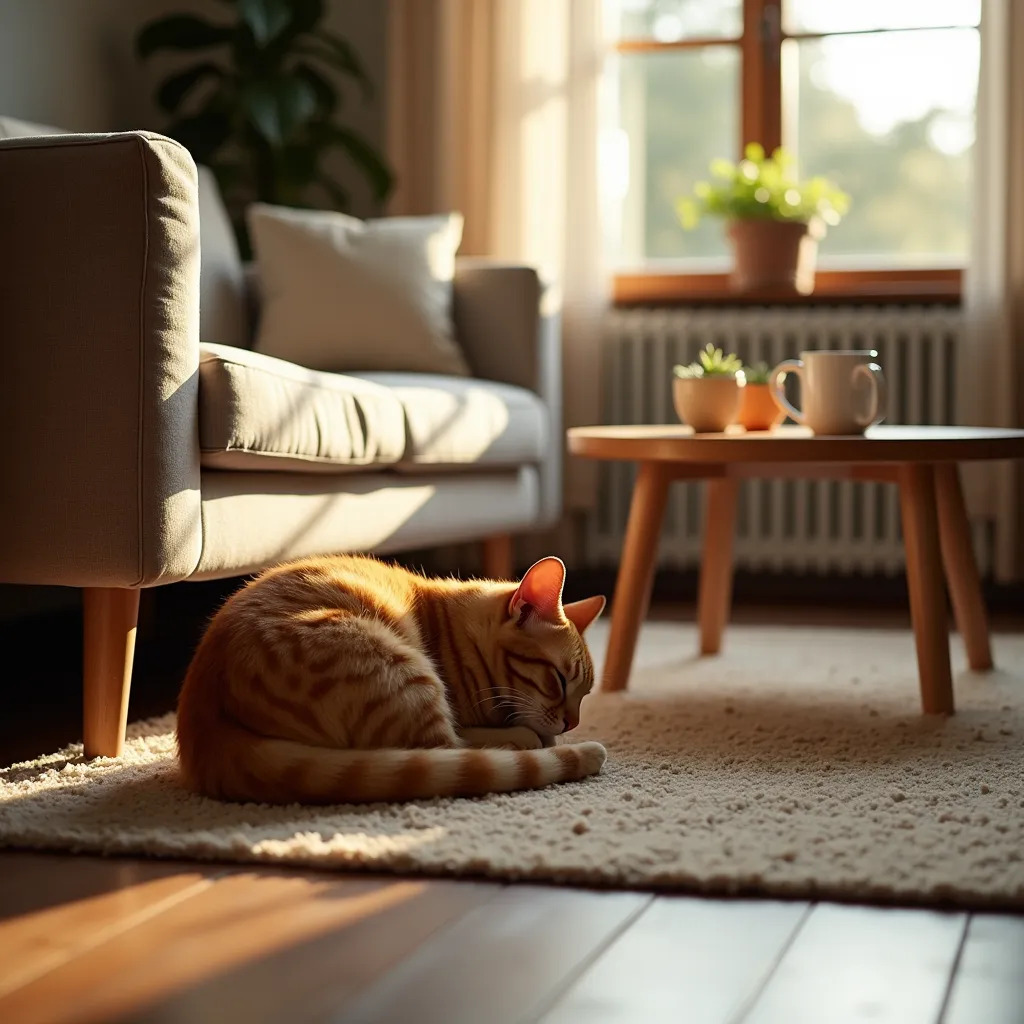} &
  \includegraphics[width=\imgw]{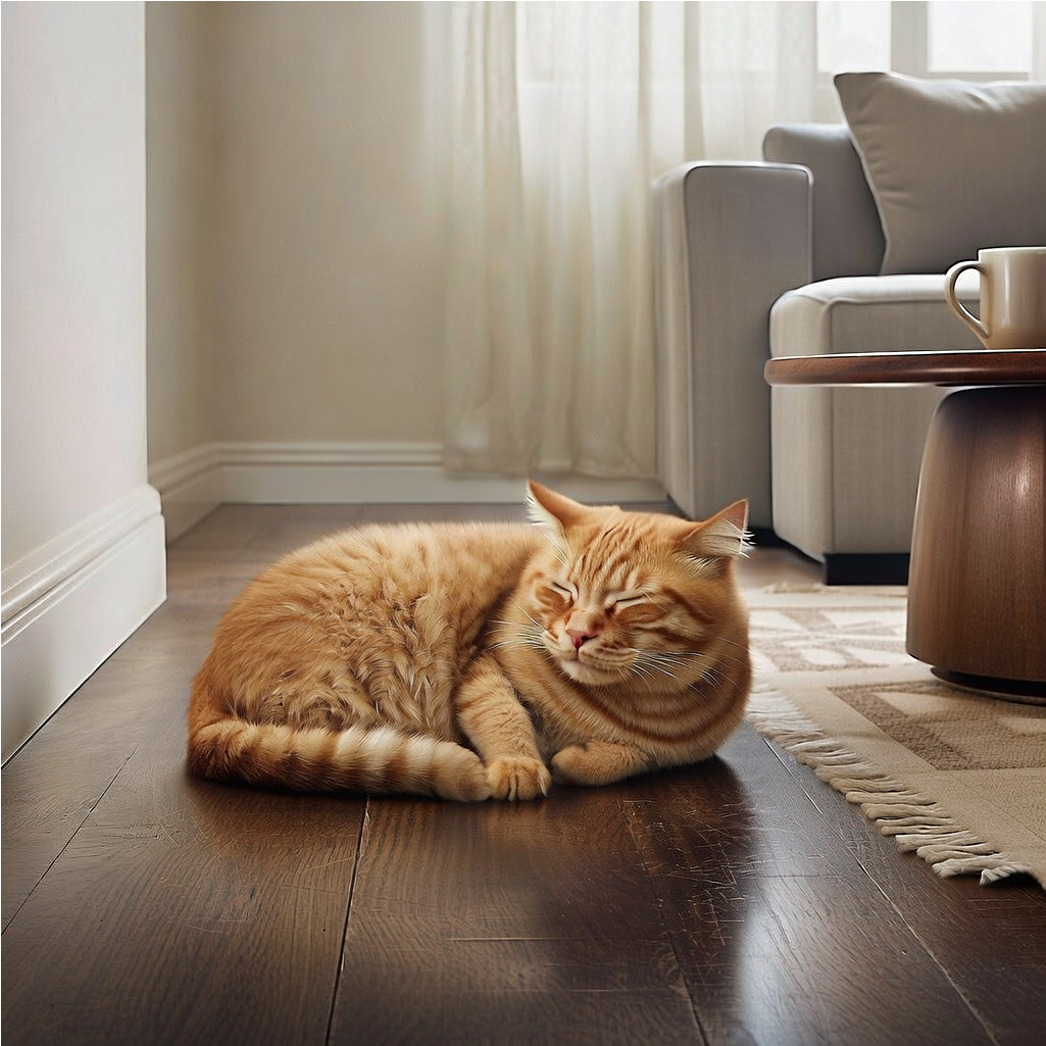}

  \\

  \parbox{\imgw}{\scriptsize\centering \textbf{FLUX}} &
  \parbox{\imgw}{\scriptsize\centering \textbf{\modelname{} (Ours)}} &
  \parbox{\imgw}{\scriptsize\centering \textbf{Flux}} &
  \parbox{\imgw}{\scriptsize\centering \textbf{\modelname{} (Ours)}}
  \\
\end{tabular}

\newcommand{\imgwB}{0.48\linewidth} 
\begin{tabular}{@{}cc@{}}
\multicolumn{2}{>{\centering\arraybackslash}p{0.96\linewidth}}{%
  \footnotesize\textit{``Three people: Left one holding a goose while screaming. Middle is deeply surprised. Third is crying while holding two piglets.''}%
}\\
\includegraphics[width=\imgwB]{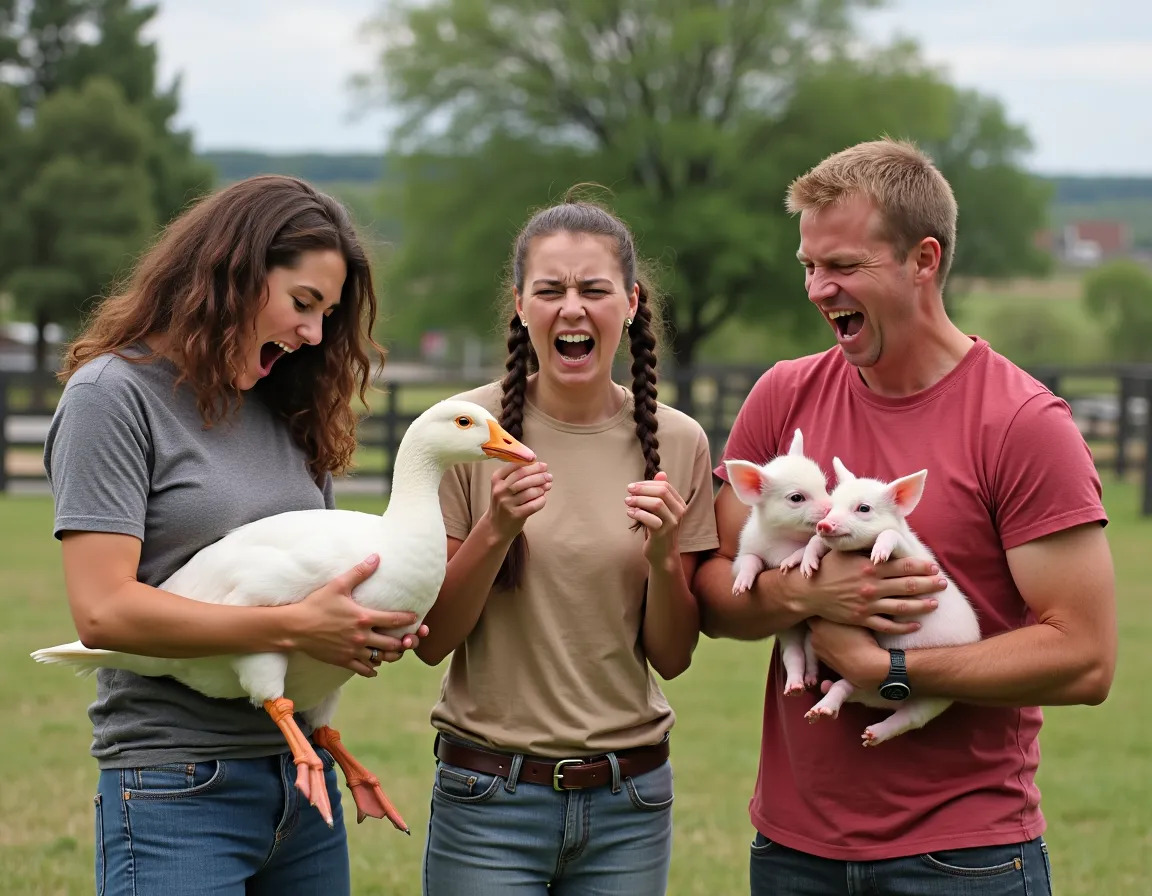} &
\includegraphics[width=\imgwB]{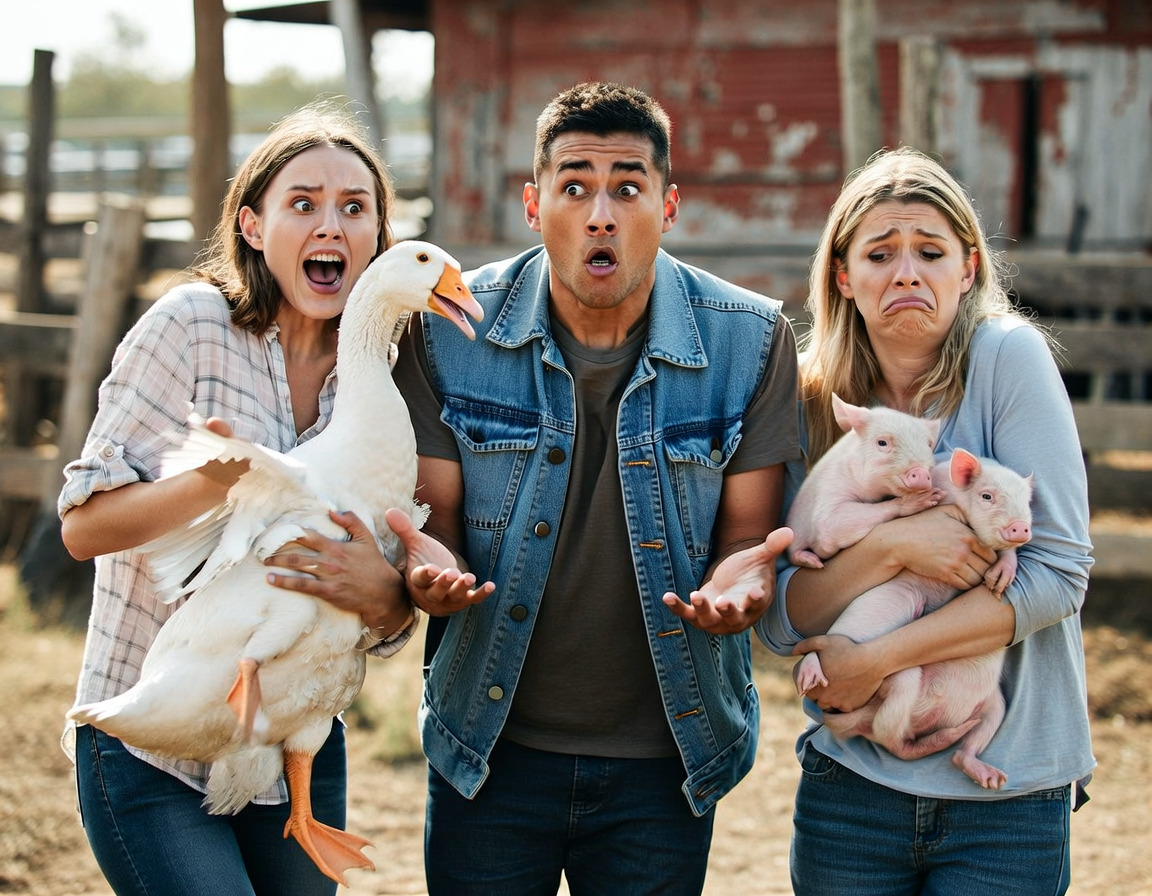} \\
\parbox{\imgwB}{\scriptsize\centering \textbf{FLUX}} &
\parbox{\imgwB}{\scriptsize\centering \textbf{\modelname{} (Ours)}}
\end{tabular}

\vspace{-3pt}
\caption{\textbf{Controllability examples.} \emph{Top:} Our model (\modelname{}) enables precise depth-of-field control, from shallow to deep, whereas FLUX~\cite{flux2024} does not consistently produce deep depth of field. \emph{Bottom:} \modelname{} allows simultaneous control of facial expressions for multiple subjects within the same scene.}
\label{fig:control}
\end{figure}

%% file: tables/caption-length-fid-comparison.tex

\setlength{\tabcolsep}{0.5pt}
\renewcommand{\arraystretch}{1.0}

\begin{figure}[t]
\centering
\newcommand{\imgw}{0.24\linewidth} 

\begin{tabular}{@{}cccc@{}}
  \multicolumn{4}{@{}l@{}}{{\scriptsize\textbf{Long structured captions training}}} \\[-1pt]
  \includegraphics[width=\imgw]{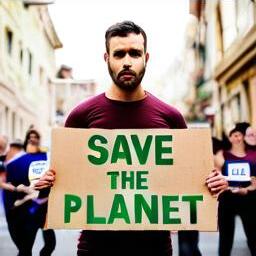} &
  \includegraphics[width=\imgw]{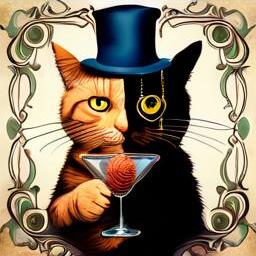} &
  \includegraphics[width=\imgw]{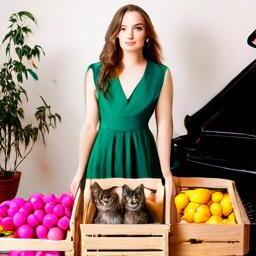} &
  \includegraphics[width=\imgw]{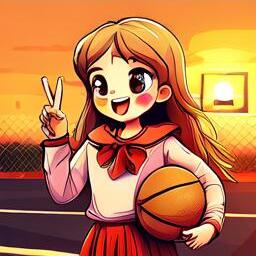} \\[-2pt]
  \multicolumn{4}{@{}l@{}}{{\scriptsize\textbf{Short captions training}}} \\[-1pt]
  \includegraphics[width=\imgw]{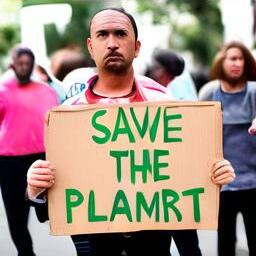} &
  \includegraphics[width=\imgw]{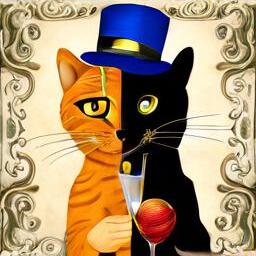} &
  \includegraphics[width=\imgw]{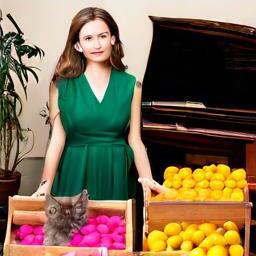} &
  \includegraphics[width=\imgw]{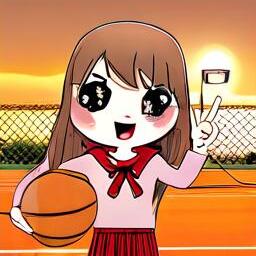} \\
\end{tabular}

\vspace{4pt}

{\setlength{\tabcolsep}{8pt}\renewcommand{\arraystretch}{1.05}
\begin{tabular}{lc}
\toprule
\textbf{Caption Type} & \textbf{FID} ↓ \\
\midrule
Long structured captions & \textbf{19.01} \\
Short captions & 34.04 \\
\bottomrule
\end{tabular}
}

\vspace{-2pt}
\caption{ \textbf{Training with long structured captions vs. short captions.} \emph{Top:} Qualitative comparison ($256X256$ resolution). Training with long captions produces more coherent and visually detailed images, showing faster convergence and better alignment. \emph{Bottom:} Quantitative results on $30$K COCO-2014 val images~\cite{lin2014microsoft}: long captions yield lower (better) FID.}
\label{fig:caption-length-merged}
\end{figure}

%% file: Figures/disentanglement/dis_fig.tex
\setlength{\tabcolsep}{0.5pt}
\renewcommand{\arraystretch}{1.0}

\begin{figure}[t]
\centering
\newcommand{\imgw}{0.24\linewidth} 
\begin{tabular}{@{}cccc@{}}

  \includegraphics[width=\imgw]{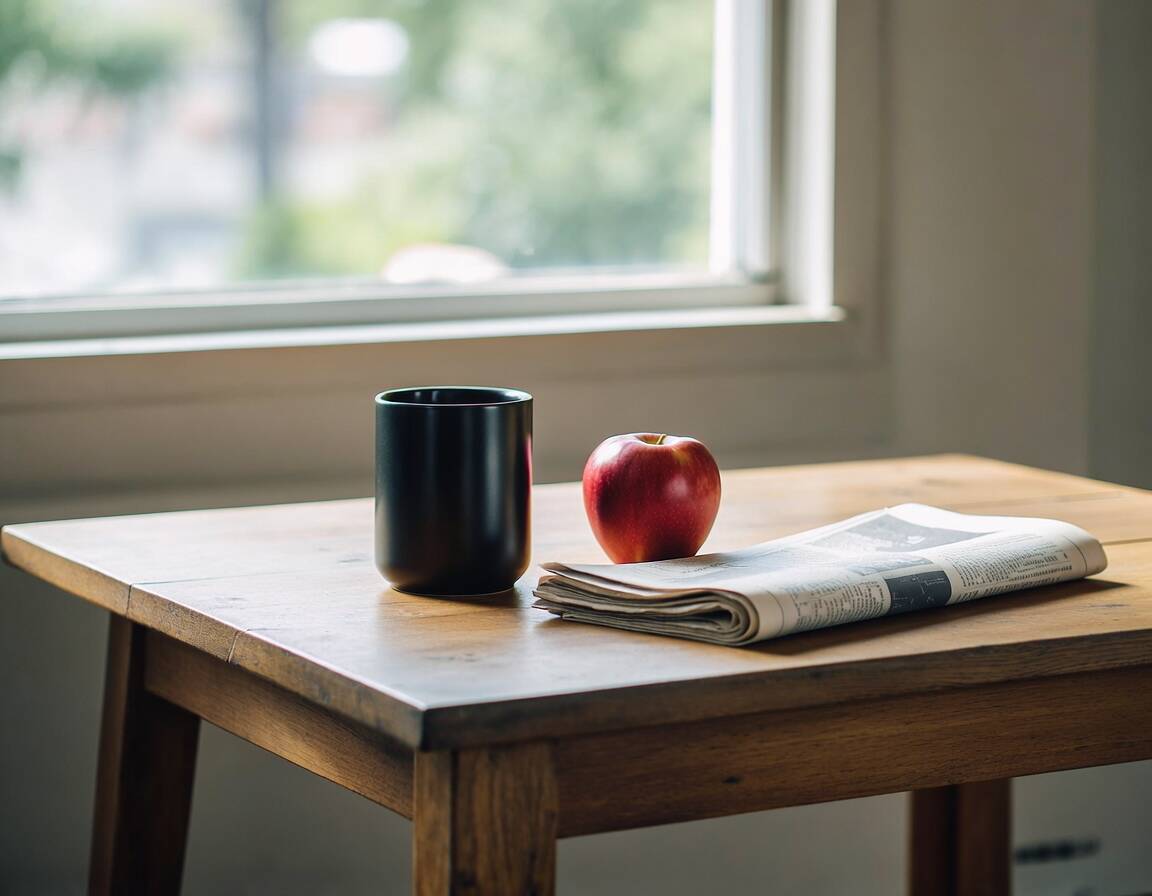} &
  \includegraphics[width=\imgw]{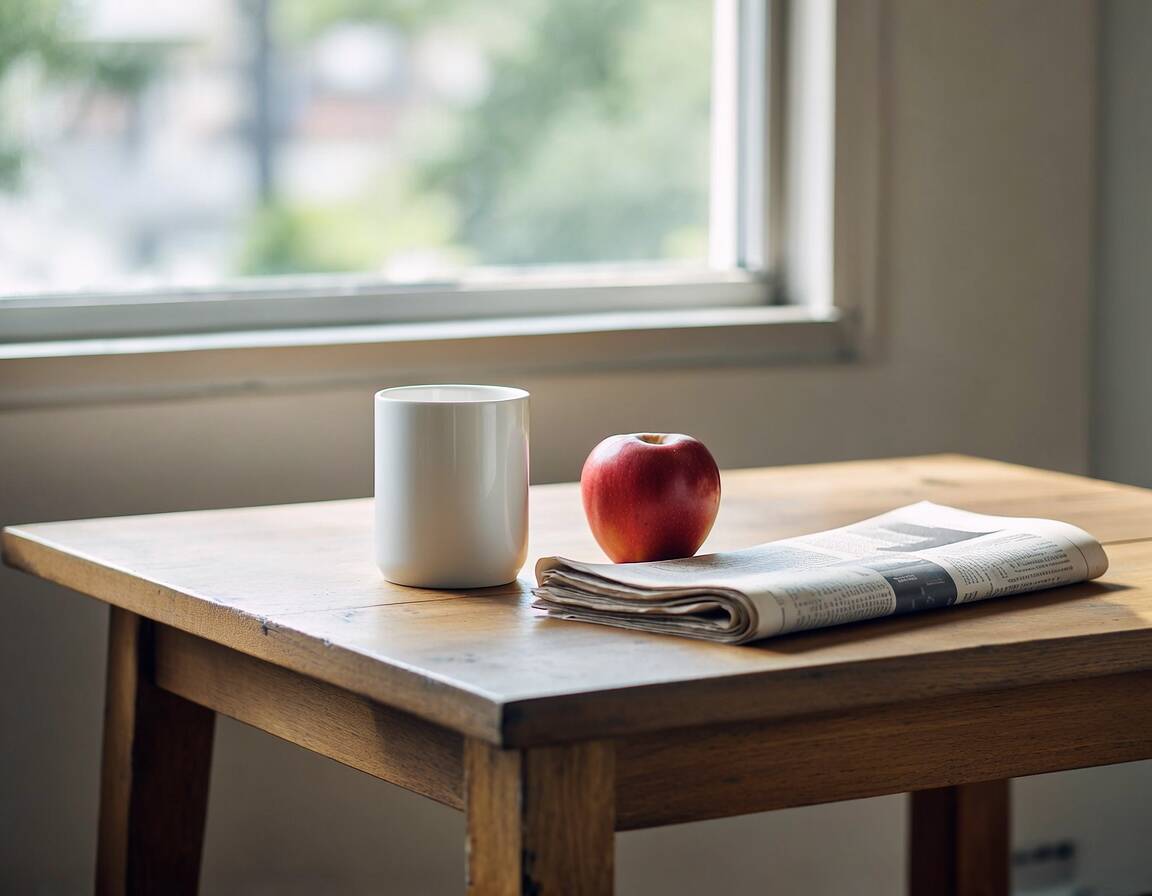} &
  \includegraphics[width=\imgw]{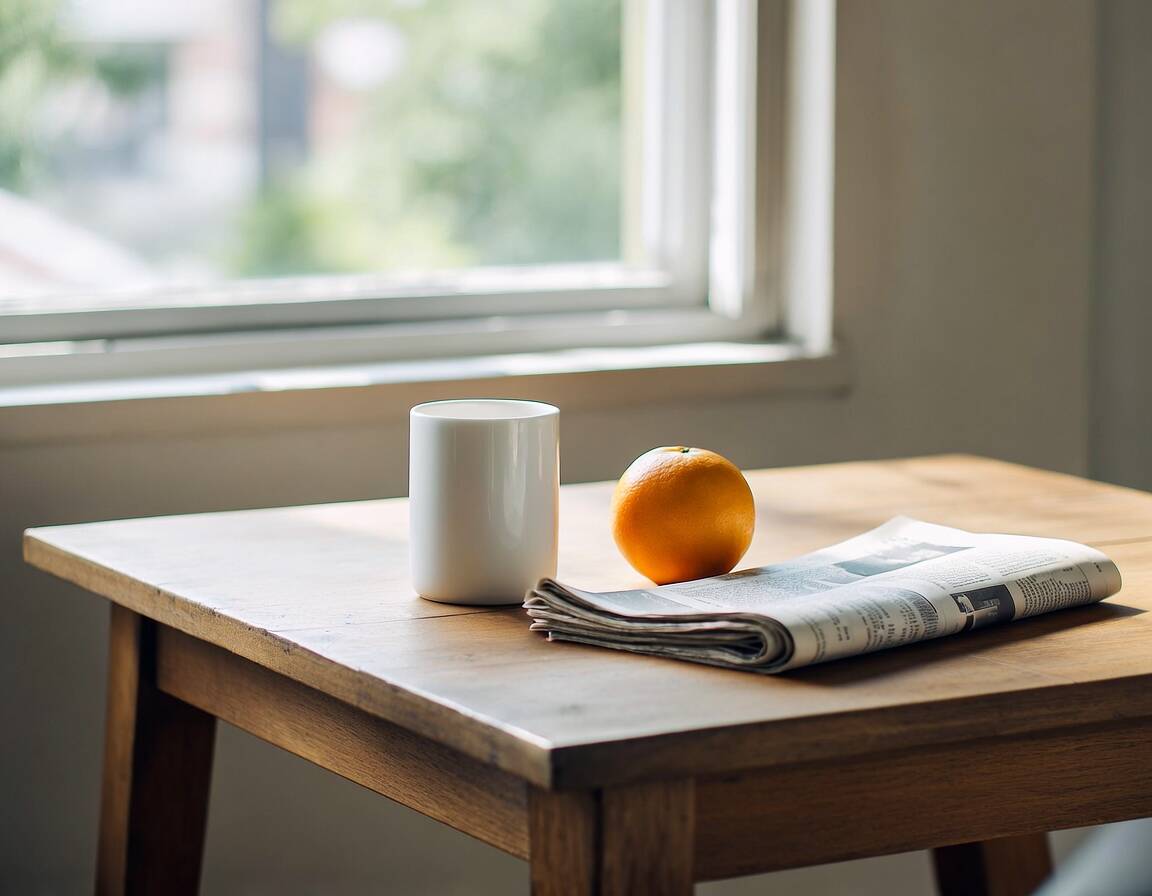} &
  \includegraphics[width=\imgw]{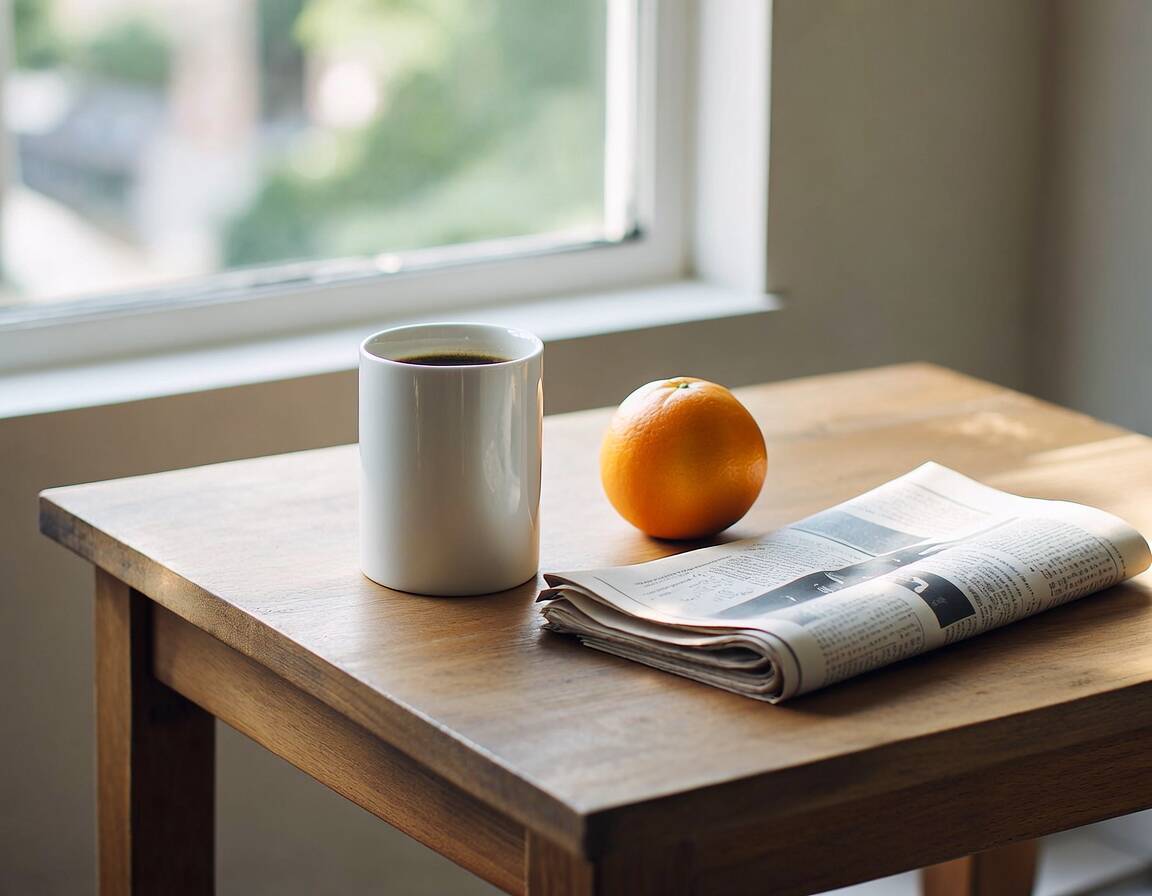}
  \\[-2pt]

  \parbox{\imgw}{\scriptsize\centering \textbf{}} &
  \parbox{\imgw}{\scriptsize\centering \textbf{black mug to white}} &
  \parbox{\imgw}{\scriptsize\centering \textbf{apple to orange}} &
  \parbox{\imgw}{\scriptsize\centering \textbf{add coffee to mug}}
  \\[+2pt]

  \includegraphics[width=\imgw]{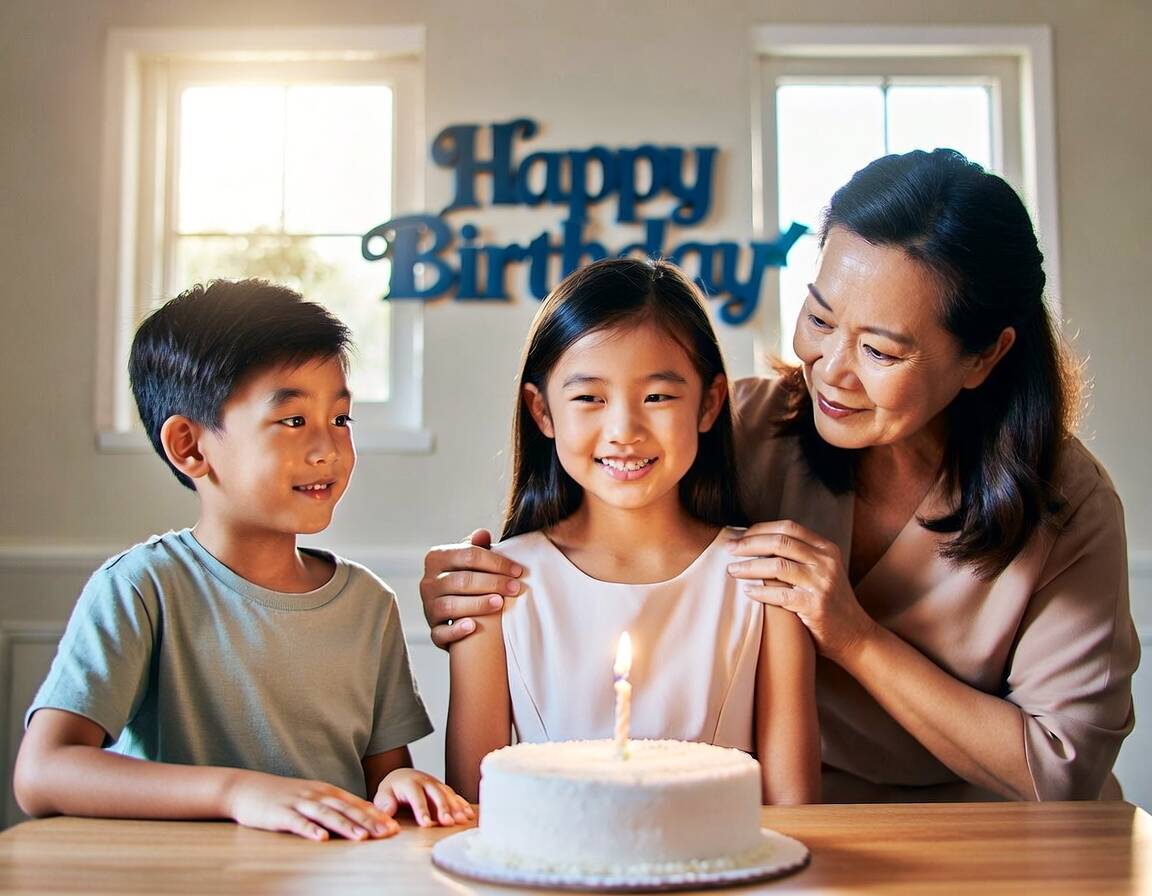} &
  \includegraphics[width=\imgw]{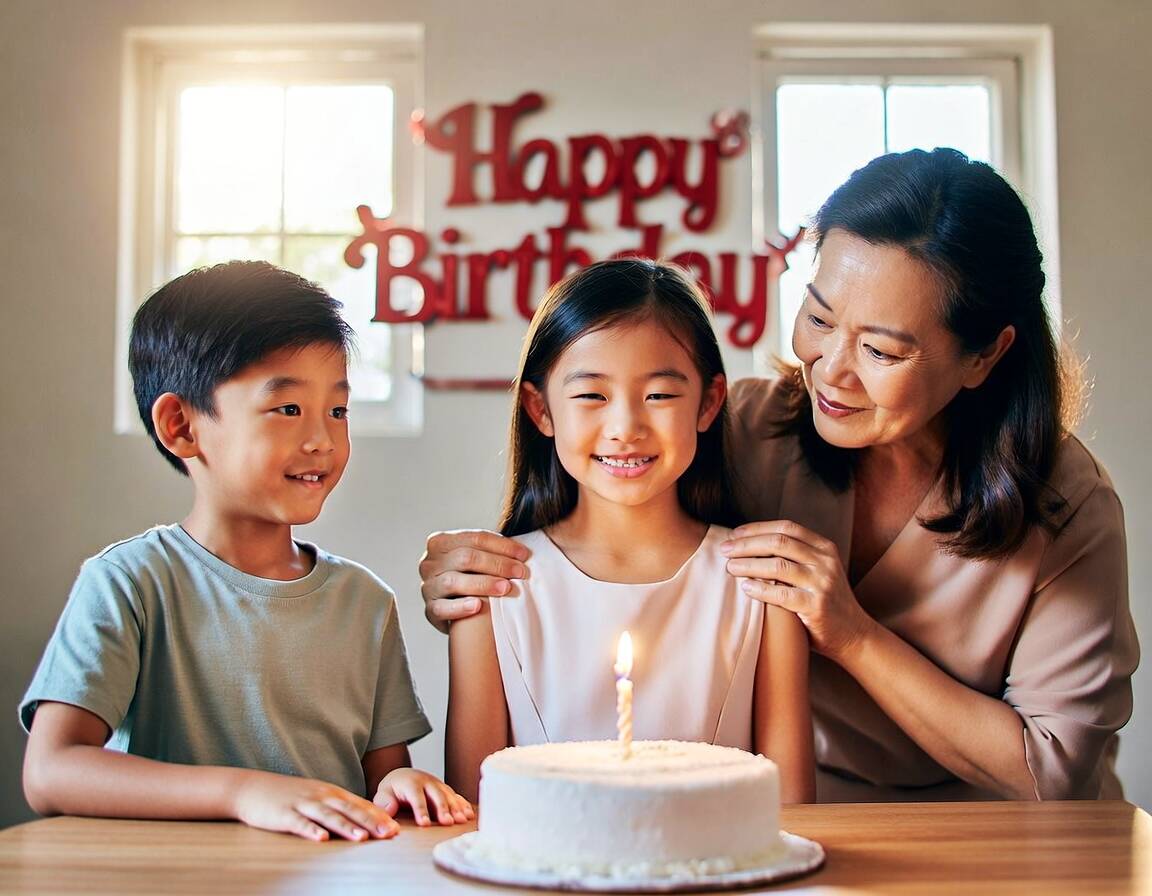} &
  \includegraphics[width=\imgw]{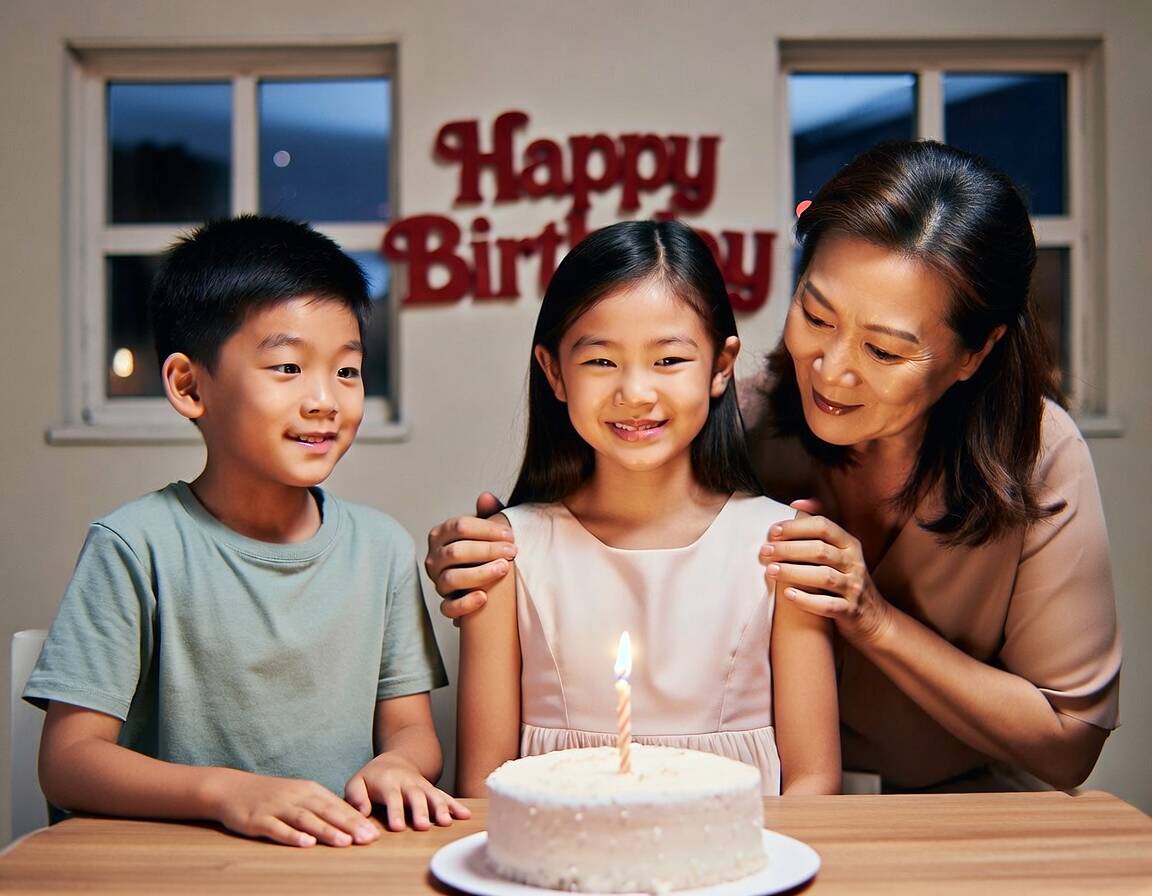} &
  \includegraphics[width=\imgw]{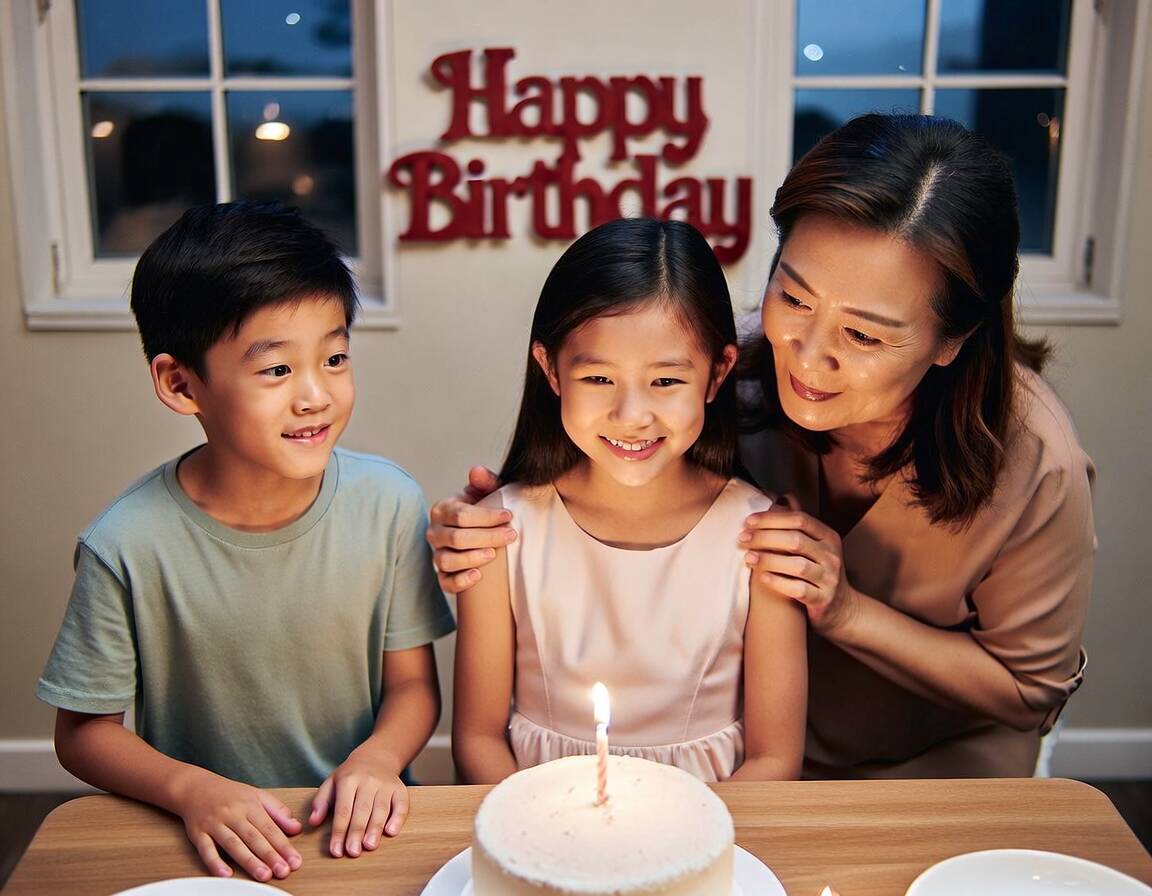}
  \\[-2pt]

  \parbox{\imgw}{\scriptsize\centering \textbf{}} &
  \parbox{\imgw}{\scriptsize\centering \textbf{blue sign to red}} &
  \parbox{\imgw}{\scriptsize\centering \textbf{make it evening}} &
  \parbox{\imgw}{\scriptsize\centering \textbf{set camera from above}}
  \\[+2pt]

  \includegraphics[width=\imgw]{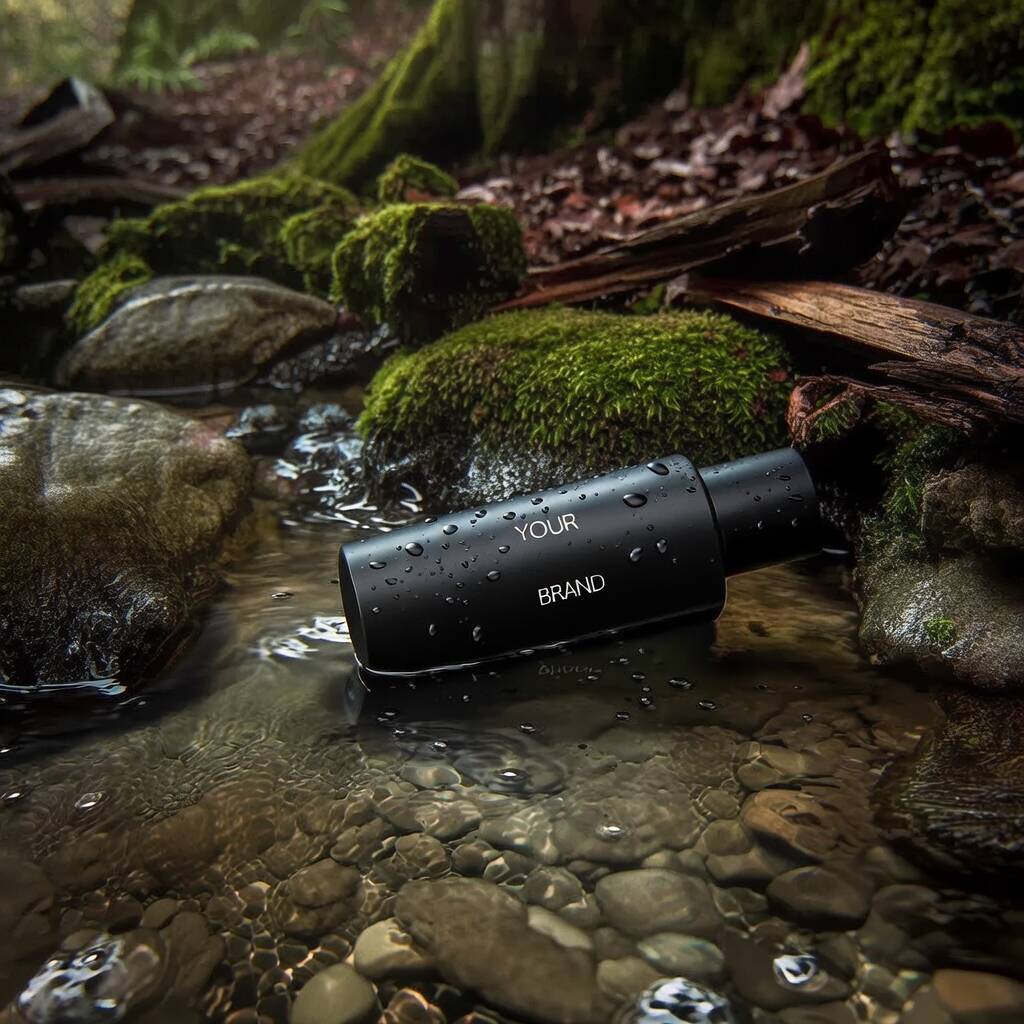} &
  \includegraphics[width=\imgw]{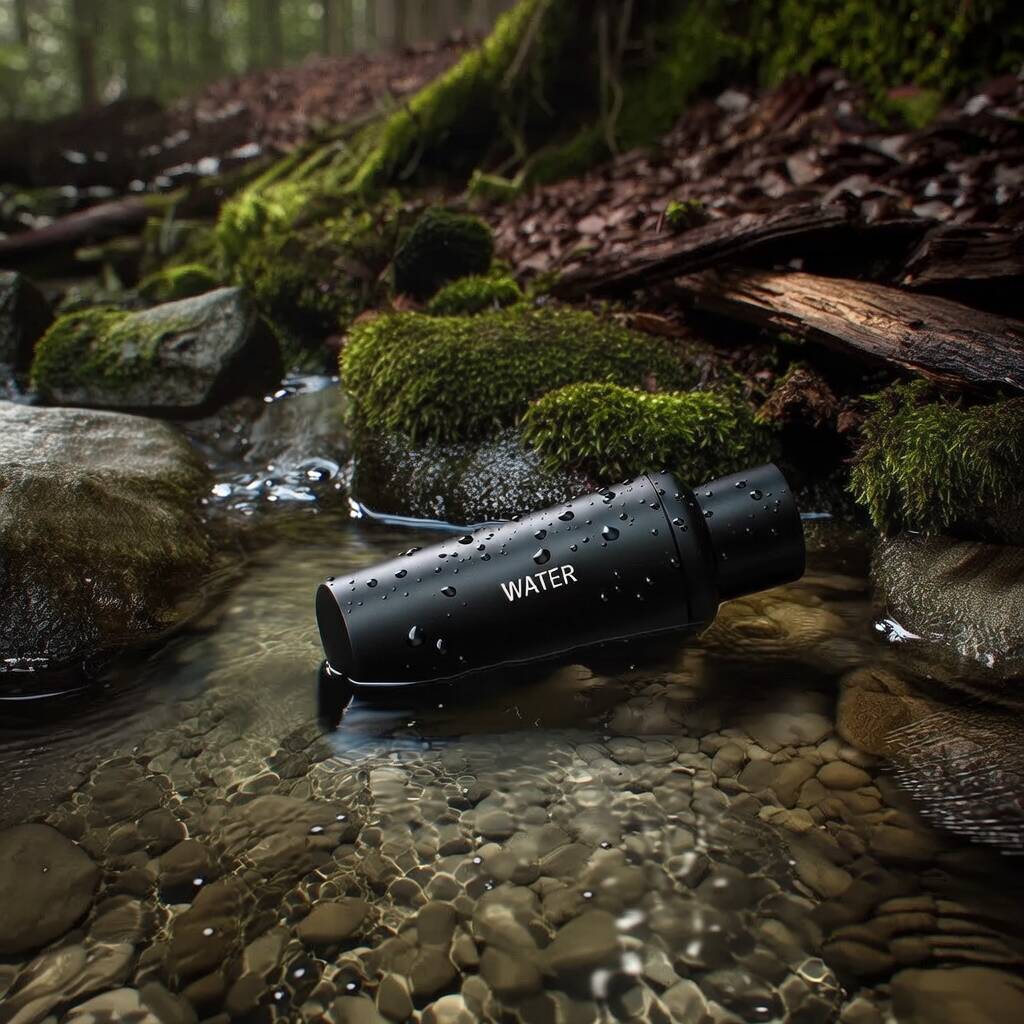} &
  \includegraphics[width=\imgw]{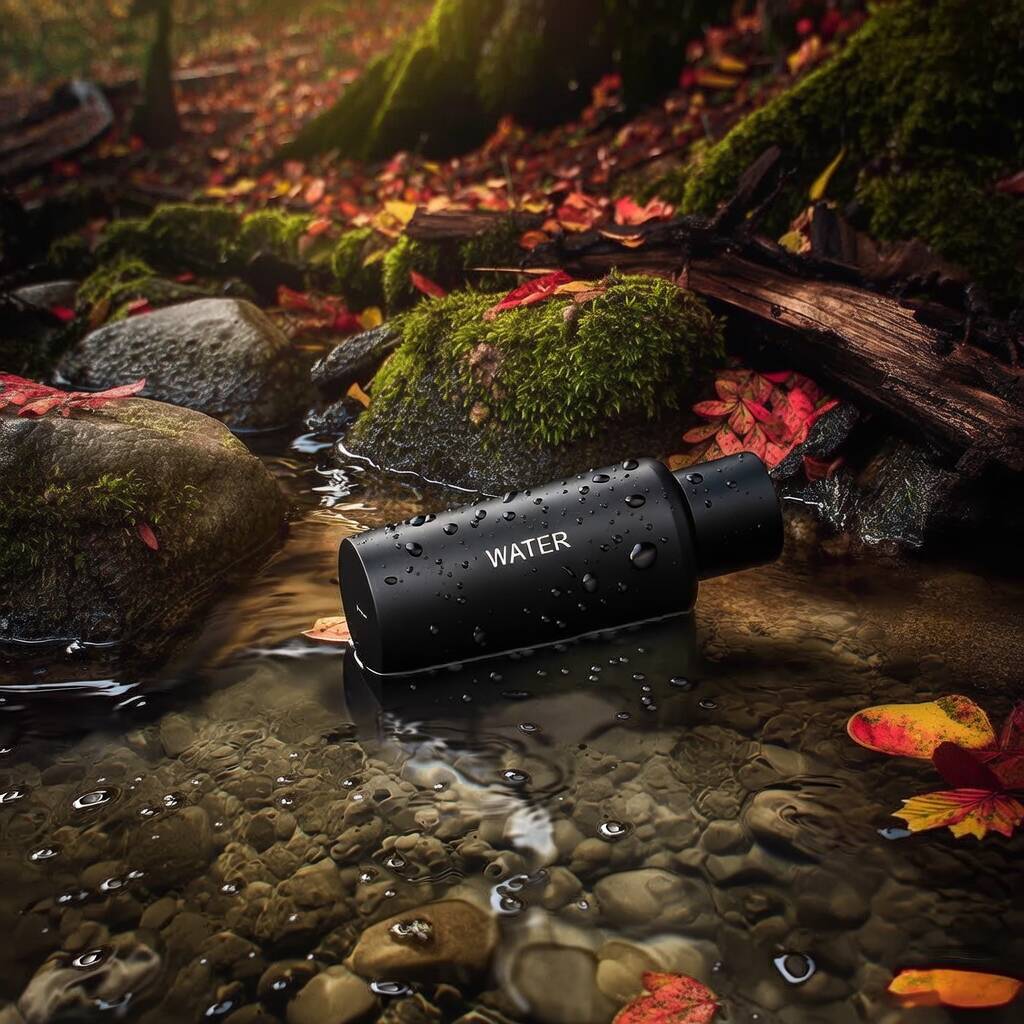} &
  \includegraphics[width=\imgw]{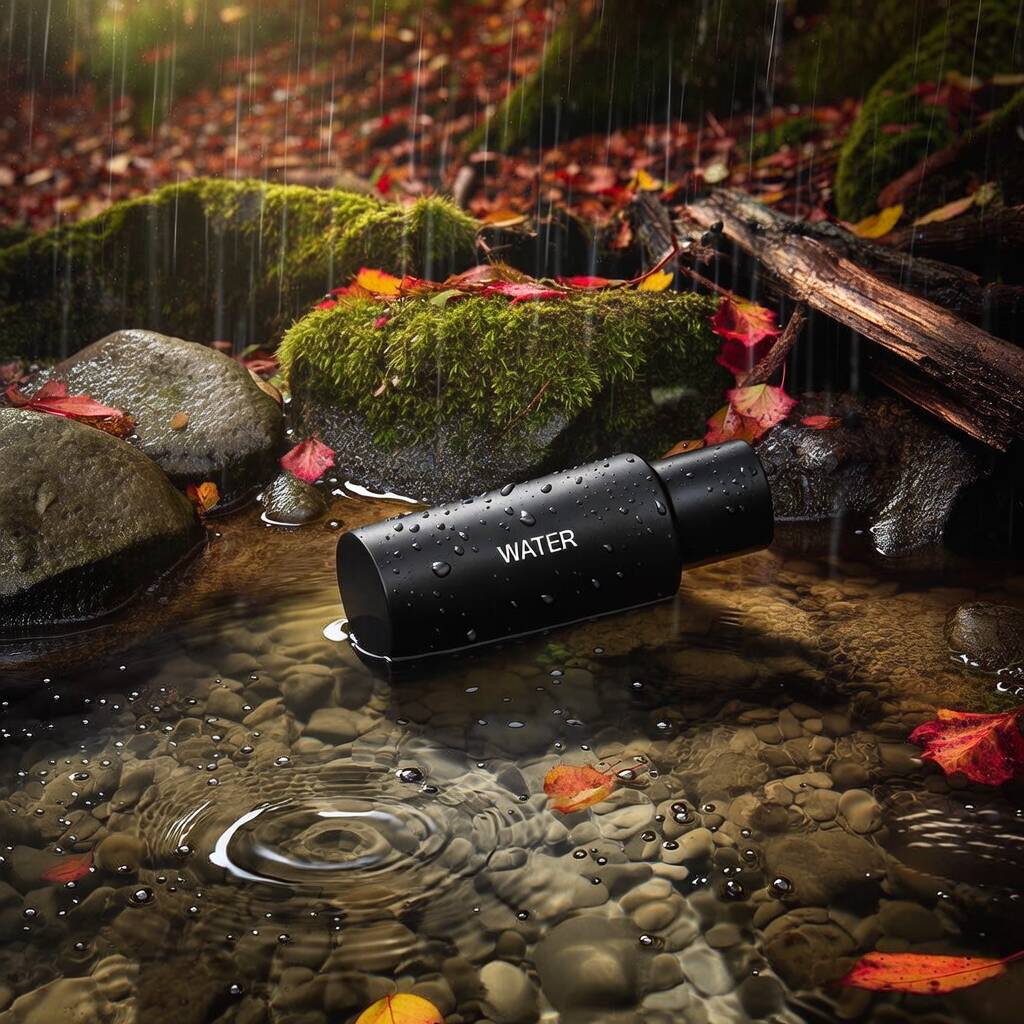}
  \\[-2pt]

  \parbox{\imgw}{\scriptsize\centering \textbf{}} &
  \parbox{\imgw}{\scriptsize\centering \textbf{change text to ``water''}} &
  \parbox{\imgw}{\scriptsize\centering \textbf{season is autumn}} &
  \parbox{\imgw}{\scriptsize\centering \textbf{add rain drops}}
  \\[+2pt]



  \includegraphics[width=\imgw]{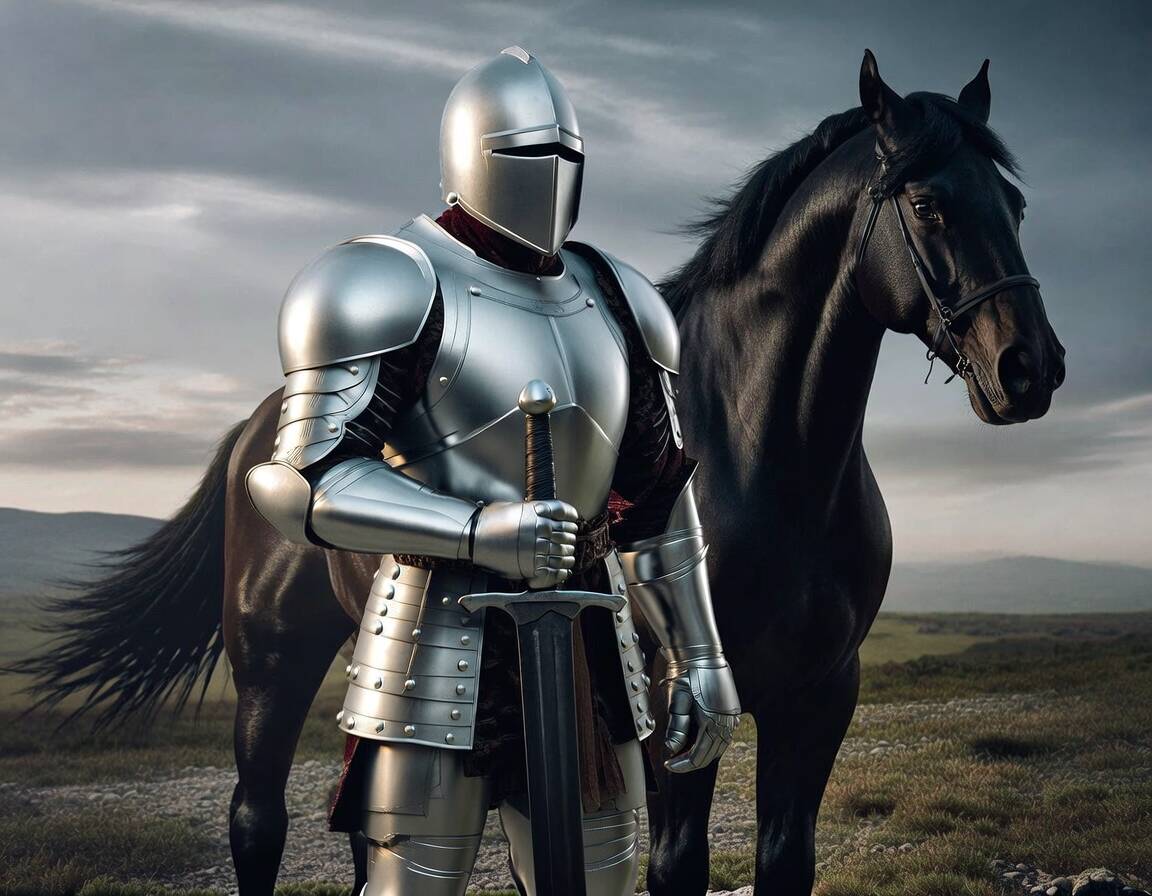} &
  \includegraphics[width=\imgw]{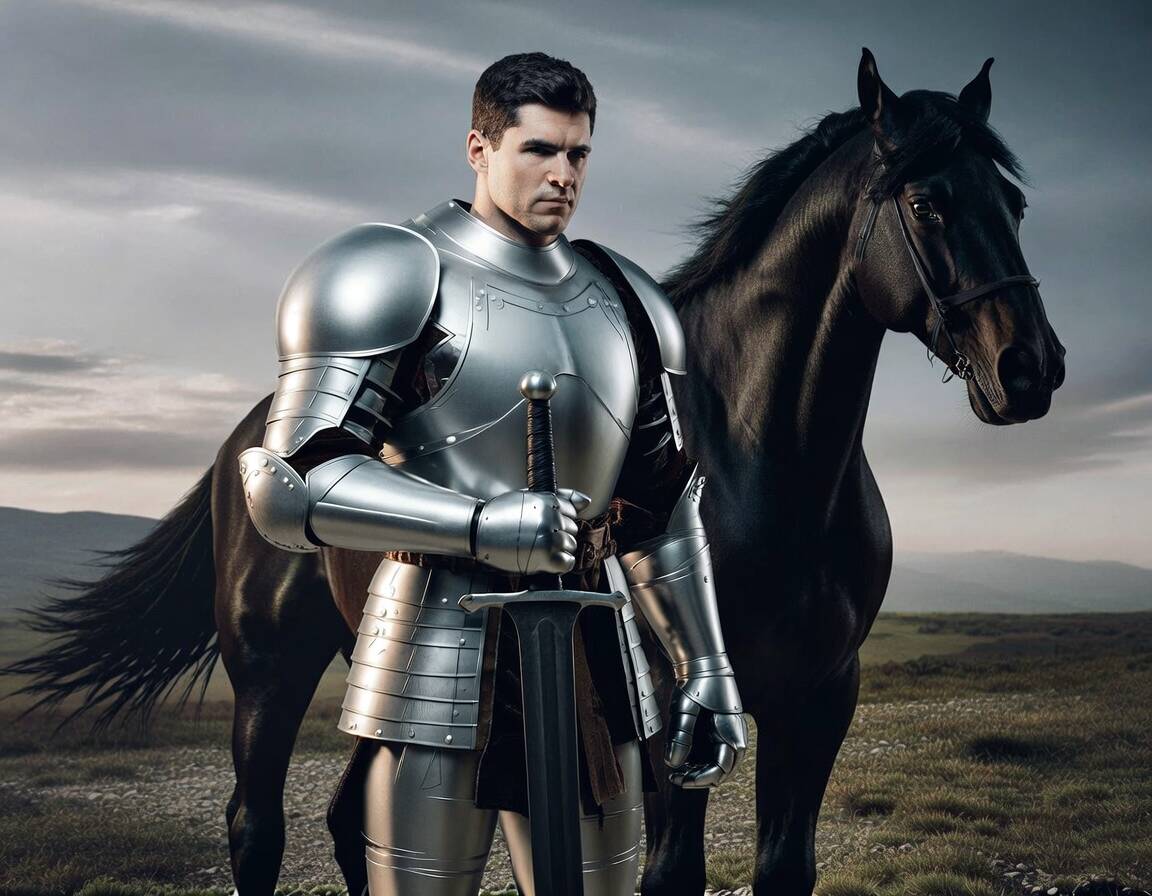} &
  \includegraphics[width=\imgw]{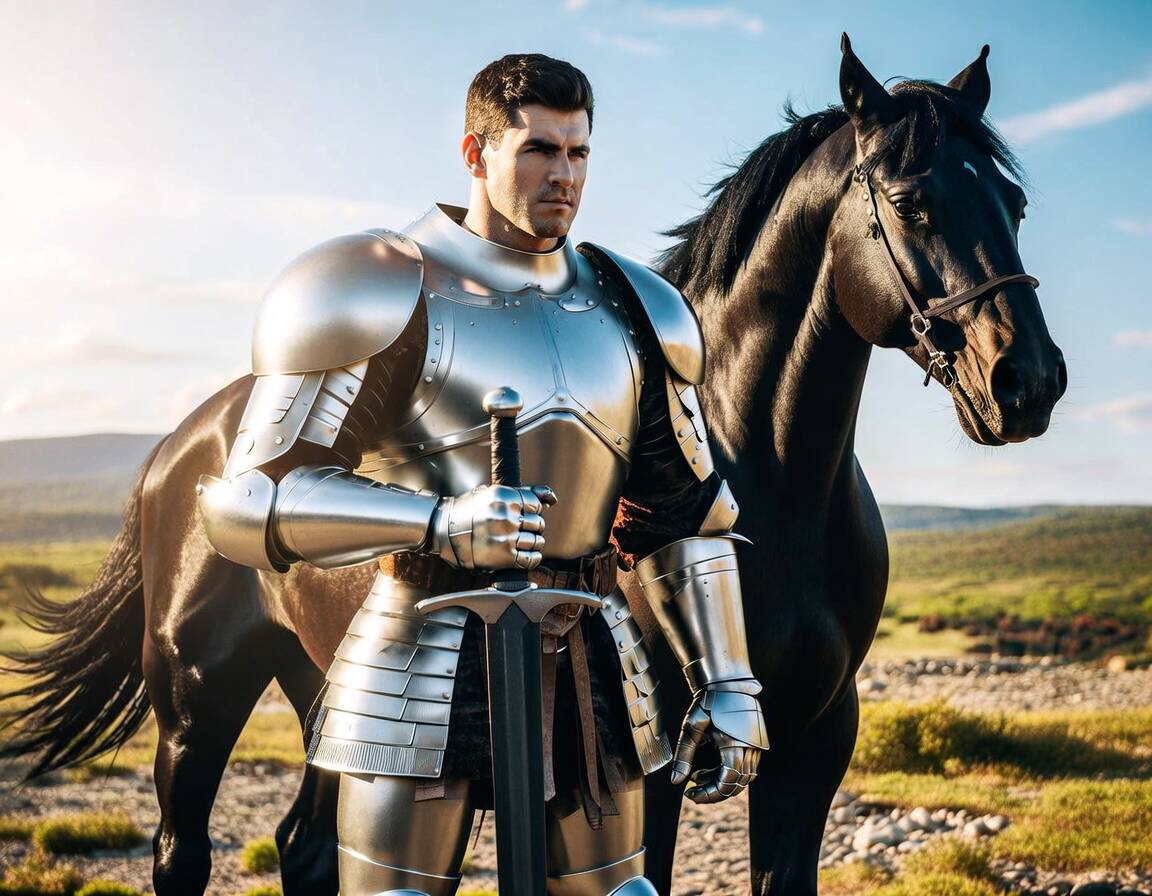} &
  \includegraphics[width=\imgw]{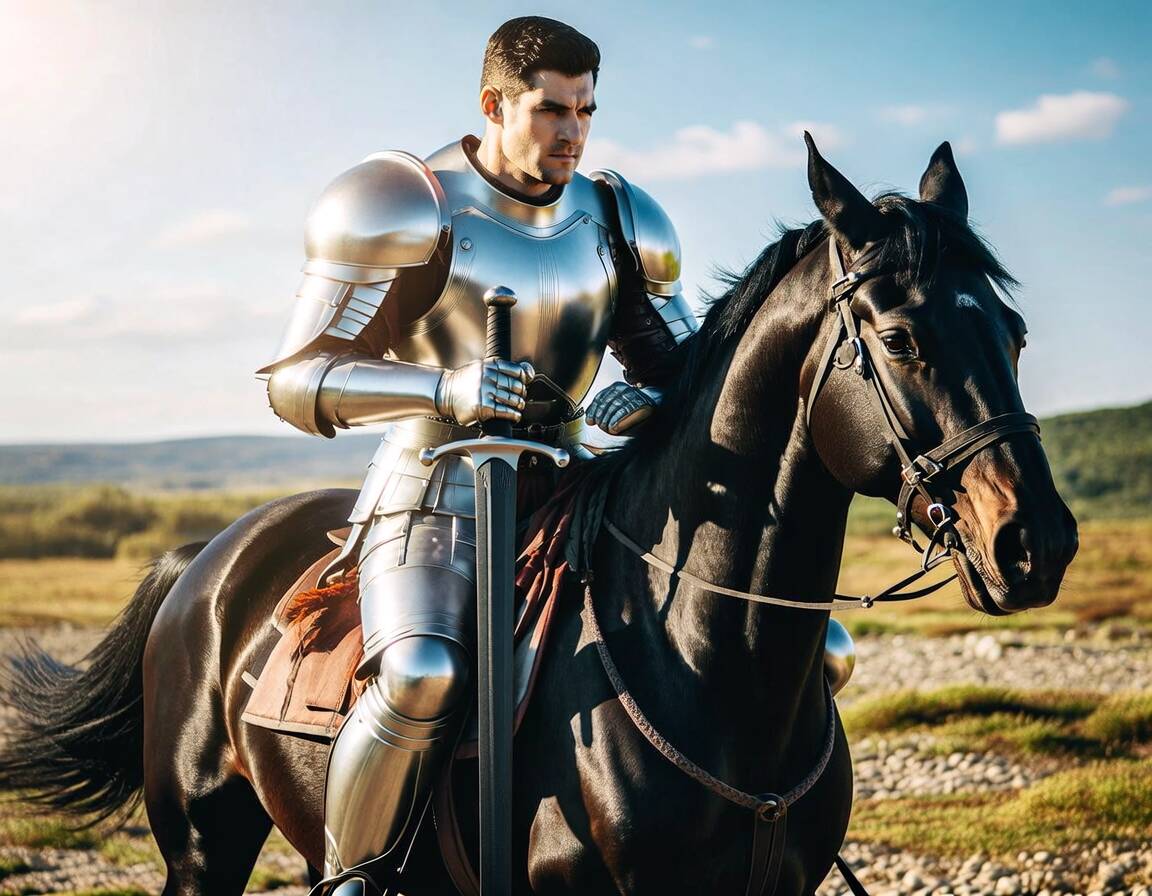}
  \\[-2pt]

  \parbox{\imgw}{\scriptsize\centering \textbf{}} &
  \parbox{\imgw}{\scriptsize\centering \textbf{remove helmet}} &
  \parbox{\imgw}{\scriptsize\centering \textbf{make sunny}} &
  \parbox{\imgw}{\scriptsize\centering \textbf{knight riding horse}}
  \\[+2pt]

\end{tabular}

\vspace{-3pt}
\caption{\textbf{Structured captions encourage disentanglement.} Our model allows iterative refinement: altering one attribute in the JSON typically affects only the corresponding visual factor. Starting from the left images, we demonstrate iteratively editing, where the label below describes the requested change in the JSON.}
\label{fig:disentanglement}
\end{figure}

%% file: tables/captions_stats.tex
\begin{table}[t]
\centering
{\setlength{\tabcolsep}{4pt} 
\begin{tabular}{cccccc}
\toprule
& \textbf{Avg.} & \textbf{Median} & \textbf{Std. Dev.} & \textbf{Min} & \textbf{Max} \\
\midrule
$\#$\textbf{Tokens} & 1160.3 & 1176 & 316.4 & 192 & 1800 \\
\bottomrule
\end{tabular}
}
\caption{\textbf{Structured captions statistics.}}
\label{tab:caption_stats}
\end{table}

%% file: Figures/dimfusion_arch.tex
\begin{figure}[t]
  \centering
  \includegraphics[width=\linewidth]{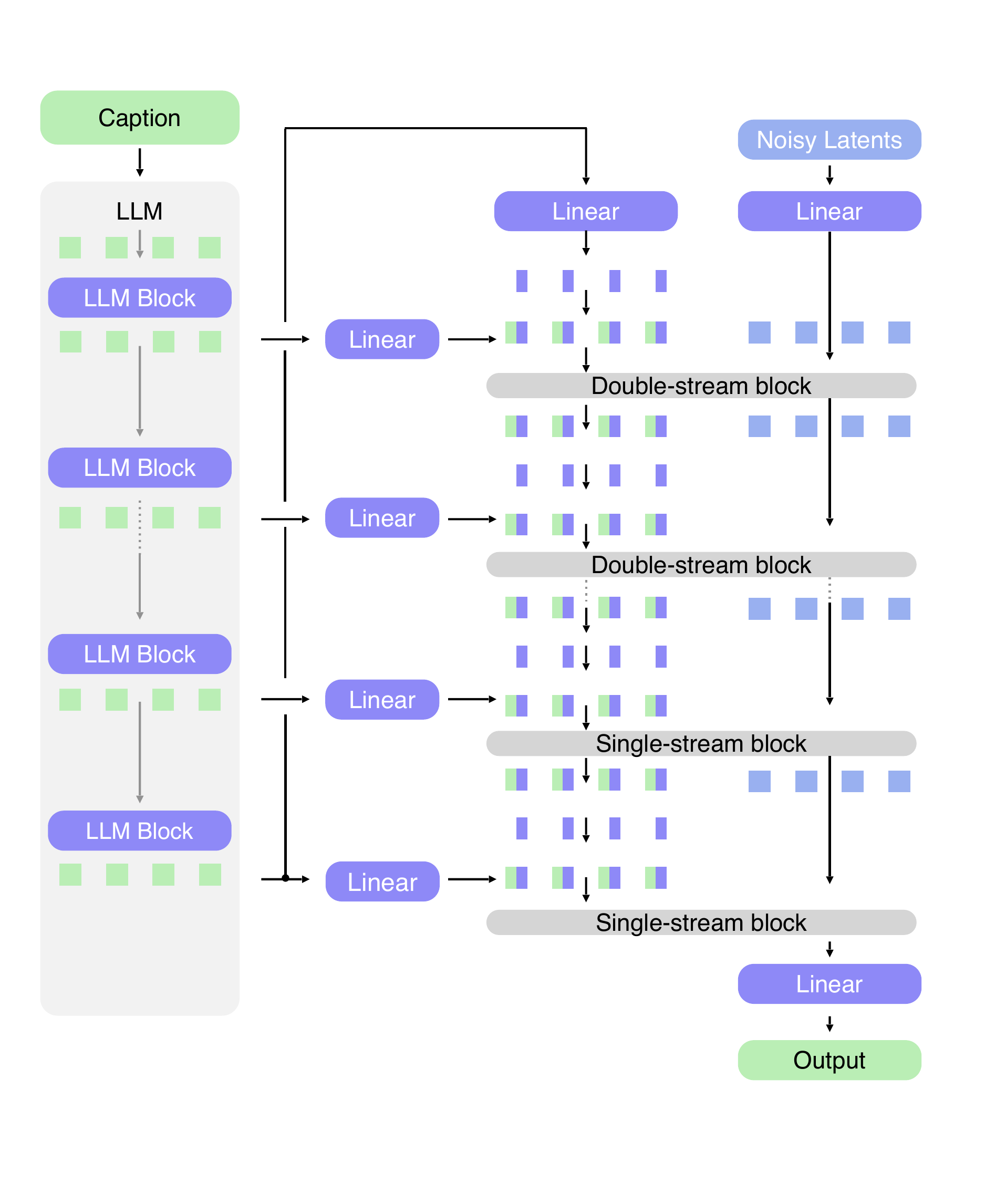}
  \caption{\textbf{DimFusion architecture.} In each layer, the \textbf{\textcolor{textencoding}{text encoding}} is concatenated with the corresponding LLM \textbf{\textcolor{llmtokens}{hidden states}} along the embedding dimension. The resulting representation is then jointly processed with the \textbf{\textcolor{imagetokens}{noisy latents}} through bi-directional mixing in the Dual- and Single-stream blocks. After each block, the appended LLM hidden states are discarded, restoring the original text embedding dimension.}
  \label{fig:arch}
\end{figure}

%% file: Figures/dimfusion_ablation.tex



\begin{figure}[t]
\centering

{\setlength{\tabcolsep}{8pt}\renewcommand{\arraystretch}{1.05}
\begin{tabular}{lcc}
\toprule
\textbf{Architecture} & \textbf{FID} ↓ & \textbf{Avg. Time per Step (sec)} ↓\\
\midrule
T5 & 36.46  & 1.28\\
TokenFusion & 15.90 & 0.8\\
DimFusion & \textbf{15.58} & \textbf{0.5}\\
\bottomrule
\end{tabular}
}

\vspace{6pt}

\resizebox{\linewidth}{!}{%
  \includegraphics[width=0.7\textwidth]{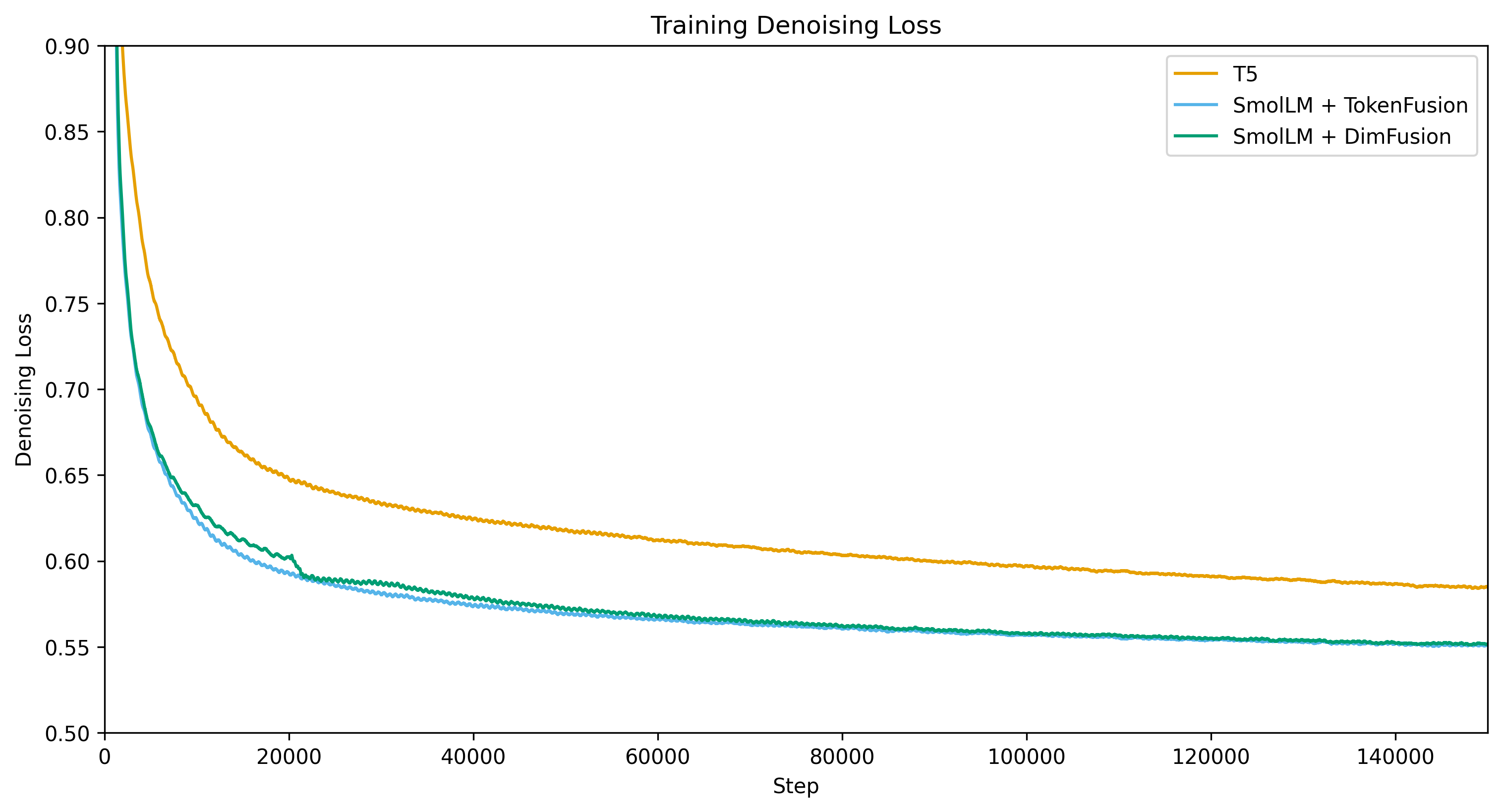}
}
\resizebox{\linewidth}{!}{%
  \includegraphics[width=0.7\textwidth]{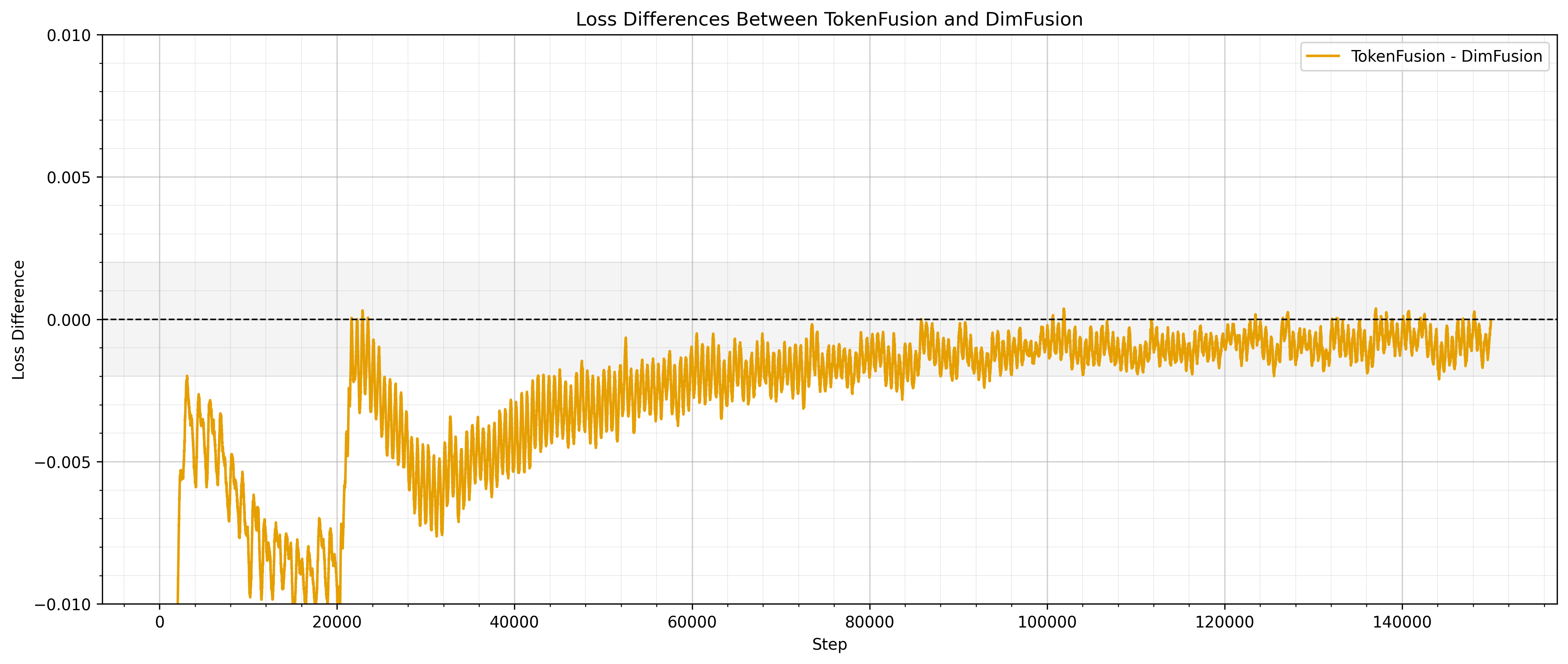}
}

\vspace{-2pt}
\caption{\textbf{DimFusion vs. TokenFusion} \emph{Top:} FID on $30$K COCO-2014 val images~\cite{lin2014microsoft}: both fusion strategies (TokenFusion, DimFusion) substantially outperform T5, with \textbf{DimFusion} achieving the best FID while maintaining lower time per step. \emph{Middle:} Training loss comparison for T5 (baseline), SmolLM3-3B + TokenFusion, and SmolLM3-3B + DimFusion. Both fusion strategies outperform T5, with DimFusion closely matching TokenFusion. \emph{Bottom:} Loss difference relative to TokenFusion zoomed in, showing DimFusion converges to a similar value while requiring significantly lower computational cost.}
\label{fig:dimfusion-merged}
\end{figure}

%% file: Figures/tabr/tabr.tex
\setlength{\tabcolsep}{0.5pt}
\renewcommand{\arraystretch}{1.0}

\begin{figure}[t]
\centering
\newcommand{\imgw}{0.24\linewidth} 
\begin{tabular}{@{}cccc@{}}

  \includegraphics[width=\imgw]{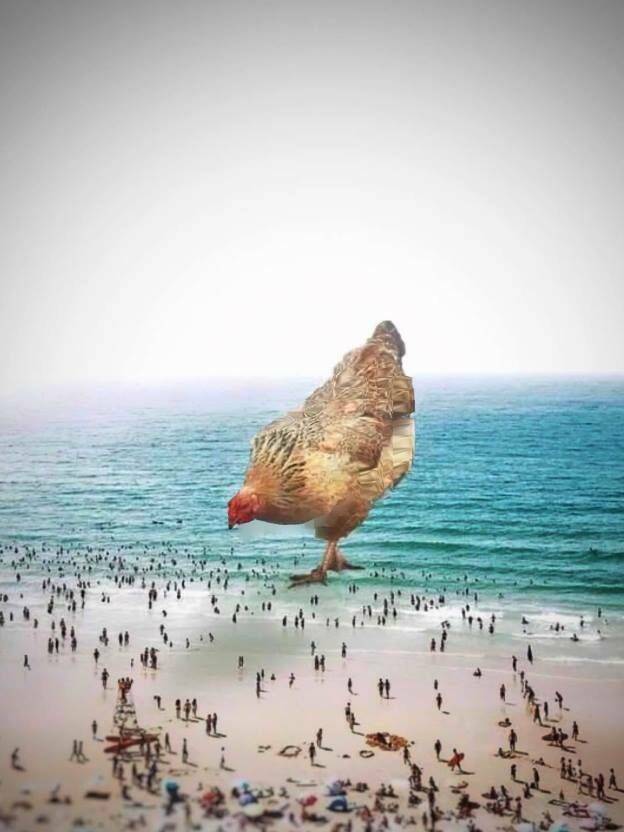} &
  \includegraphics[width=\imgw]{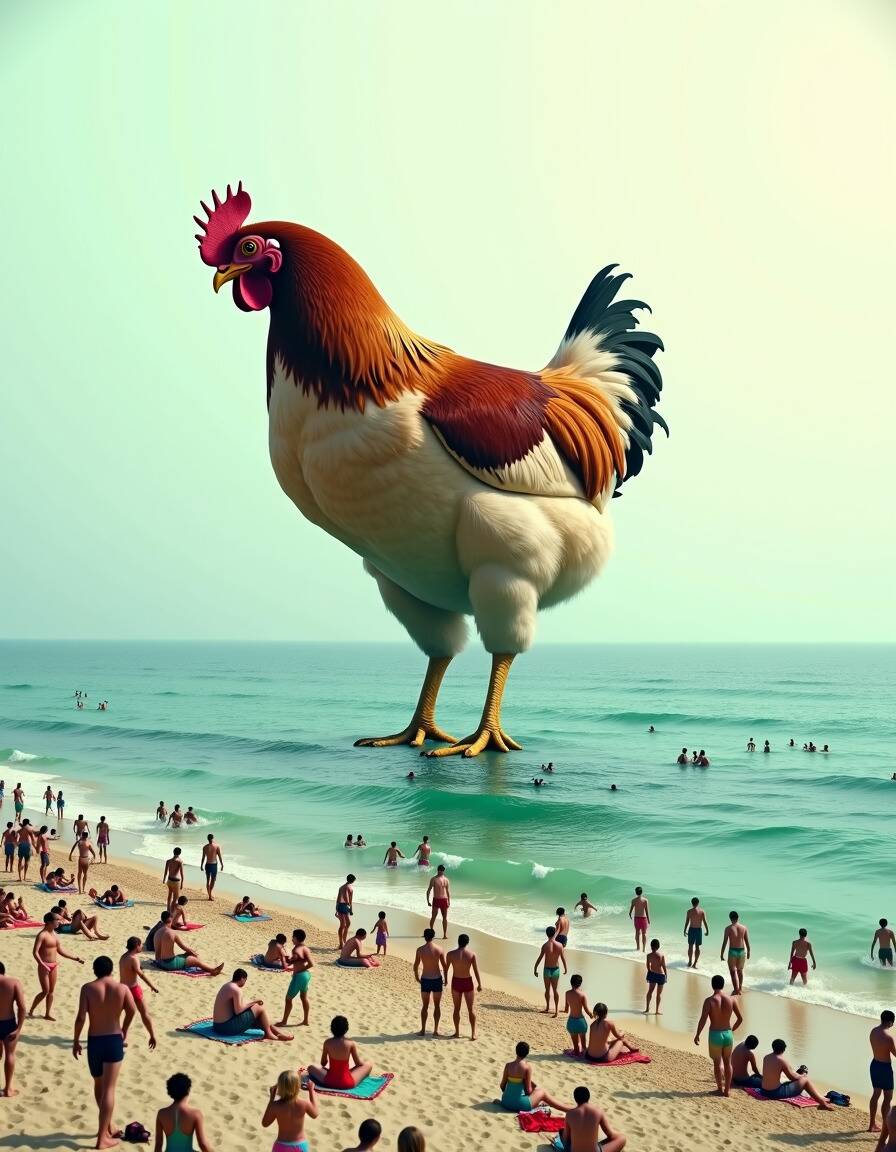} &
  \includegraphics[width=\imgw]{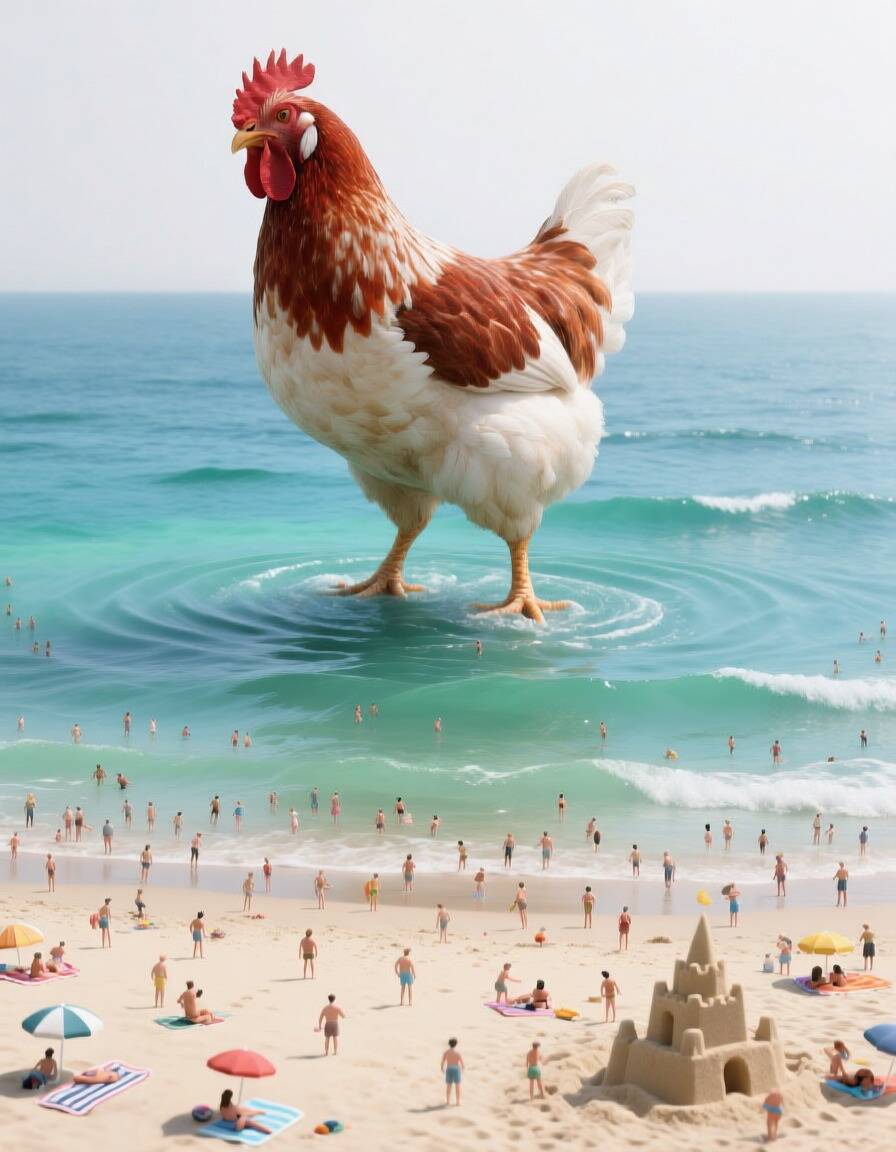} &
  \includegraphics[width=\imgw]{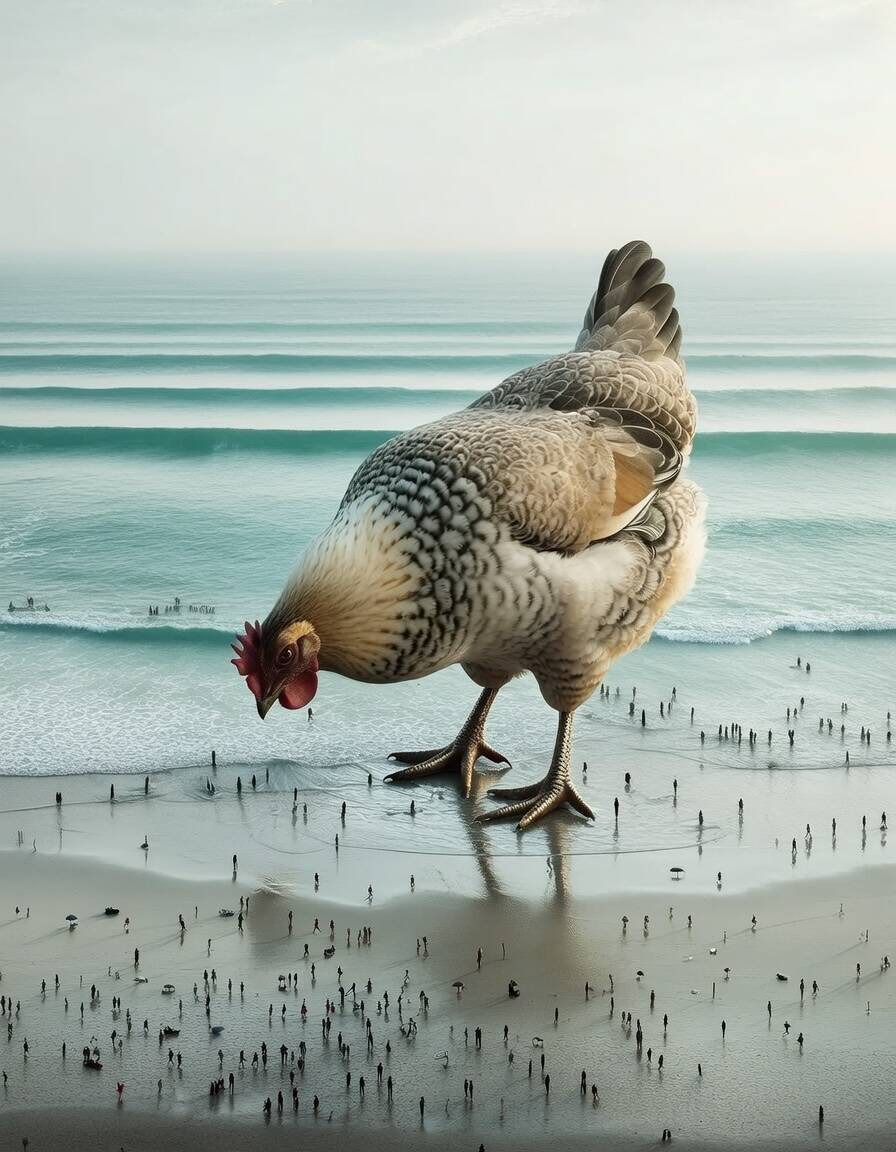}
  \\[-2pt]

  \includegraphics[width=\imgw]{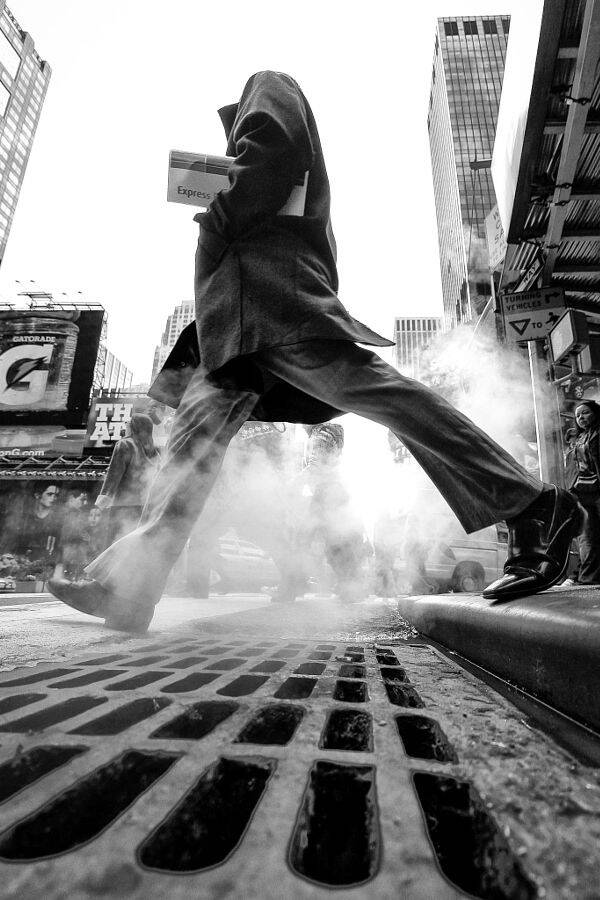} &
  \includegraphics[width=\imgw]{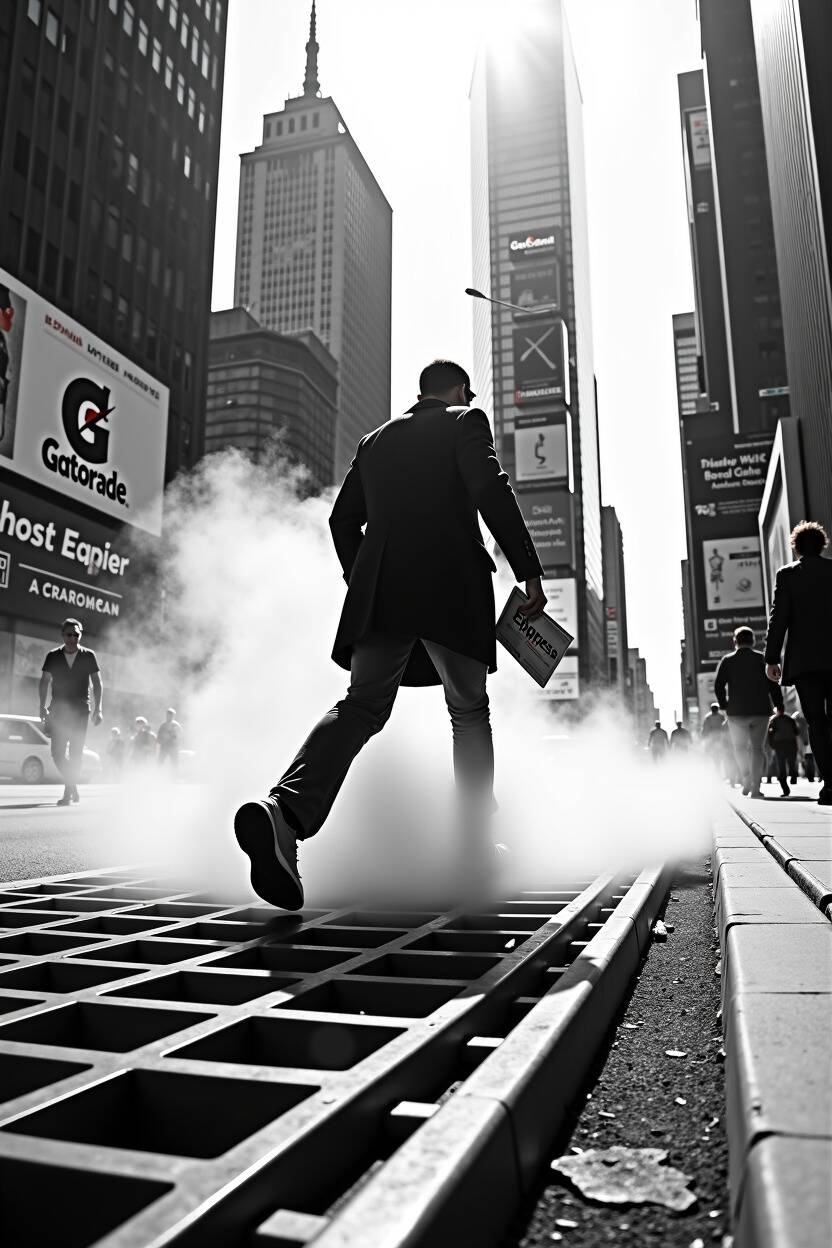} &
  \includegraphics[width=\imgw]{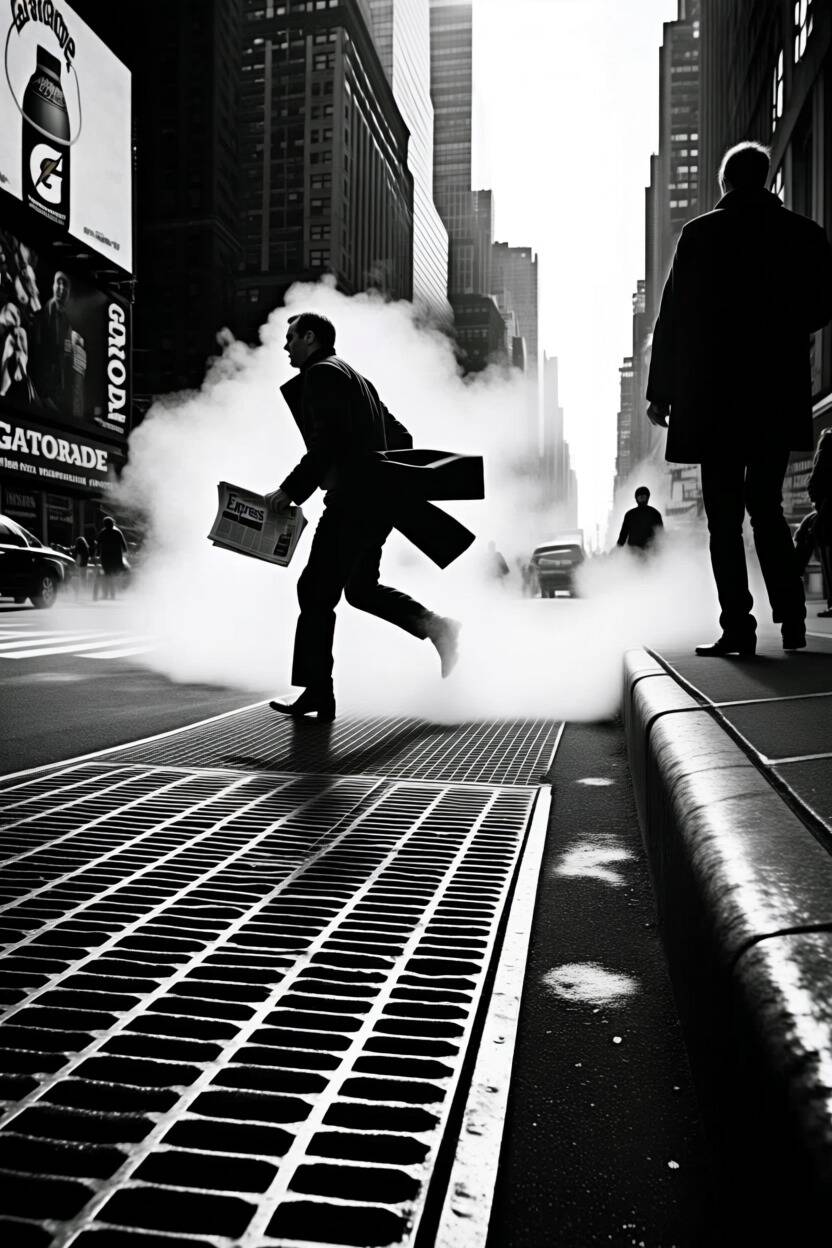} &
  \includegraphics[width=\imgw]{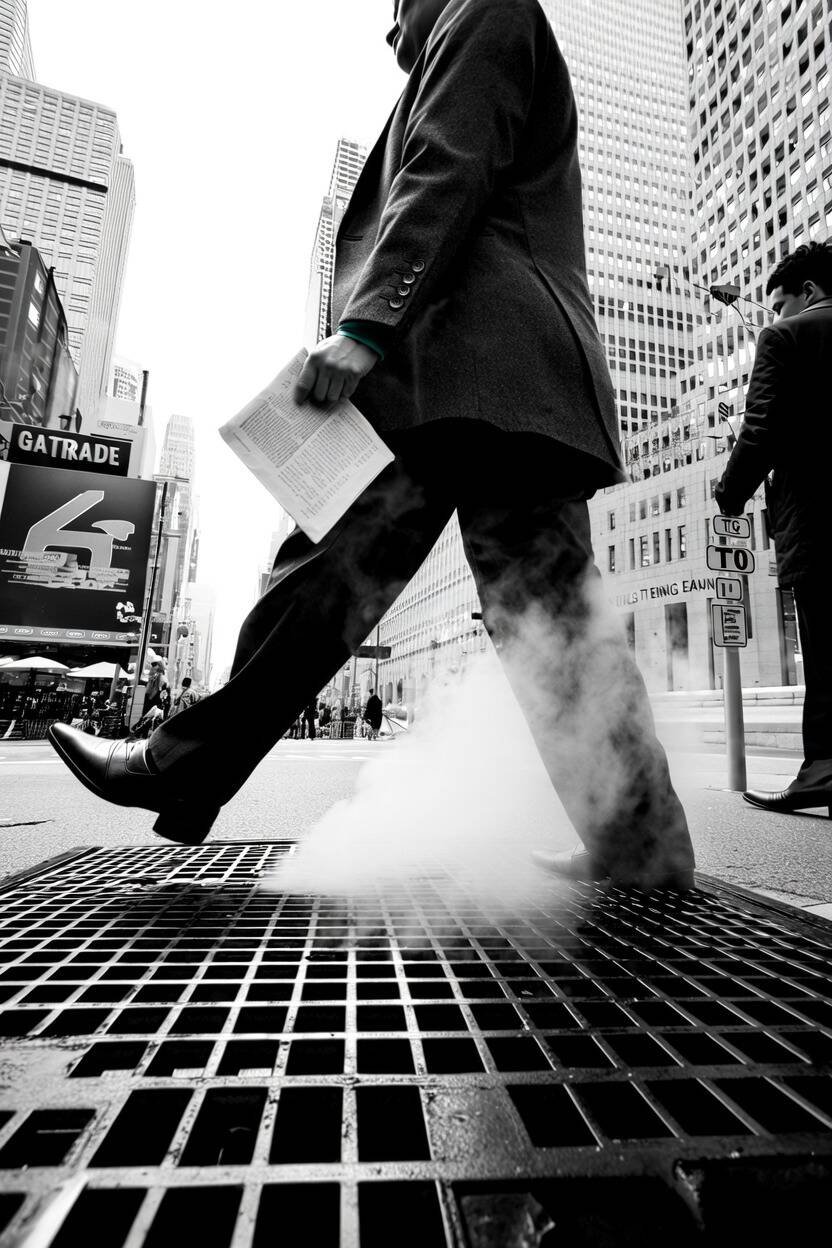}
  \\

  \includegraphics[width=\imgw]{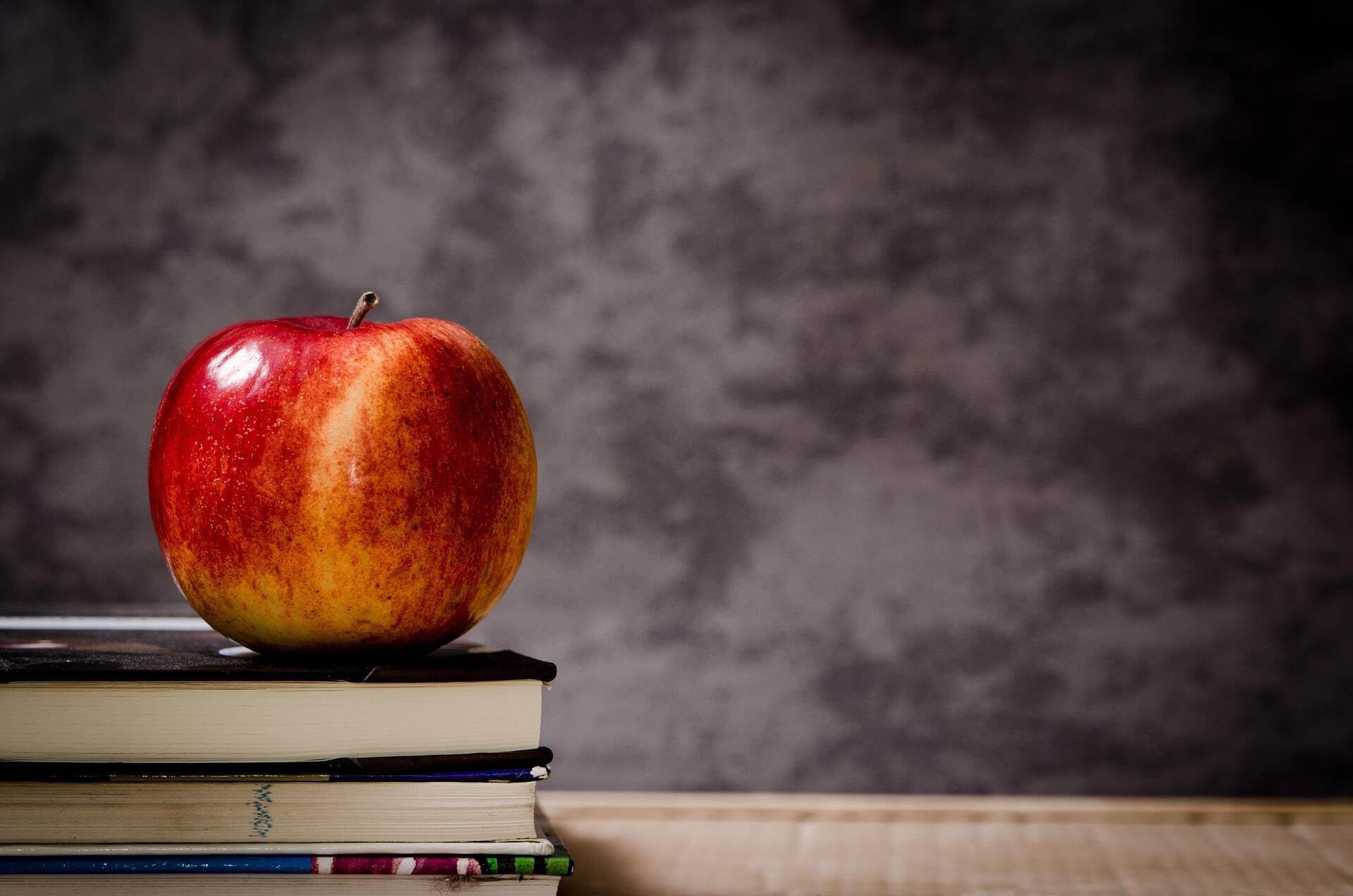} &
  \includegraphics[width=\imgw]{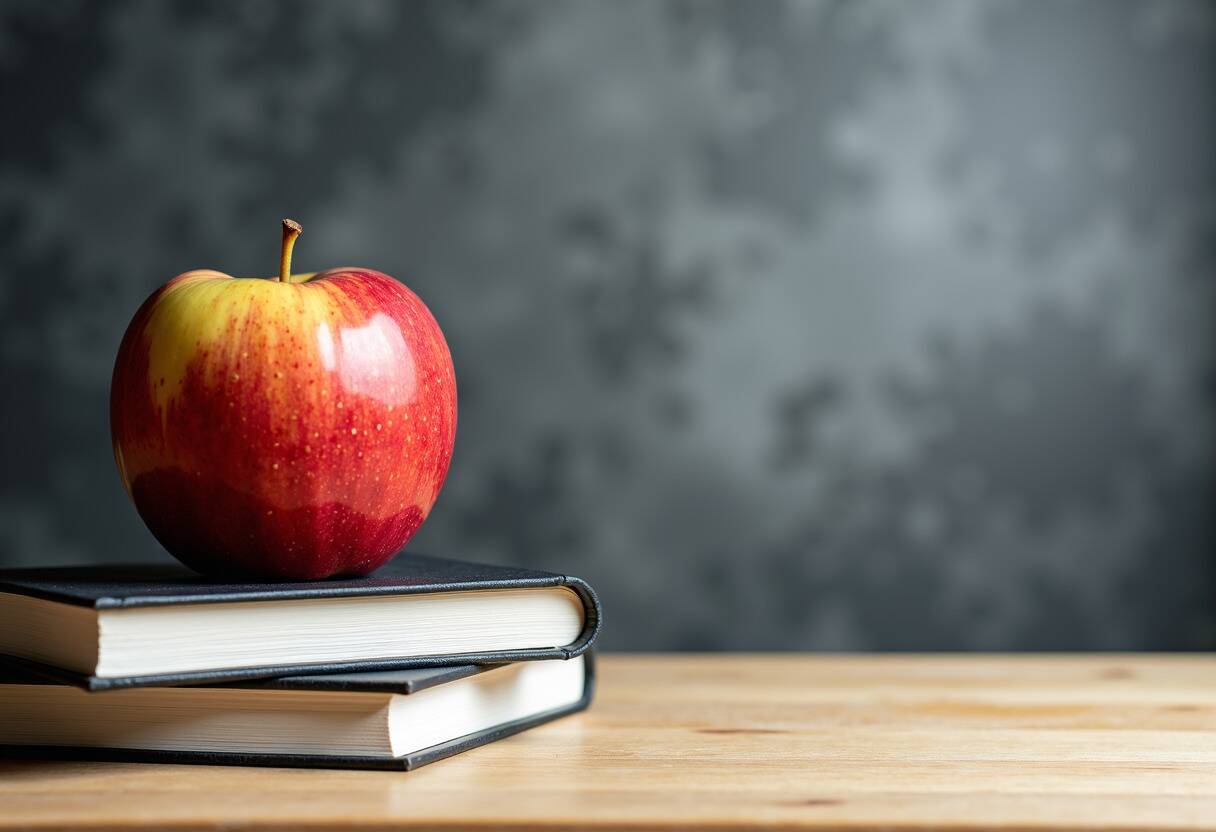} &
  \includegraphics[width=\imgw]{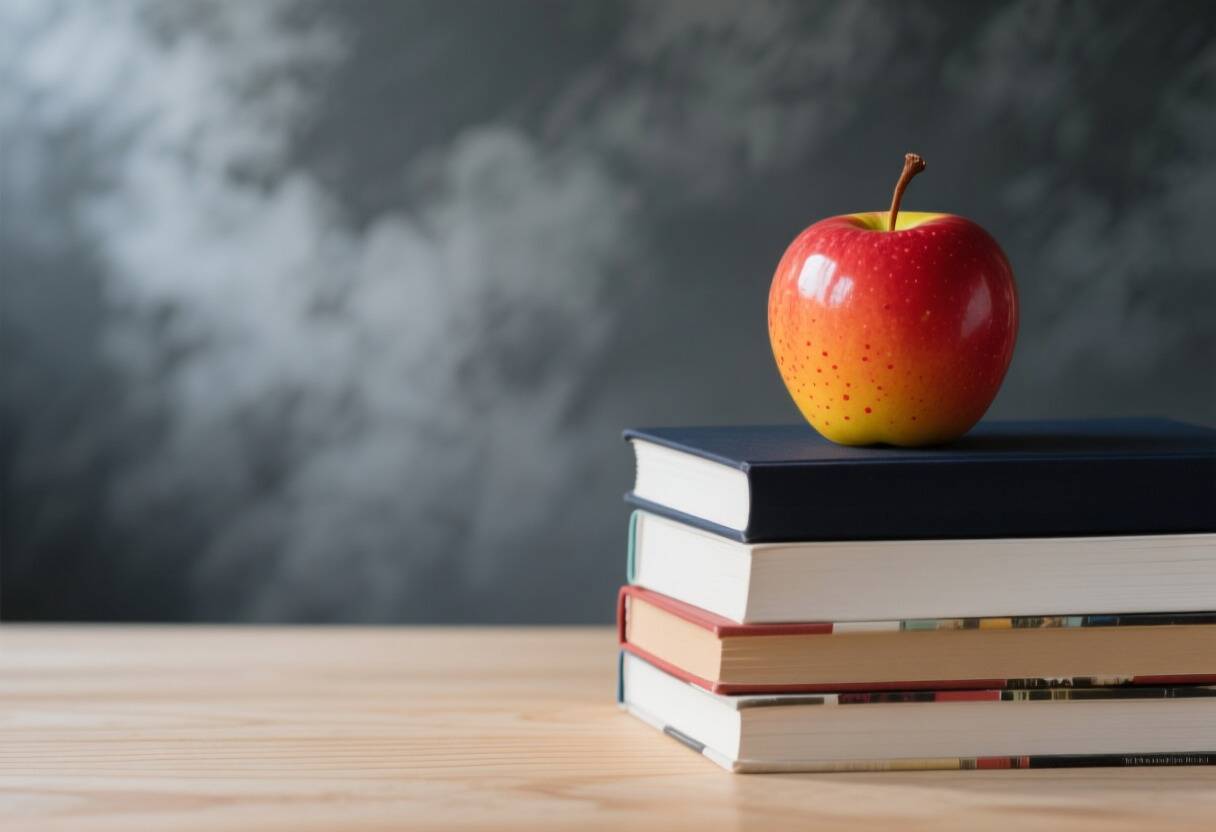} &
  \includegraphics[width=\imgw]{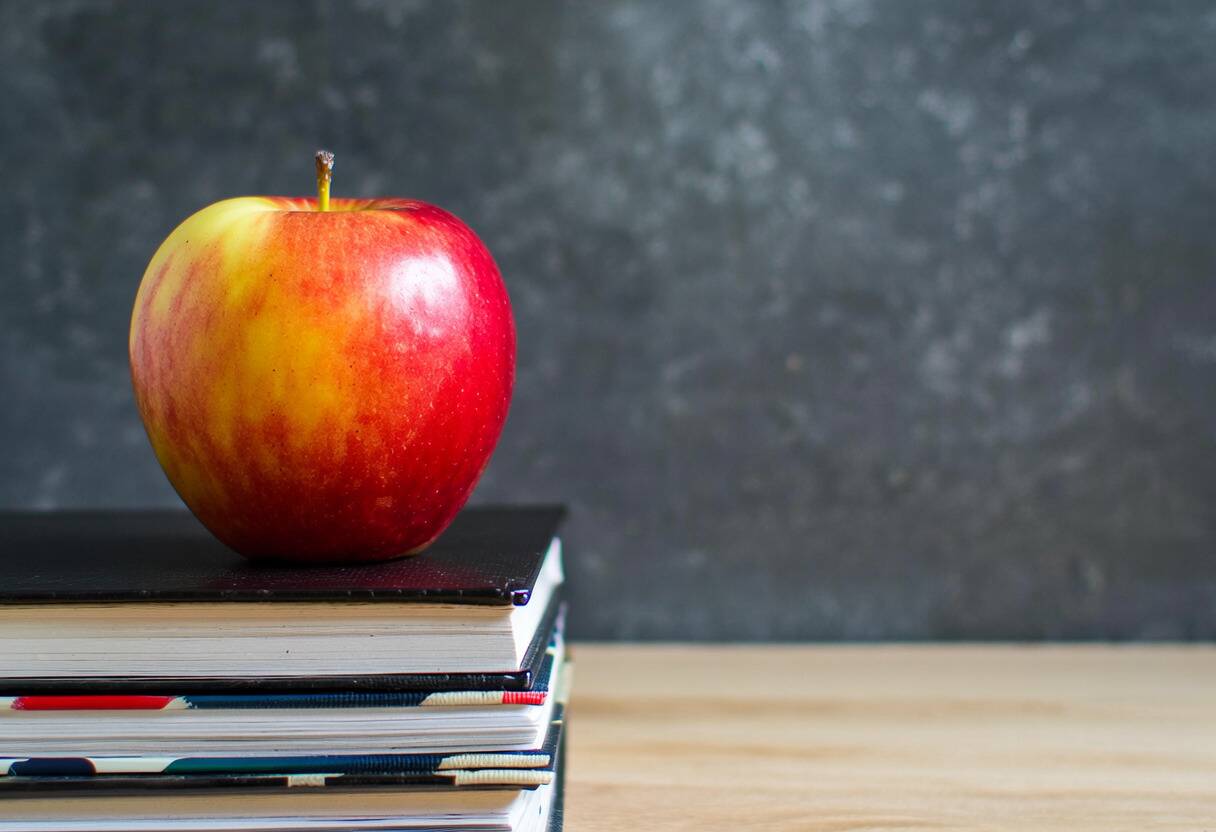}
  \\

  \includegraphics[width=\imgw]{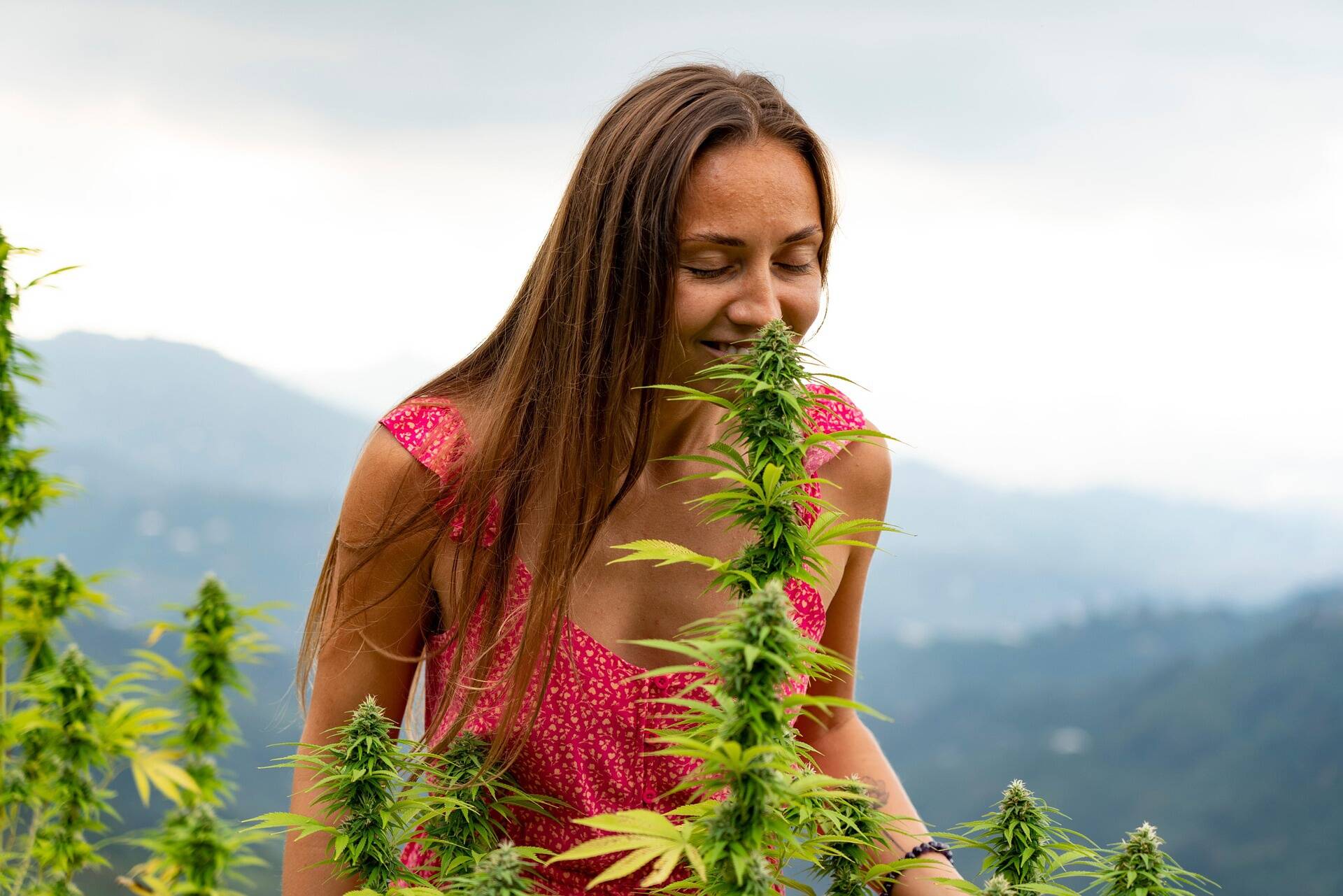} &
  \includegraphics[width=\imgw]{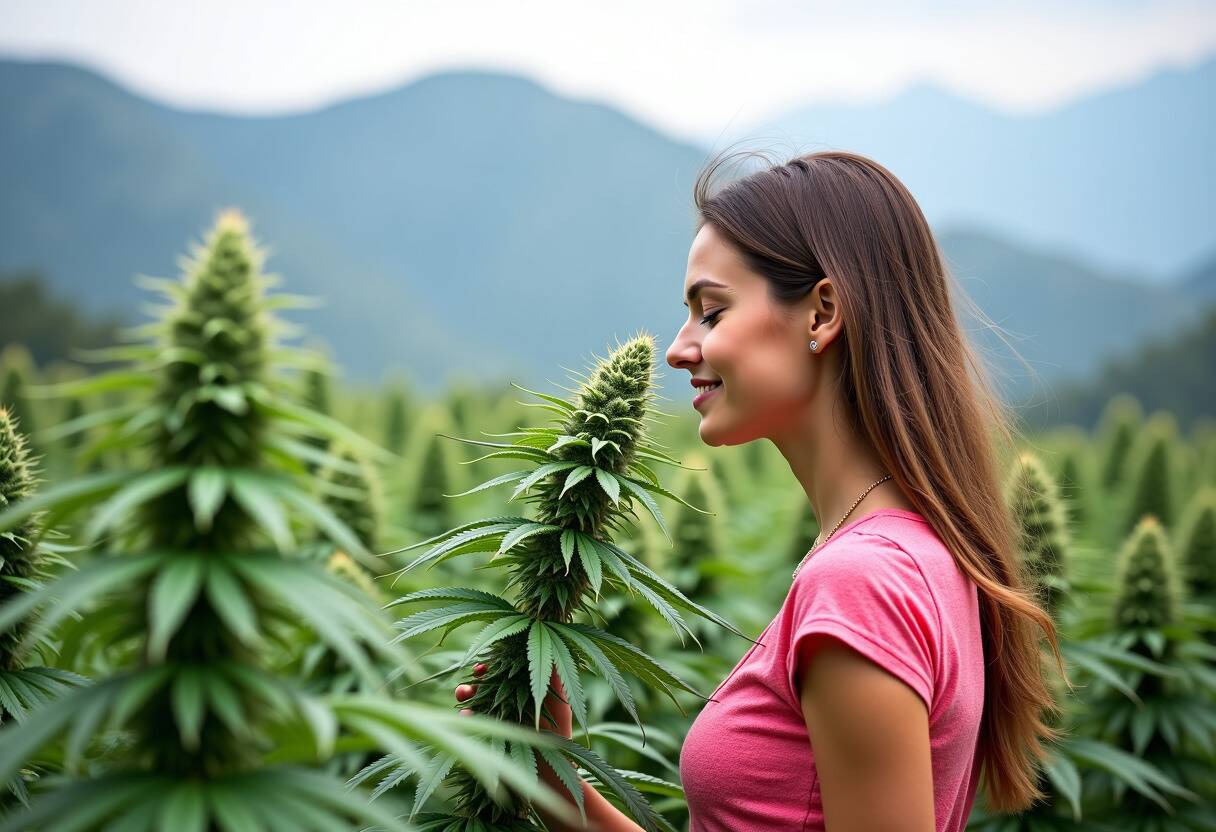} &
  \includegraphics[width=\imgw]{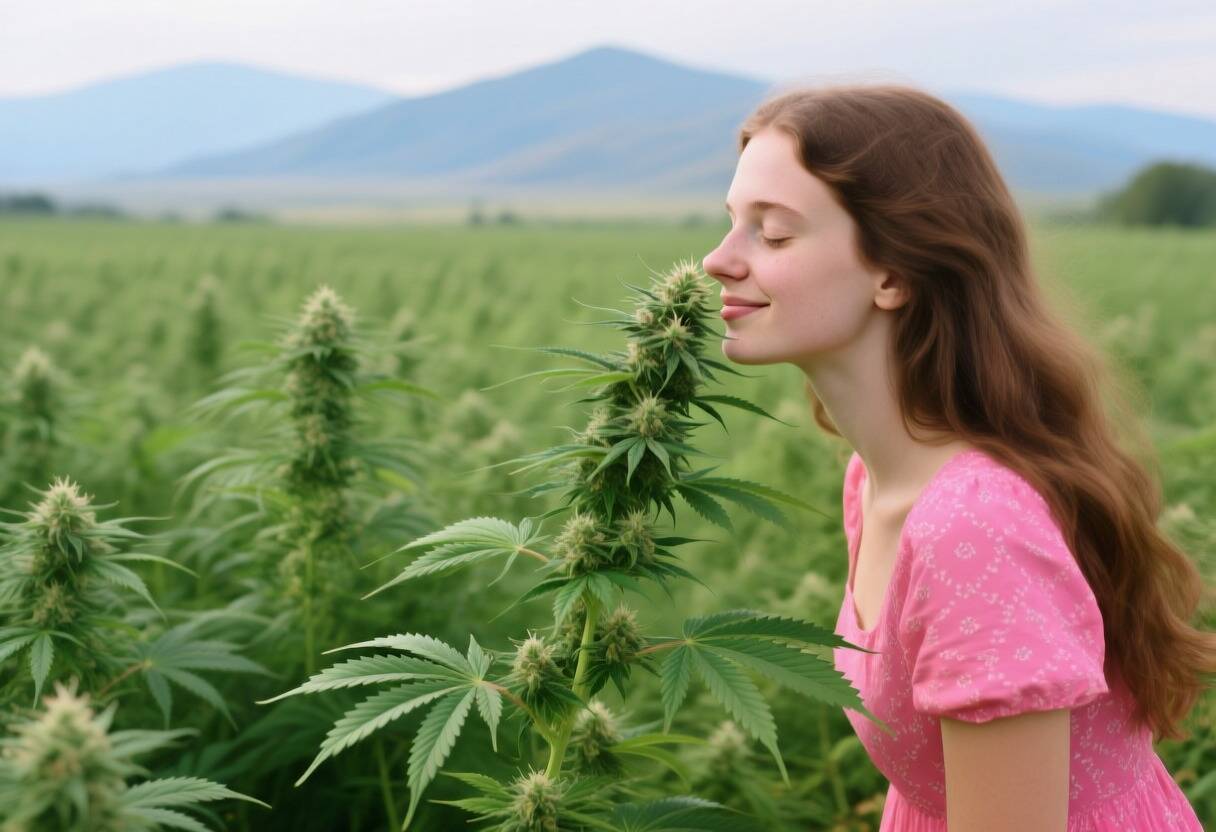} &
  \includegraphics[width=\imgw]{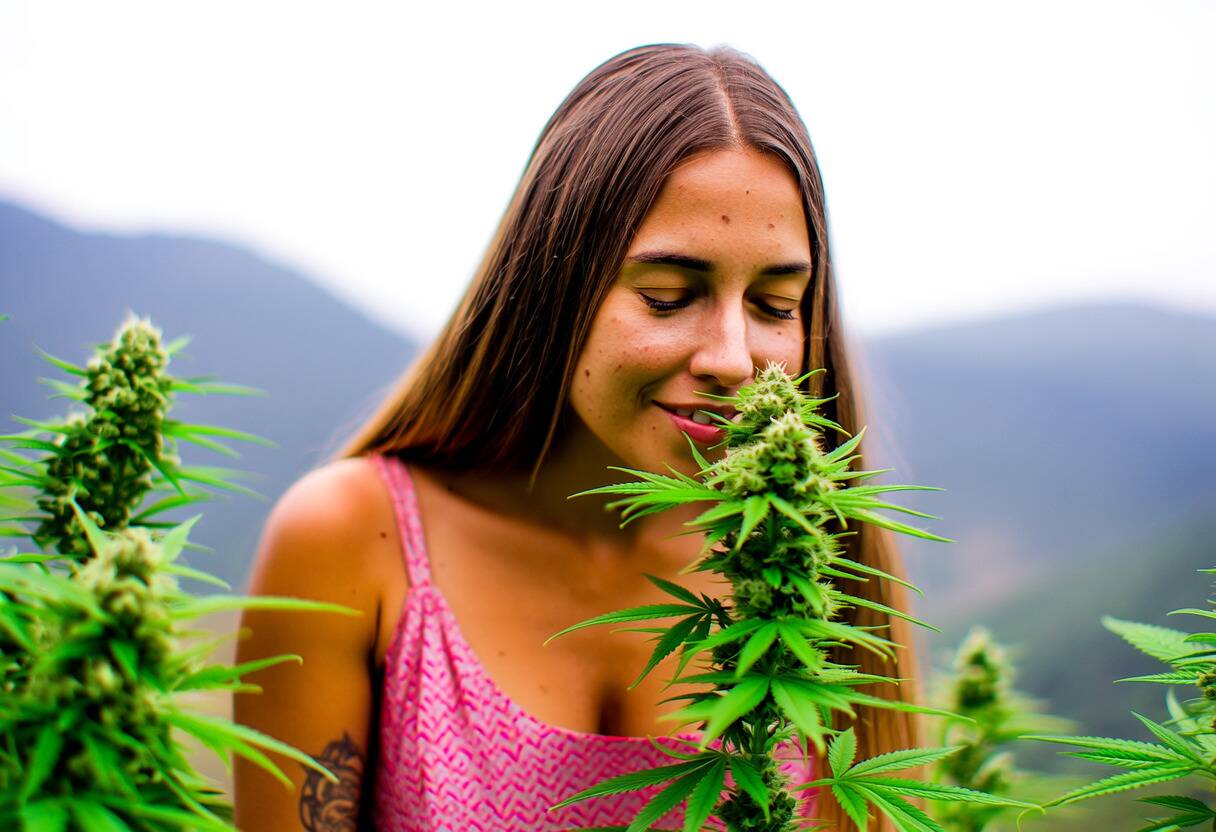}
  \\

  \parbox{\imgw}{\scriptsize\centering \textbf{Original}} &
  \parbox{\imgw}{\scriptsize\centering \textbf{Flux}} &
  \parbox{\imgw}{\scriptsize\centering \textbf{Qwen-Image}} &
  \parbox{\imgw}{\scriptsize\centering \textbf{\modelname{} (Ours)}}
  \\
\end{tabular}

\vspace{-3pt}
\caption{\textbf{Text-as-a-Bottleneck Reconstruction (TaBR).} We caption the original image and use the resulting text as a bottleneck input to the generator, producing a reconstructed image. This evaluation allows humans to assess a model’s expressive power objectively without reading extremely long captions. As shown, our model achieves noticeably higher similarity to the original image.}
\label{fig:tabr}
\end{figure}

%% file: Figures/arch/arch_table.tex
\begin{table}[t]
  \centering
  \footnotesize
  \begin{tabular}{cccccc}
    \toprule
    \textbf{\makecell{Dual-stream \\ Blocks}} &
    \textbf{\makecell{Single-stream \\ Blocks}} &    
    \textbf{\makecell{FFN \\ Dimension}} &
    \textbf{\makecell{Attention \\ Heads}} &
    \textbf{\makecell{Head dim}} &
    $\left(d_t, d_h, d_w\right)$ \\
    \midrule
    8 & 38 & 12288 & 24 & 128 & $(16,\,56,\,56)$ \\
    \bottomrule
  \end{tabular}
  \caption{\textbf{\modelname{} transformer architecture.}}
  \label{tab:model_arch}
\end{table}

%% file: tables/prism_licensed.tex
\begin{table*}[!t]\centering
\renewcommand{\arraystretch}{1.7} 
\setlength{\tabcolsep}{3pt}

\centering
\begin{adjustbox}{width=\textwidth}
\begin{tabular}{l|ccc|ccc|ccc|ccc|ccc|ccc|ccc}
\toprule
\multirow{2}{*}{\textbf{Model}}
  & \multicolumn{3}{c|}{\textbf{Imagination}}
  & \multicolumn{3}{c|}{\textbf{Text rendering}}
  & \multicolumn{3}{c|}{\textbf{Style}}
  & \multicolumn{3}{c|}{\textbf{Affection}}
  & \multicolumn{3}{c|}{\textbf{Composition}}
  & \multicolumn{3}{c|}{\textbf{Long text}}
  & \multicolumn{3}{c}{\textbf{Overall}} \\

\cmidrule(lr){2-22}

  & Ali. & Aes.& Avg.
  & Ali. & Aes.& Avg.   
  & Ali. & Aes.& Avg.     
  & Ali. & Aes.& Avg.        
  & Ali. & Aes.& Avg.     
  & Ali. & Aes.& Avg.     
  & Ali. & Aes.& Avg.     
  \\

\midrule

SD3.5-Large~\citep{SD35} & 75.4 & 75.2 & 75.3 & 56.1 & 68.3 & 62.2 & 78.3 & 81.0 & 79.7 & 87.7 & 86.4 & 87.0 & 86.6 & 83.1 & 84.9 & 67.8 & 59.2 & 63.5 & 78.8 & 77.6 & 78.2\\

FLUX.1-dev~\citep{flux2024} & 65.4 & 71.8 & 68.6 & 57.4 & 63.5 & 60.4 & 71.4 & 80.4 & 75.9 & 88.5 & \underline{89.3} & 88.9 & 89.4 & 85.4 & 87.4 & 73.3 & 66.8 & 70.1 & 77.0 & 78.7 & 77.8\\

FLUX.1-Krea-dev~\citep{fluxkrea} & 71.4 & 73.3 & 72.4 & 52.6 & 62.6 & 57.6 & 79.4 & 84.3 & 81.9 & 86.8 & \textbf{89.6} & 88.2 & 90.0 & 86.6 & 88.3 & 77.1 & 65.9 & 71.5 & 79.7 & 79.7 & 79.7\\

HiDream-I1-Full~\citep{cai2025hidream} & 74.1 & 74.4 & 74.2 & 64.8 & 72.2 & 68.5 & 79.3 & \underline{88.8} & 84.1 & \underline{90.8} & 87.8 & 89.3 & 90.3 & \underline{86.7} & 88.5 & 64.0 & 53.7 & 58.8 & 80.0 & 79.1 & 79.5\\

Qwen-Image~\citep{wu2025qwenimagetechnicalreport} & 79.4 & \underline{81.0} & 80.2 & \textbf{76.1} & \underline{76.5} & \textbf{76.3} & 79.9 & 86.4 & 83.1 & 85.8 & 87.2 & 86.5 & \textbf{94.2} & 84.4 & \underline{89.3} & 82.4 & 65.9 & 74.1 & 84.1 & 81.5 & 82.8\\


\midrule
Gemini2.5-Flash-Image~\citep{comanici2025gemini} & \textbf{91.4} & \textbf{86.9} & \textbf{89.1} & \underline{71.3} & \textbf{78.7} & \underline{75.0} & \textbf{89.9} & \textbf{90.6} & \textbf{90.2} & \textbf{97.4} & 88.4 & \textbf{92.9} & \underline{92.6} & \textbf{90.3} & \textbf{91.5} & \textbf{88.1} & \textbf{81.4} & \textbf{84.8} & \textbf{91.2} & \textbf{87.3} & \textbf{89.3}\\

\midrule

\modelname{} (Ours) & \underline{90.5} & 77.6 & \underline{84.0} & 66.5 & 72.2 & 69.3 & \underline{86.9} & 86.9 & \underline{86.9} & 90.3 & 88.4 & \underline{89.4} & 91.1 & 85.5 & 88.3 & \underline{83.2} & \underline{72.1} & \underline{77.7} & \underline{87.8} & \underline{82.1} & \underline{84.9}\\
\modelname{} open-VLM & 88.5 & 76.2 & 82.3 & 65.2 & 68.7 & 67.0 & 85.1 & 88.5 & 86.8 & 89.5 & 87.3 & 88.4 & 90.1 & 85.1 & 87.6 & 82.1 & 67.5 & 74.8 & 86.4 & 80.9 & 83.6\\
\bottomrule

\end{tabular}
\end{adjustbox}
\caption{\textbf{Quantitative results on PRISM-Bench-licensed:} a licensed variant of PRISM~\cite{fang2025flux}. We report alignment (Ali.), aesthetics (Aes.), and average (Avg.)  across categories. Best results are in \textbf{bold}; second best are \underline{underlined}. \modelname{} outperforms all open-source baselines.}
\label{tab:prism}

\end{table*}

%% file: tables/geneval.tex
\begin{table*}
\setlength{\tabcolsep}{3pt}
  \centering
  \begin{adjustbox}{width=\textwidth}
  \begin{tabular}{lcccccccc}
    \toprule
    \textbf{Model} & \textbf{Size (Parameters)} & \textbf{Single object} & \textbf{Two object} & \textbf{Counting} & \textbf{Colors} & \textbf{Position} & \textbf{Color attribution} & \textbf{Overall} \\    
    \midrule
    SD3.5-Large~\citep{SD35} & 8B & \textbf{1.00} & 0.91 & \underline{0.79} & 0.81 & 0.23 & 0.54 & 0.71 \\
    FLUX.1-dev~\citep{flux2024} & 12B & \underline{0.99} & 0.80 & 0.75 & 0.77 & 0.19 & 0.49 & 0.66 \\
    FLUX.1-Krea-dev~\citep{fluxkrea} & 12B & \underline{0.99} & 0.91 & 0.70 & 0.88 & 0.31 & 0.69 & 0.75 \\
    HiDream-I1-Full~\citep{cai2025hidream} & 17B & \textbf{1.00} & \textbf{0.98} & \underline{0.79} & \textbf{0.91} & 0.60 & \underline{0.72} & 0.83 \\
    Qwen-Image~\citep{wu2025qwenimagetechnicalreport} & 20B & \underline{0.99} & \underline{0.94} & \textbf{0.91} & 0.87 & \underline{0.77} & \textbf{0.75} & \textbf{0.87} \\
    \midrule
    \modelname{} (Ours) & 8B & \textbf{1.00} & 0.88 & \underline{0.79} & \underline{0.90} & \textbf{0.80} & 0.71 & \underline{0.85} \\
    \bottomrule
  \end{tabular}
  \end{adjustbox}
\caption{\textbf{Quantitative results on the GenEval benchmark.} Best results are shown in \textbf{bold}, and second best are \underline{underlined}. Unlike PRISM-Bench, GenEval focuses on relatively short captions. Accordingly, our model \modelname{} achieves comparable performance to other open-source models despite using fewer parameters.}

  \label{tab:geneval}
\end{table*}

%% file: sec/4_experiments.tex
\section{Experiments}
\label{sec:exp}
In this section, we present a comprehensive evaluation of \modelname{}. Evaluation method are described in Section~.\ref{sec:eval_metrics}, Qualitative results are provided in Section~\ref{sec:Qualitative_Results}, quantitative analyses in Section~\ref{sec:Quantitative_Results}, and ablation studies examining the effect of individual design choices are discussed within the Method Section~\ref{sec:method}.

\subsection{Evaluation metrics.}
\label{sec:eval_metrics}
We evaluate \modelname{} using four complementary protocols: PRISM-Bench-licensed~\cite{fang2025flux}, GenEval~\cite{ghosh2023geneval}, TaBR (Section~\ref{sec:eval}), and overall user preference—covering prompt adherence, compositional reasoning, aesthetics, and expressiveness. 
PRISM-Bench-licensed is an automatically derived variant of PRISM-Bench in which prompts referencing copyrighted or unlicensed concepts are filtered out, ensuring a fair comparison as \modelname{} is trained on 100\% licensed data (see Section~\ref{sec:ethics}). PRISM-Bench was designed to expose alignment gaps that saturate on earlier benchmarks; for example, GenEval’s single-object category often reaches near-perfect scores, masking differences across models. 
For completeness, we still report GenEval, which probes object count, color attribution, and spatial relations with controlled prompts. We follow the official protocols: PRISM is scored automatically with GPT-4.1-Vision, and GenEval using detectors and CLIP-based similarity.

TaBR measures long-caption expressiveness via a caption$\rightarrow$generation$\rightarrow$reconstruction loop with human raters comparing reconstructions to the original image (Section ~\ref{sec:eval}). Annotators are shown the original and two reconstructions (from different models) and asked: \textit{``Which Image is more similar to the one above?''}  We perform this measurement on a test-set of $60$ image that are not part of our training data. We also report overall user preference using blinded, side-by-side comparisons, asking: \textit{``Which image do you prefer?''} The evaluation covers a test set of 150 captions spanning diverse visual and stylistic domains. For both studies, image order and model identity are randomized per trial. We aggregate responses by computing, for each model, the percentage of questions won (win rate) and the percentage of ties (draw rate). In total, we collect over 5{,}000 votes to ensure a robust evaluation.


\subsection{Qualitative Results}
\label{sec:Qualitative_Results}

Figure~\ref{fig:tabr} shows qualitative examples from the TaBR evaluation set, with original reference images alongside reconstructions produced by FLUX, Qwen-Image, and \modelname{}. As illustrated, \modelname{} achieves the closest resemblance to the originals, preserving global structure and fine-grained details. In particular, it maintains pose, camera angle, and overall scene layout more faithfully than the baselines.

Figure~\ref{fig:contextual} presents \textit{Contextual Contradictions} drawn from ContraBench~\cite{huberman2025image} and Whoops~\cite{Bitton_Guetta_2023}, which probe robustness to atypical relations, whereas prior models often default to plausible but incorrect configurations. \modelname{} preserves the full set of specified attributes and relations, e.g., bear performing handstand and women writing with a dart. We attribute this fidelity to training on structured captions that explicitly encode concepts and context. Additional visual results are provided in the Appendix, Figure~\ref{fig:app_samples}, highlighting \modelname{}’s creative range. Figure~\ref{fig:app_disentanglement} provide additional examples for controllable refinement.


\subsection{Quantitative Results}
\label{sec:Quantitative_Results}

Table~\ref{tab:prism} reports results on PRISM-Bench-licensed. \modelname{} attains the highest alignment score among open-source models, substantially narrowing the prompt-alignment gap to closed systems such as Gemini~2.5~Flash~Image. Despite operating at a moderate 8B parameter scale, \modelname{} rivals larger models in prompt understanding and text–image grounding, underscoring the impact of structured-caption supervision.
We additionally report scores for \modelname{} when paired with a closed-source VLM~\cite{comanici2025gemini} and with our fine-tuned open-source alternative. Surprisingly, the open-source variant achieves competitive results while relying on a significantly smaller and more cost-effective VLM to produce the structured prompts.

On GenEval (Table~\ref{tab:geneval}), \modelname{} delivers consistently strong results across tasks, with a pronounced gain in positional understanding, a known weakness for most models~\cite{ghosh2023geneval}. We attribute this improvement to training on structurally detailed captions that explicitly encode spatial relations.

Table~\ref{tab:tabr-comparison} presents TaBR results. \modelname{} achieves the highest reconstruction fidelity across baselines, winning over 90\% of comparisons against SD3.5-Large and 88.9\% against HiDream-II-Full, while maintaining strong performance against FLUX and Qwen-Image. These findings indicate superior expressive power derived from long structured-caption training. Finally, in overall user preference (Table~\ref{tab:user-preference}), \modelname{} is consistently favored over SD3.5, FLUX.1-dev, and FLUX.1-Krea-dev, and attains comparable ratings to Qwen-Image and HiDream-II-Full—demonstrating competitive perceptual quality alongside strong prompt alignment.

\input{tables/TaBR}
\input{tables/user_preference}

%% file: tables/TaBR.tex
\begin{table}
{\footnotesize
\setlength{\tabcolsep}{5pt}
  \centering
  \begin{tabular}{lccccc}
    \toprule
    & \textbf{SD3.5} & \textbf{FLUX-dev} & \textbf{FLUX-Krea} & \textbf{HiDream} & \textbf{Qwen} \\    
    \midrule
    \modelname{} & \textbf{90.5\%} & \textbf{76.9\%} & \textbf{66.4\%} & \textbf{88.9\%} & \textbf{59.2\%} \\
    Baseline & 2.9\% & 8.0\% & 13.9\% & 4.5\% & 22.0\% \\
    Draw & 6.6\% & 15.1\% & 19.7\% & 6.6\% & 18.8\% \\
    \bottomrule
  \end{tabular}
  }
\caption{\textbf{Text-as-a-Bottleneck Reconstruction (TaBR) evaluation.} Win rates of \modelname{} against each model. \modelname{} outperforms all open-source baselines, reflecting greater expressive power due to training on long structured captions.}

  \label{tab:tabr-comparison}

\end{table}




%% file: tables/user_preference.tex
\begin{table}
{\footnotesize
\setlength{\tabcolsep}{4.5pt}
  \centering
  \begin{tabular}{lccccc}
    \toprule
     & \textbf{SD3.5} &  \textbf{FLUX-dev} & \textbf{FLUX.1-Krea} & \textbf{HiDream} & \textbf{Qwen} \\    
    \midrule
    \modelname{} & \textbf{69.1\%} & \textbf{58.6\%} & \textbf{47.5\%} & 42.3\% & \textbf{42.1\%} \\
    Baseline & 19\% & 30.4\% & 35.1\% & \textbf{43.5\%} & 40.2\% \\
    Draw & 11.9\% & 11\% & 17.4\% & 19.7\% & 17.7\% \\
    \bottomrule
  \end{tabular}
  }
  \caption{\textbf{Overall user preference.} \modelname{} clearly surpasses SD3.5, FLUX.1-dev and Flux.1-Krea-dev, and achieves comparable performance to Qwen-Image and HiDream.}
  \label{tab:user-preference}
\end{table}

%% file: sec/5_conclusion.tex
\vspace{-0.2cm}

\section{Conclusions}

Most prior work improves image generators under a fixed dataset regime. We take a different path: we scale the language itself by training on long structured captions that encode far richer visual and compositional detail than any prior open-source approach. This shift changes learning dynamics and behavior: long captions accelerate convergence and yield native disentanglement, so single-attribute edits (lighting, depth of field, expression) modify only the intended factor while the rest remains stable. To make this feasible, we introduced DimFusion, a novel architecture that integrates intermediate and final LLM representations without increasing token length. And because thousand-word prompts outstrip human evaluators, we proposed Text-as-a-Bottleneck Reconstruction (TaBR)—an image-grounded protocol to measure expressiveness and controllability. All components culminate in \modelname{}, a large-scale text-to-image model that outperforms larger baselines, demonstrating that caption scale and structure are levers for controllable generation. Looking ahead, our results suggest a broader design space: we can build intermediate languages that unlock new user interfaces. Users continue to write natural language, which is translated into a structured intermediate dialect that enables new forms of creative control. This paves the way for predictable graphics primitives atop image generation models.

\section{Ethical Statement}
\label{sec:ethics}

Most text-to-image models are trained on data scraped online without permission or licenses from the rightful owners, a practice that violates intellectual property rights and yields models that cannot guarantee the copyright status of their outputs. In contrast, our model is trained solely on licensed data, substantially reducing the risk of infringement. We believe the AI ethics community should prioritize licensed, consent-based training, as responsible data use is essential for a sustainable AI ecosystem. Complementing this stance, our work investigates how structured synthetic captions and LLM fusion can improve fidelity and controllability: by strengthening prompt adherence and semantic grounding, our approach can help reduce unintended biases and misalignment. At the same time, greater controllability introduces dual-use risks, as it may also be exploited to generate misleading, or inappropriate content more reliably. We therefore advocate openly studying both the benefits and risks of these techniques to develop effective safeguards.

%% file: sec/X_suppl.tex
\clearpage

\begin{appendices}
\section{Structured Captions}
\label{app:structured-json}

In this section we describe our structured captioning strategy. 
Each caption is a long, detailed JSON that decomposes the image into interpretable fields, covering the objects, background, lighting, aesthetics, photographic attributes, style, and text. This structure guarantees full semantic coverage and improves controllability during training and inference.
The structure of our JSON is described in Table~\ref{tab:JSON-strcuture}.\input{tables/JSON_structure}

Our structured captions are generated using Gemini 2.5 with a dedicated system prompt (see Table~\ref{tab:system_prompt}) and using Gemini's structured outputs configured with our JSON schema (see Table~\ref{tab:strcutured_outputs}) that enforce completeness, consistency, and adherence to the predefined schema. 
\input{tables/JSON_system_prompt}
\input{tables/JSON_gemini_strcture_output}

\section{Additional Training Details}\label{sec:additional-training-details}

The data distribution is illustrated in Figure~\ref{fig:data_distribution}.
\input{Figures/data_distribution/data_distribution}

We train the model using the AdamW optimizer~\cite{loshchilov2017decoupled} with weight decay of $1\times10^{-4}$, $\beta_1=0.9$, $\beta_2=0.999$, and $\epsilon=1\times10^{-15}$. The learning rate is set to $1\times10^{-4}$ with a constant schedule and a warmup of $10$K steps, followed by an additional $2.5$K warmup steps when increasing resolution. Training follows the flow-matching formulation~\cite{lipman2023flowmatchinggenerativemodeling}, with a logit-normal noise schedule combined with resolution-dependent timestep shifting~\cite{esser2024scaling}.

We train the model in stages of increasing resolution. The first stage is conducted at $256^2$ resolution for $652$k steps, where the initial $300$k steps employ REPA~\cite{yu2024representation} using DINOv2-L encodings~\cite{oquab2023dinov2} and REPA coefficient of 0.1. Training then continues for an additional $100$k steps at $512^2$ resolution, followed by $50$k steps at $1024^2$. At each resolution stage, the effective batch size starts at approximately $1$K and is gradually increased to $2$K as training progresses.

Post-training, we perform aesthetic finetuning with 3,000 hand-picked images, followed by DPO training~\cite{wallace2024diffusion} with dynamic beta~\cite{liu2025improving} to improve text rendering.

\section{Additional Ablation Studies Details}
\label{sec:appendix_ablation}

We provide implementation details for the ablation studies discussed in Section~\ref{sec:captions} (long structured captions vs. short captions) and Section~\ref{sec:dimfusion} (convergence behavior across architectures).
To reduce compute cost, all ablations were conducted under identical training and sampling conditions using 1B-parameter models trained with the SDXL VAE \cite{podell2023sdxl}.
Table \ref{tab:ablations_model_arch} lists the architectural hyperparameters used in all experiments.
\input{tables/ablation_studies_arch}

In Section \ref{sec:captions}, we examine the effect of caption length using the \emph{DimFusion} architecture with \emph{SmolLM3-3B} as the LLM, trained once with long structured captions and once with short captions.
In Section \ref{sec:dimfusion}, we compare three architectures trained on long structured captions.
The baseline follows \cite{esser2024scaling} and uses T5 \cite{raffel2020exploring} as the text encoder.
We further evaluate two large-language-model fusion variants that share the same hyperparameters as the baseline:
(1) \emph{DimFusion} with \emph{SmolLM3-3B} as the LLM; and
(2) \emph{TokenFusion}, following HiDream \cite{cai2025hidream}, concatenates SmolLM3-3B hidden states with the text embeddings along the \emph{sequence} dimension, effectively doubling the number of text tokens in each attention layer.

All models were trained on 70M licensed image-caption pairs, with the same optimizer, learning rate and noise scheduling described in Appendix~\ref{sec:additional-training-details}, and effective batch size of 1024. We also employed REPA~\cite{yu2024representation} with DINOv2-L encodings~\cite{oquab2023dinov2} and a REPA coefficient of 0.5.
For evaluation, FID was computed on 30K samples from the COCO-2014~\cite{lin2014microsoft} validation split, using 50 inference steps and no classifier-free guidance.

\section{Additional Samples from \modelname{}}

Additional samples from \modelname{} in various aspect ratios appear in Figure~\ref{fig:app_samples}, and additional examples for refinement and disentanglement appear in Figure~\ref{fig:app_disentanglement}.
\input{Figures/samples/app_samples}
\input{Figures/disentanglement/app_dis_fig}



\end{appendices}

%% file: tables/JSON_structure.tex
\begin{table*}[t]
\small
\setlength{\tabcolsep}{6pt}
\renewcommand{\arraystretch}{1.02}
\centering
\begin{tabularx}{\textwidth}{@{}p{3.9cm}p{2.4cm}X@{}}
\toprule
\textbf{Field} & \textbf{Type} & \textbf{Description} \\
\midrule
\texttt{short\_description} & String & A short summary of the image content. \\

\texttt{objects} & Array of Objects &
List of prominent objects. For each object, we record the following fields:
a detailed \texttt{description} of the object, the object's \texttt{location}, \texttt{relative\_size}, \texttt{shape\_and\_color}, \texttt{texture} (opt.), \texttt{appearance\_details} (opt.),
\texttt{relationship} with other objects, and the object's \texttt{orientation} (opt.).
When a human is described, the following fields are also included: the human's \texttt{pose} (opt.), \texttt{expression} (opt.), \texttt{clothing} (opt.), \texttt{action} (opt.), \texttt{gender} (opt.) and \texttt{skin\_tone\_and\_texture} (opt.).
If the image is a cluster of objects then also includes the
\texttt{number\_of\_objects} (opt.).\\

\texttt{background\_setting} & String &
Description of the environment/background. \\

\texttt{lighting} & Object &
Illumination details, including
\texttt{conditions} (e.g., daylight, studio),
\texttt{direction} (key-light orientation), and
\texttt{shadows} (opt.). \\

\texttt{aesthetics} & Object &
Global visual properties, including
\texttt{composition} (e.g., rule of third, symmetrical),
\texttt{color\_scheme} (dominant palette) and
\texttt{mood\_atmosphere} (emotional tone). \\

\shortstack[l]{\texttt{photographic\_}\\\texttt{characteristics}} & Object (optional) &
Camera/focus parameters:
\texttt{depth\_of\_field},
\texttt{focus},
\texttt{camera\_angle}, and
\texttt{lens\_focal\_length}. \\

\texttt{style\_medium} & String &
Artistic/rendering medium (e.g., photograph, oil painting, 3D render). \\

\texttt{text\_render} & Array of Objects &
A list of prominent text renders in the image. For each text render it includes the
\texttt{text},
\texttt{location},
\texttt{size},
\texttt{color},
\texttt{font} and
\texttt{appearance\_details} (opt.). \\

\texttt{context} & String & Additional context that helps understand the image better. \\

\texttt{artistic\_style} & String & High-level artistic/stylistic description (e.g., cinematic, minimalism). \\
\bottomrule
\end{tabularx}
\caption{\textbf{Structure of the JSON-based captions.} Each field captures a distinct aspect of the image, ensuring semantic completeness and consistency.}
\label{tab:JSON-strcuture}
\end{table*}

%% file: tables/JSON_system_prompt.tex
\begin{table*}[t]
\centering
\begin{tcolorbox}[
    colback=gray!3,
    colframe=gray!30,
    sharp corners,
    left=3pt,right=3pt,top=2pt,bottom=2pt,  
    before skip=2pt, after skip=2pt,        
    fontupper=\ttfamily\scriptsize,         
]
You are a meticulous and perceptive Visual Art Director working for a leading Generative AI company. Your expertise lies in analyzing images and extracting detailed, structured information.
Your primary task is to analyze provided images and generate a comprehensive JSON object describing them. Adhere strictly to the following structure and guidelines:

The output MUST be ONLY a valid JSON object. Do not include any text before or after the JSON object (e.g., no "Here is the JSON:", no explanations, no apologies).

IMPORTANT: When describing human body parts, positions, or actions, always describe them from the PERSON'S OWN PERSPECTIVE, not from the observer's viewpoint. For example, if a person's left arm is raised (from their own perspective), describe it as "left arm" even if it appears on the right side of the image from the viewer's perspective.

The JSON object must contain the following keys precisely:

1. \textbf{`short\_description`}: (String) A concise summary of the image content, 200 words maximum.

2. \textbf{`objects`}: (Array of Objects) List a maximum of 5 prominent objects if there are more than 5, list them in the background. For each object, include:

\quad    * \textbf{`description`}: (String) A detailed description of the object, 100 words maximum.
    
\quad    * \textbf{`location`}: (String) E.g., "center", "top-left", "bottom-right foreground".
    
\quad    * \textbf{`relative\_size`}: (String) E.g., "small", "medium", "large within frame".
    
\quad    * \textbf{`shape\_and\_color`}: (String) Describe the basic shape and dominant color.
    
\quad    * \textbf{`texture`}: (String) E.g., "smooth", "rough", "metallic", "furry".
    
\quad    * \textbf{`appearance\_details`}: (String) Any other notable visual details.
    
\quad    * \textbf{`relationship`}: (String) Describe the relationship between the object and the other objects in the image.
    
\quad    * \textbf{`orientation`}: (String) Describe the orientation or positioning of the object, e.g., "upright", "tilted 45 degrees", "horizontal", "vertical", "facing left", "facing right", "upside down", "lying on its side".
    
If the object is a human or human-like entity, include:

\quad * \textbf{`pose`}: (String) Describe the body position.

\quad * \textbf{`expression`}: (String) Describe facial expression and emotion. E.g., "winking", "joyful", "serious", "surprised", "calm".

\quad        * \textbf{`clothing`}: (String) Describe attire.

\quad        * \textbf{`action`}: (String) Describe the action of the human.

\quad        * \textbf{`gender`}: (String) Describe the gender of the human.
        
\quad        * \textbf{`skin\_tone\_and\_texture`}: (String) Describe the skin tone and texture.

If the object is a cluster of objects, include the following:

\quad        * \textbf{`number\_of\_objects`}: (Integer) The number of objects in the cluster.

3. \textbf{`background\_setting`}: (String) Describe the overall environment, setting, or background, including any notable background elements that are not part of the objects section.

4. \textbf{`lighting`}: (Object)

\quad    * \textbf{`conditions`}: E.g., "bright daylight", "dim indoor", "studio lighting", "golden hour".
    
\quad    * \textbf{`direction`}: E.g., "front-lit", "backlit", "side-lit from left".
    
\quad    * \textbf{`shadows`}: Describe the presence of shadows.

5. \textbf{`aesthetics`}: (Object)

\quad    * \textbf{`composition`}: E.g., "rule of thirds", "symmetrical", "centered", "leading lines".

\quad    * \textbf{`color\_scheme`}: E.g., "monochromatic blue", "warm complementary colors", "high contrast".

\quad    * \textbf{`mood\_atmosphere`}: E.g., "serene", "energetic", "mysterious", "joyful".

6. \textbf{`photographic\_characteristics`}: (Object)

\quad    * \textbf{`depth\_of\_field`}: (String) E.g., "shallow", "deep", "bokeh background".

\quad    * \textbf{`focus`}: (String) E.g., "sharp focus on subject", "soft focus", "motion blur".

\quad    * \textbf{`camera\_angle`}: (String) E.g., "eye-level", "low angle", "high angle", "dutch angle".

\quad    * \textbf{`lens\_focal\_length`}: (String) E.g., "wide-angle", "telephoto", "macro", "fisheye".

7. \textbf{`style\_medium`}: (String) Identify the artistic style or medium (e.g., "photograph", "oil painting", "watercolor", "3D render", "digital illustration", "pencil sketch") If the style is not "photograph", but artistic, please describe the specific artistic characteristics under 'artistic\_style', 50 words maximum.

8. \textbf{`artistic\_style`}: (String) Describe specific artistic characteristics, 3 words maximum.

9. \textbf{`context`}: (String) Provide any additional context that helps understand the image better. This should include a general description of the type of image (e.g., Fashion Photography, Product Shot, Magazine Cover, Nature Photography, Art Piece, etc.), as well as any other relevant contextual information that situates the image within a broader category or intended use. For example: "This is a high-fashion editorial photograph intended for a magazine spread"

10. \textbf{`text\_render`}: (Array of Objects) List of a maximum of 5 most prominent text renders in the image. For each text render, include:

\quad    * \textbf{`text`}: (String) The text content.
    
\quad    * \textbf{`location`}: (String) E.g., "center", "top-left", "bottom-right foreground".
    
\quad    * \textbf{`size`}: (String) E.g., "small", "medium", "large within frame".
    
\quad    * \textbf{`color`}: (String) E.g., "red", "blue", "green".
    
\quad    * \textbf{`font`}: (String) E.g., "realistic", "cartoonish", "minimalist".
    
\quad    * \textbf{`appearance\_details`}: (String) Any other notable visual details.

Ensure the information within the JSON is accurate, detailed where specified, and avoids redundancy between fields.
\end{tcolorbox}
\caption{System prompt used to generate structured captions.}
\label{tab:system_prompt}
\end{table*}

%% file: tables/JSON_gemini_strcture_output.tex
\begin{table*}[t]
\centering

\begin{tcolorbox}[
    colback=gray!3,
    colframe=gray!30,
    sharp corners,
    left=3pt,right=3pt,top=2pt,bottom=2pt,
    before skip=2pt, after skip=2pt,        
    fontupper=\ttfamily\scriptsize,    
]
\char123
"type": "OBJECT",

"properties": \char123

\quad \textbf{"short\_description"}: \char123
"type": "STRING"
\char125,

\quad \textbf{"objects"}: \char123
"type": "ARRAY",
"items": \char123
"type": "OBJECT",
"properties": \char123

\quad \quad \textbf{"description"}: \char123
"type": "STRING"
\char125,

\quad \quad \textbf{"location"}: \char123
"type": "STRING"
\char125,

\quad \quad \textbf{"relationship"}: \char123
"type": "STRING"
\char125,

\quad \quad \textbf{"relative\_size"}: \char123
"type": "STRING"
\char125,

\quad \quad \textbf{"shape\_and\_color"}: \char123
"type": "STRING"
\char125,

\quad \quad \textbf{"texture"}: \char123
"type": "STRING",
"nullable": true
\char125,

\quad \quad \textbf{"appearance\_details"}: \char123
"type": "STRING",
"nullable": true
\char125,

\quad \quad \textbf{"number\_of\_objects"}: \char123
"type": "INTEGER",
"nullable": true
\char125,

\quad \quad \textbf{"pose"}: \char123
"type": "STRING",
"nullable": true
\char125,

\quad \quad \textbf{"expression"}: \char123
"type": "STRING",
"nullable": true
\char125,

\quad \quad \textbf{"clothing"}: \char123
"type": "STRING",
"nullable": true
\char125,

\quad \quad \textbf{"action"}: \char123
"type": "STRING",
"nullable": true
\char125,

\quad \quad \textbf{"gender"}: \char123
"type": "STRING",
"nullable": true
\char125,

\quad \quad \textbf{"skin\_tone\_and\_texture"}: \char123
"type": "STRING",
"nullable": true
\char125,

\quad \quad \textbf{"orientation"}: \char123
"type": "STRING",
"nullable": true
\char125
\char125,

\quad \quad "required": [
"description",
"location",
"relationship",
"relative\_size",
"shape\_and\_color",
"texture",
"appearance\_details",
"number\_of\_objects",
"pose",
"expression",
"clothing",
"action",
"gender",
"skin\_tone\_and\_texture",
"orientation"
]
\char125
\char125,

\quad \textbf{"background\_setting"}: \char123
"type": "STRING"
\char125,

\quad \textbf{"lighting"}: \char123
"type": "OBJECT",
"properties": \char123

\quad \quad \textbf{"conditions"}: \char123
"type": "STRING"
\char125,

\quad \quad \textbf{"direction"}: \char123
"type": "STRING"
\char125,

\quad \quad \textbf{"shadows"}: \char123
"type": "STRING",
"nullable": true
\char125
\char125,

\quad \quad "required": [
"conditions",
"direction",
"shadows"
]
\char125,

\quad \textbf{"aesthetics"}: \char123
"type": "OBJECT",
"properties": \char123

\quad \quad \textbf{"composition"}: \char123
"type": "STRING"
\char125,

\quad \quad \textbf{"color\_scheme"}: \char123
"type": "STRING"
\char125,

\quad \quad \textbf{"mood\_atmosphere"}: \char123
"type": "STRING"
\char125
\char125,

"required": [
"composition",
"color\_scheme",
"mood\_atmosphere"
]
\char125,

\quad \textbf{"photographic\_characteristics"}: \char123
"type": "OBJECT",
"properties": \char123

\quad\quad\textbf{"depth\_of\_field"}: \char123
"type": "STRING"
\char125,

\quad\quad\textbf{"focus"}: \char123
"type": "STRING"
\char125,

\quad\quad\textbf{"camera\_angle"}: \char123
"type": "STRING"
\char125,

\quad\quad\textbf{"lens\_focal\_length"}: \char123
"type": "STRING"
\char125
\char125,

\quad\quad"required": [
"depth\_of\_field",
"focus",
"camera\_angle",
"lens\_focal\_length"
],
"nullable": true
\char125,

\quad \textbf{"style\_medium"}: \char123
"type": "STRING"
\char125,

\quad \textbf{"text\_render"}: \char123
"type": "ARRAY",
"items": \char123
"type": "OBJECT",
"properties": \char123

\quad\quad\textbf{"text"}: \char123
"type": "STRING"
\char125,

\quad\quad\textbf{"location"}: \char123
"type": "STRING"
\char125,

\quad\quad\textbf{"size"}: \char123
"type": "STRING"
\char125,

\quad\quad\textbf{"color"}: \char123
"type": "STRING"
\char125,

\quad\quad\textbf{"font"}: \char123
"type": "STRING"
\char125,

\quad\quad\textbf{"appearance\_details"}: \char123
"type": "STRING",
"nullable": true
\char125
\char125,

\quad\quad"required": [
"text",
"location",
"size",
"color",
"font",
"appearance\_details"
]
\char125
\char125,

\quad \textbf{"context"}: \char123
"type": "STRING"
\char125,

\quad \textbf{"artistic\_style"}: \char123
"type": "STRING"
\char125
\char125,

"required": [
"short\_description",
"objects",
"background\_setting",
"lighting",
"aesthetics",
"photographic\_characteristics",
"style\_medium",
"text\_render",
"context",
"artistic\_style"
]
\char125
\end{tcolorbox}
\caption{Gemini's Structured Outputs.}\label{tab:strcutured_outputs}
\end{table*}

%% file: Figures/data_distribution/data_distribution.tex
\begin{figure*}[t]
    \centering
    \includegraphics[width=0.6\textwidth]{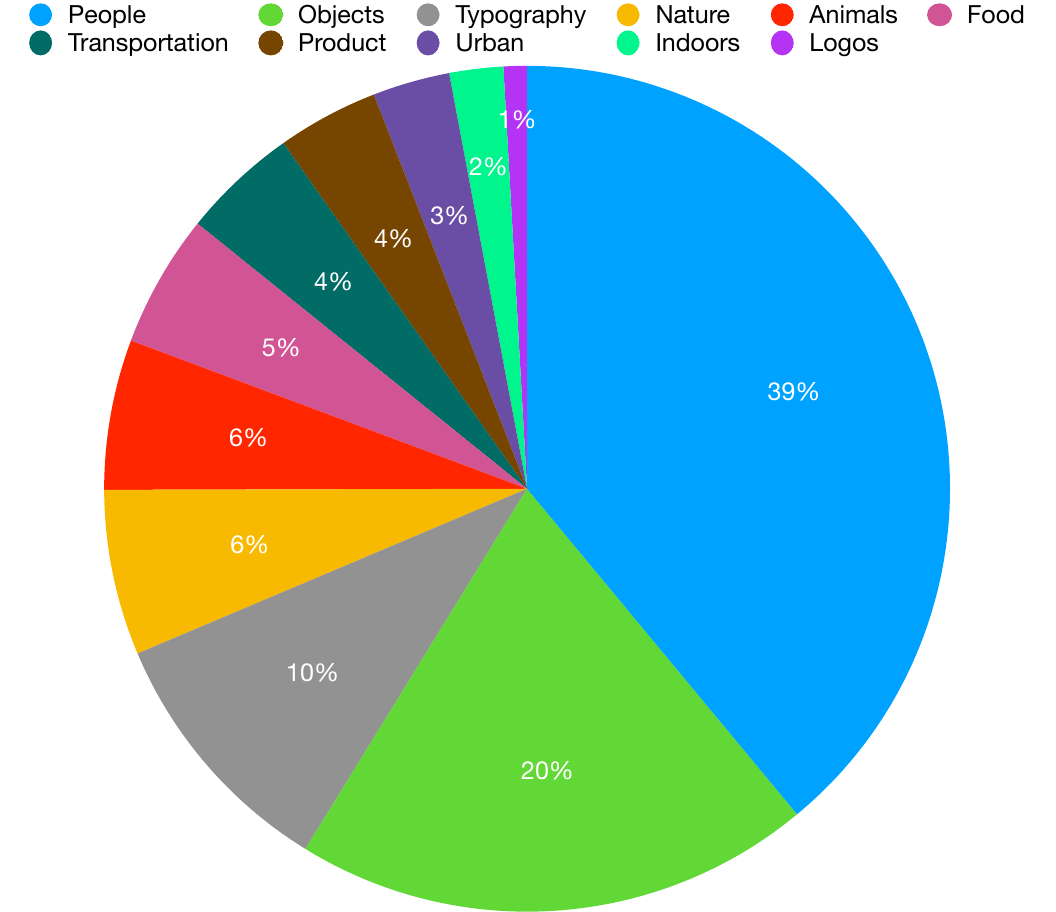}
    \vspace{-3mm}
    \caption{\textbf{Data distribution across categories.} 
    People and objects dominate the dataset (39\% and 20\%, respectively), followed by typography (10\%), nature (6\%), animals (6\%), food (5\%), transportation (4\%), product (4\%), urban scenes (3\%), indoors (2\%), and logos (1\%).}
    \label{fig:data_distribution}
    \vspace{-2mm}
\end{figure*}

%% file: tables/ablation_studies_arch.tex
\begin{table}[H]
  \centering
  \footnotesize
  \begin{tabular}{cccccc}
    \toprule
    \textbf{\makecell{Dual-stream \\ Blocks}} &
    \textbf{\makecell{Single-stream \\ Blocks}} &    
    \textbf{\makecell{FFN \\ Dimension}} &
    \textbf{\makecell{Attention \\ Heads}} &
    \textbf{\makecell{Head dim}} &
    $\left(d_t, d_h, d_w\right)$ \\
    \midrule
    8 & 20 & 6144 & 24 & 64 & $(4,\,30,\,30)$ \\
    \bottomrule
  \end{tabular}
  \caption{Model configuration for the 1B-parameter models used in the ablation studies. Each model is trained under identical conditions, differing only by caption type (long vs. short) or architectural variant (Baseline, DimFusion, TokenFusion).}
  \label{tab:ablations_model_arch}
\end{table}

%% file: Figures/samples/app_samples.tex
\newcommand{\ResRow}[4]{%
  \begin{minipage}[t]{0.24\textwidth}\centering
    \includegraphics[width=\linewidth,keepaspectratio]{#1}
  \end{minipage}\hfill
  \begin{minipage}[t]{0.24\textwidth}\centering
    \includegraphics[width=\linewidth,keepaspectratio]{#2}
  \end{minipage}\hfill
  \begin{minipage}[t]{0.24\textwidth}\centering
    \includegraphics[width=\linewidth,keepaspectratio]{#3}
  \end{minipage}\hfill
  \begin{minipage}[t]{0.24\textwidth}\centering
    \includegraphics[width=\linewidth,keepaspectratio]{#4}
  \end{minipage}
}

\begin{figure*}[t]
  \centering
  \setlength{\tabcolsep}{2pt}

  \ResRow{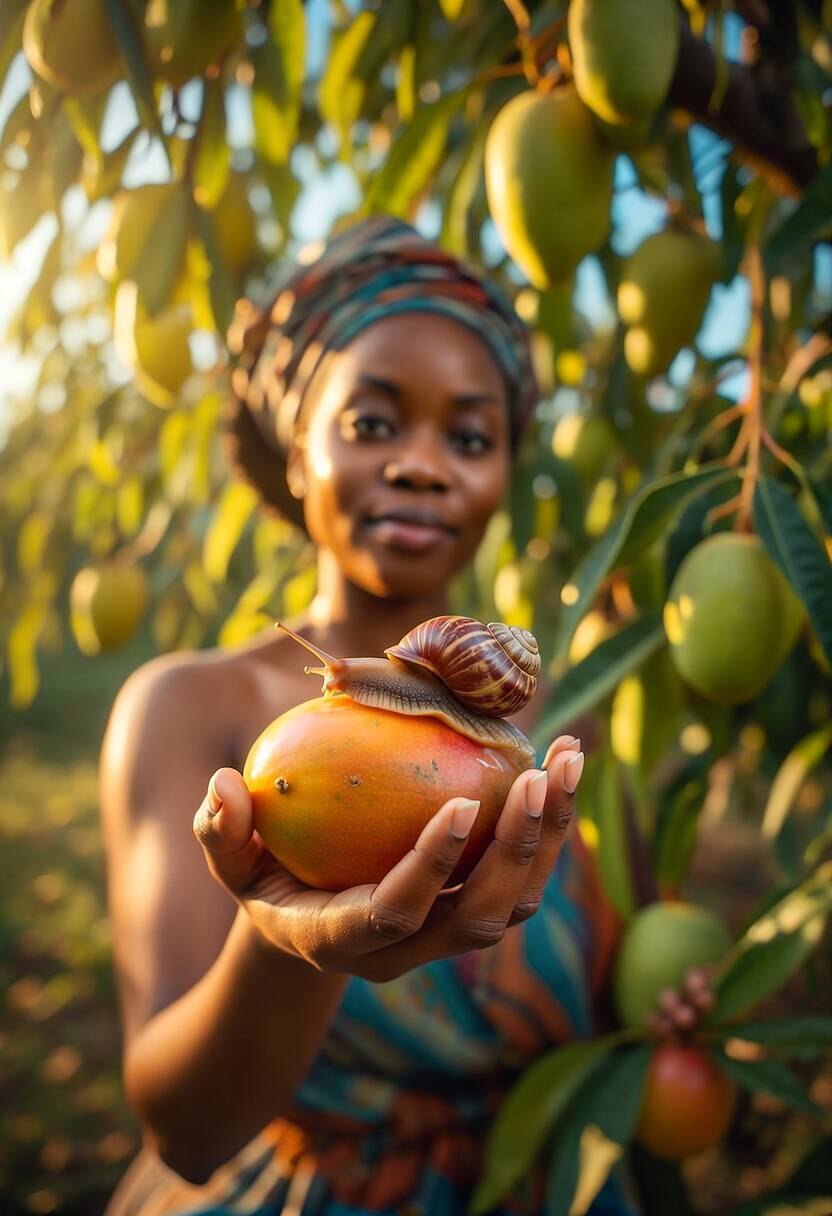}{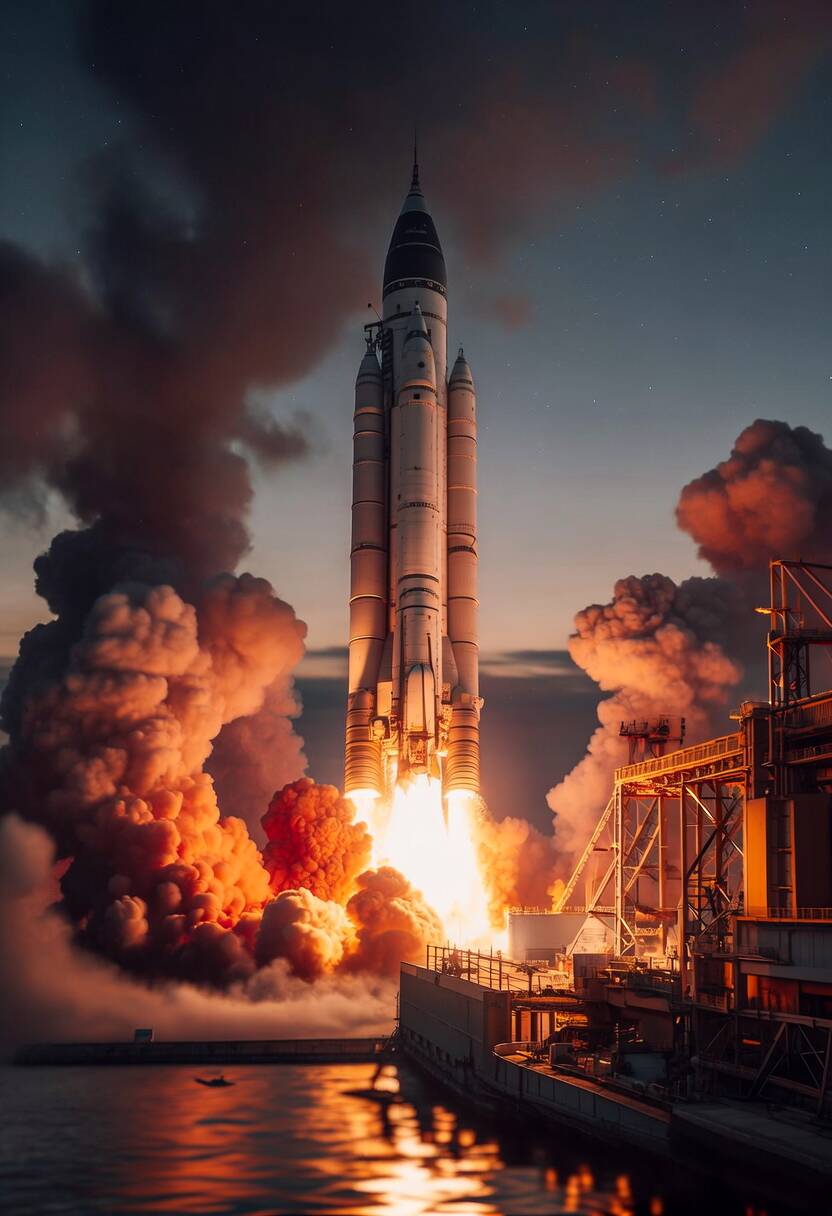}{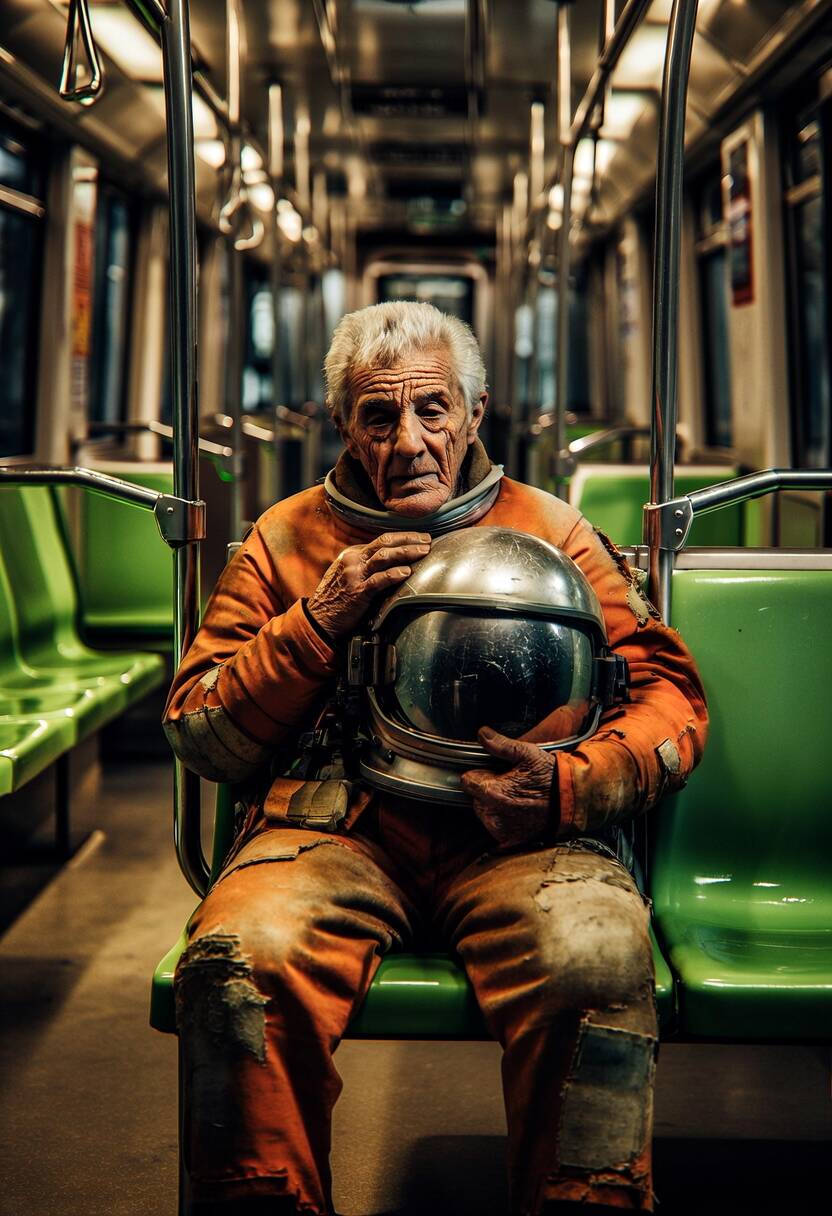}{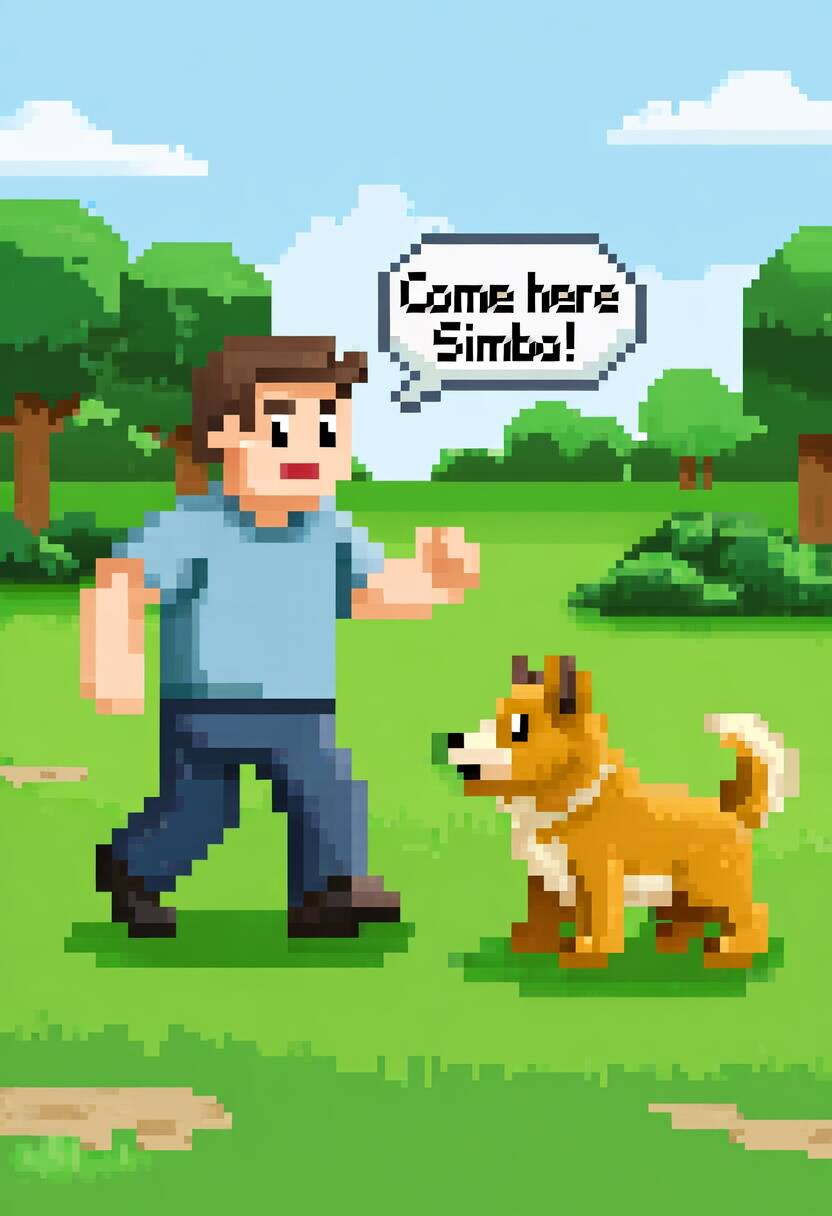}
  \\[8pt]

  \ResRow{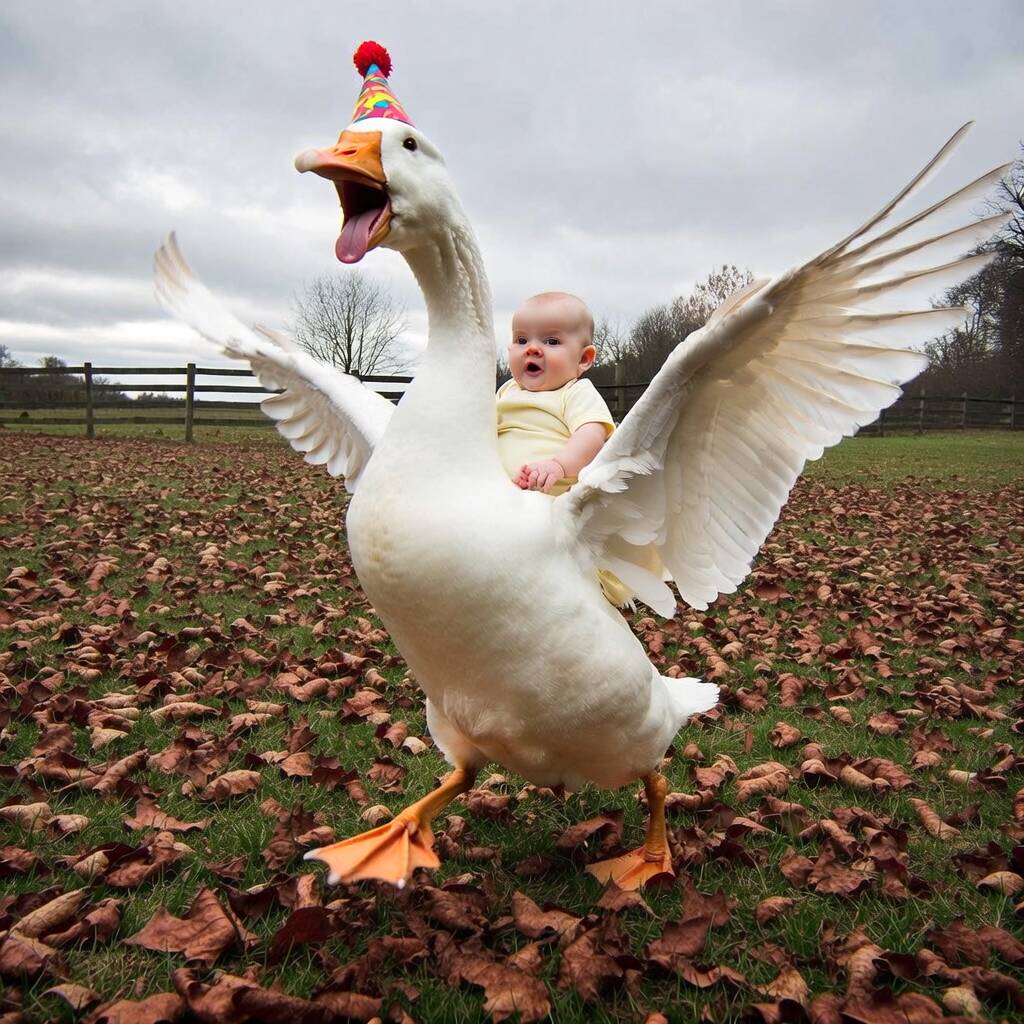}{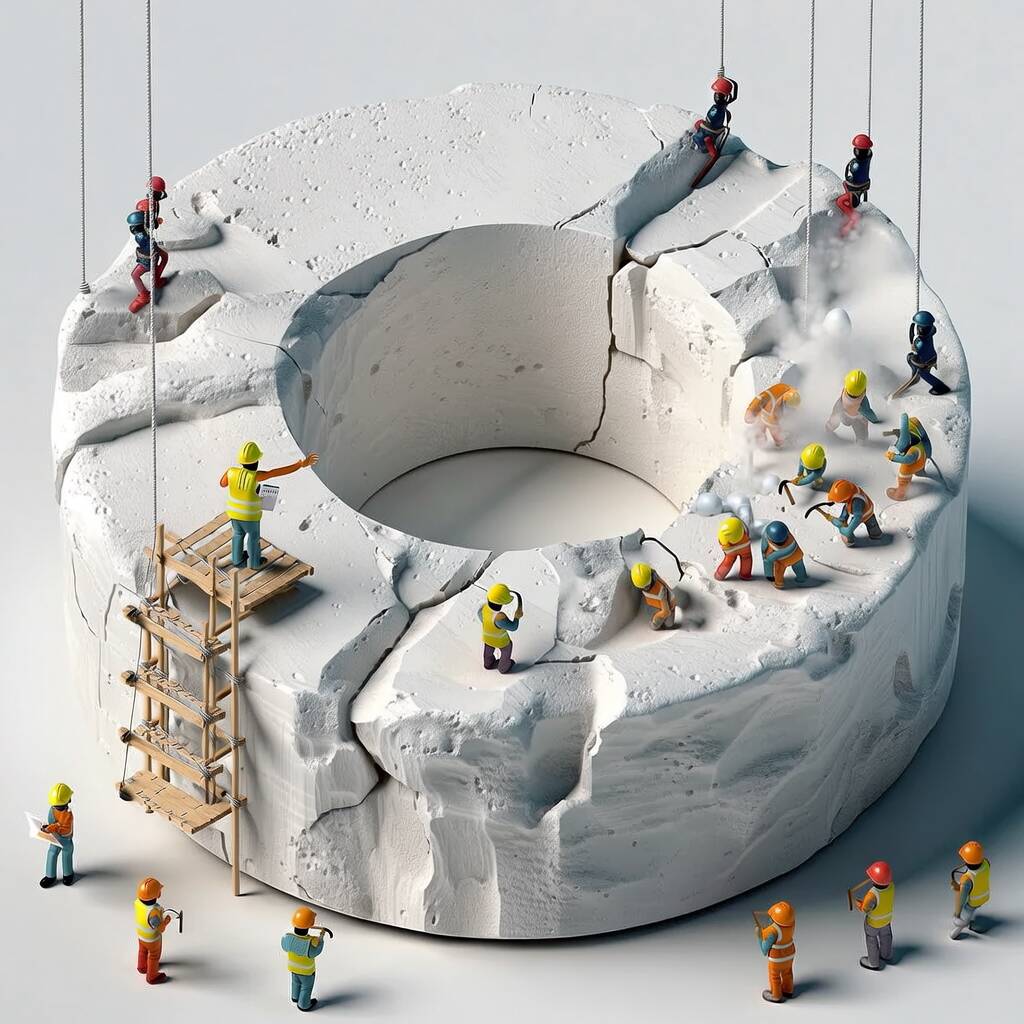}{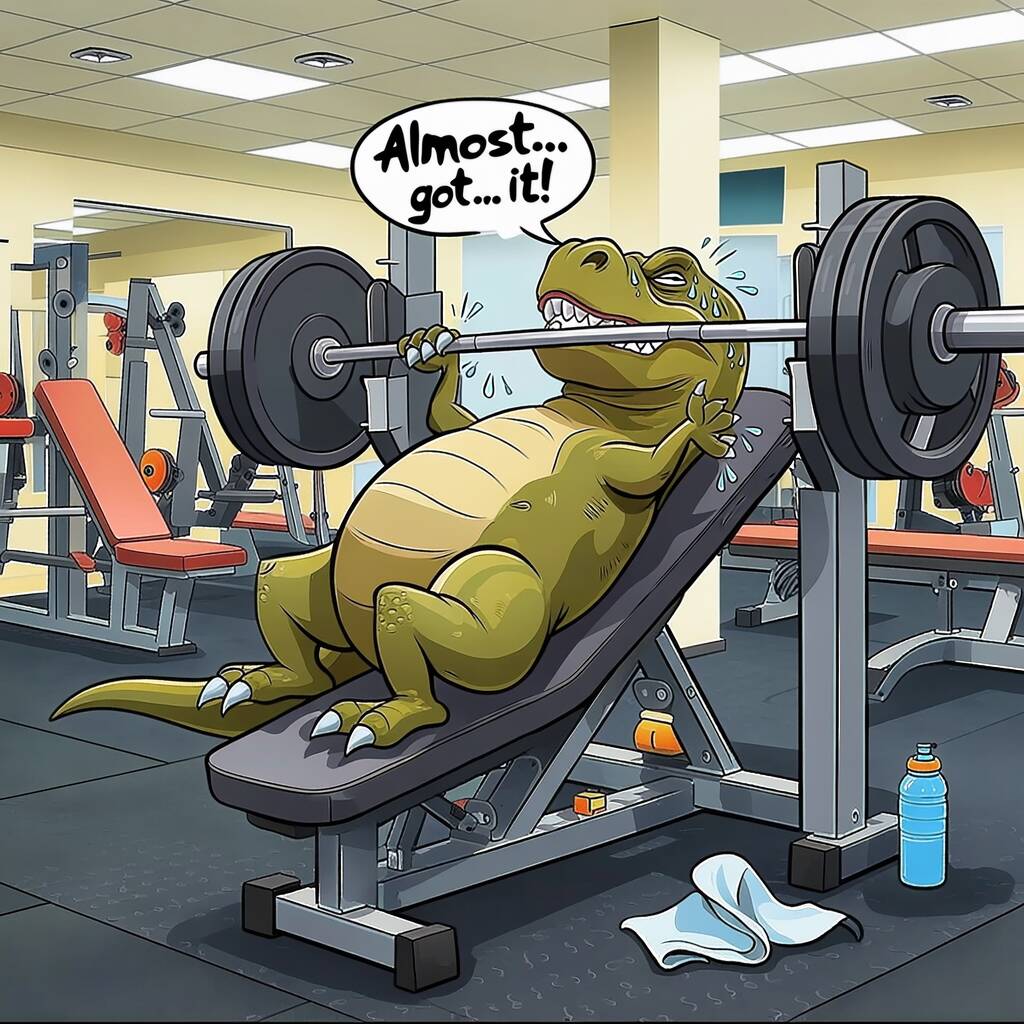}{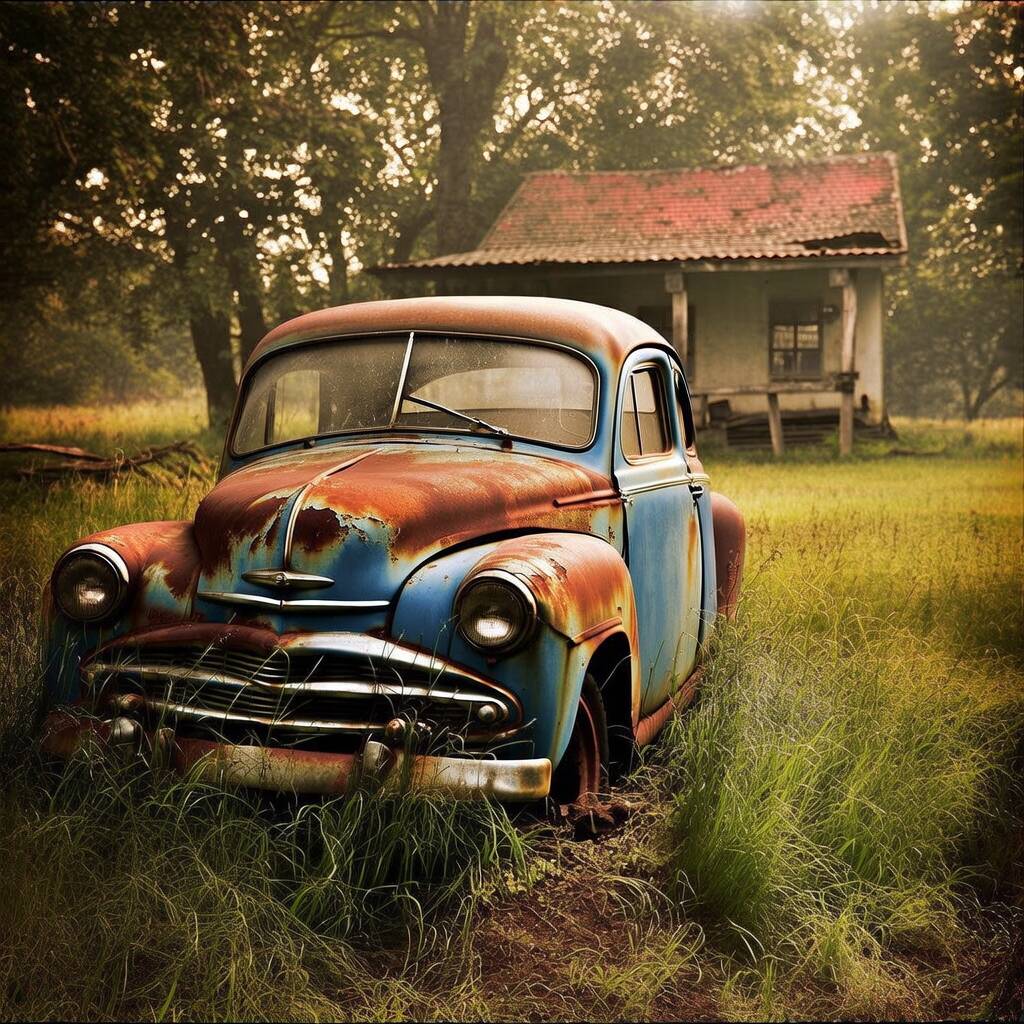}
  \\[8pt]

  \ResRow{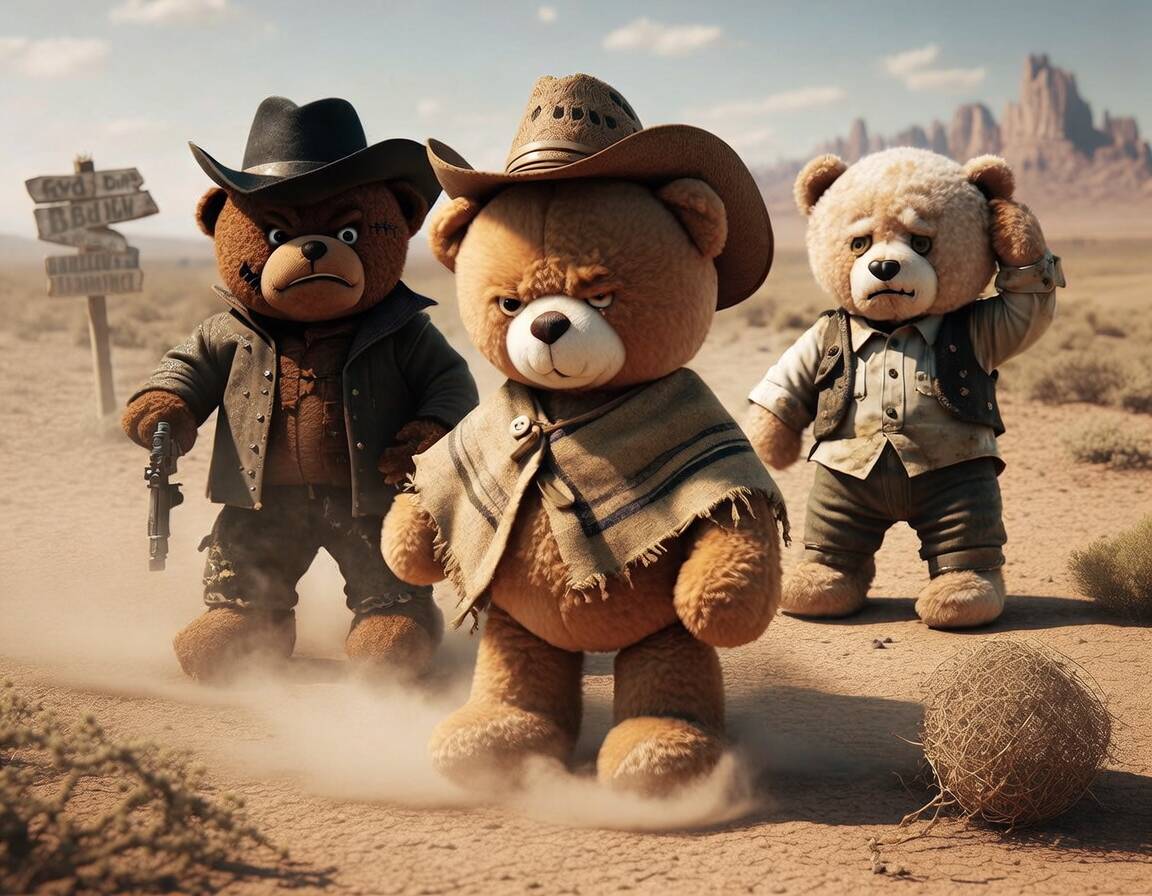}{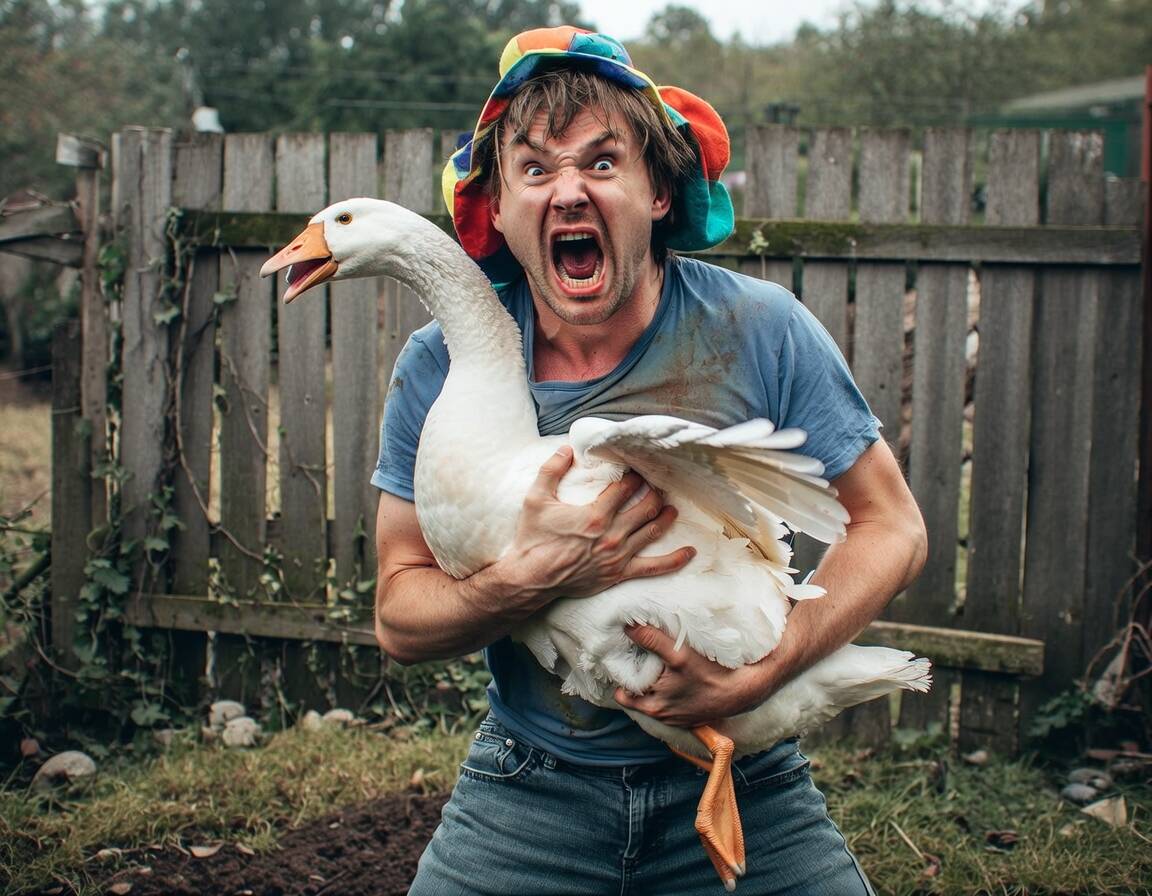}{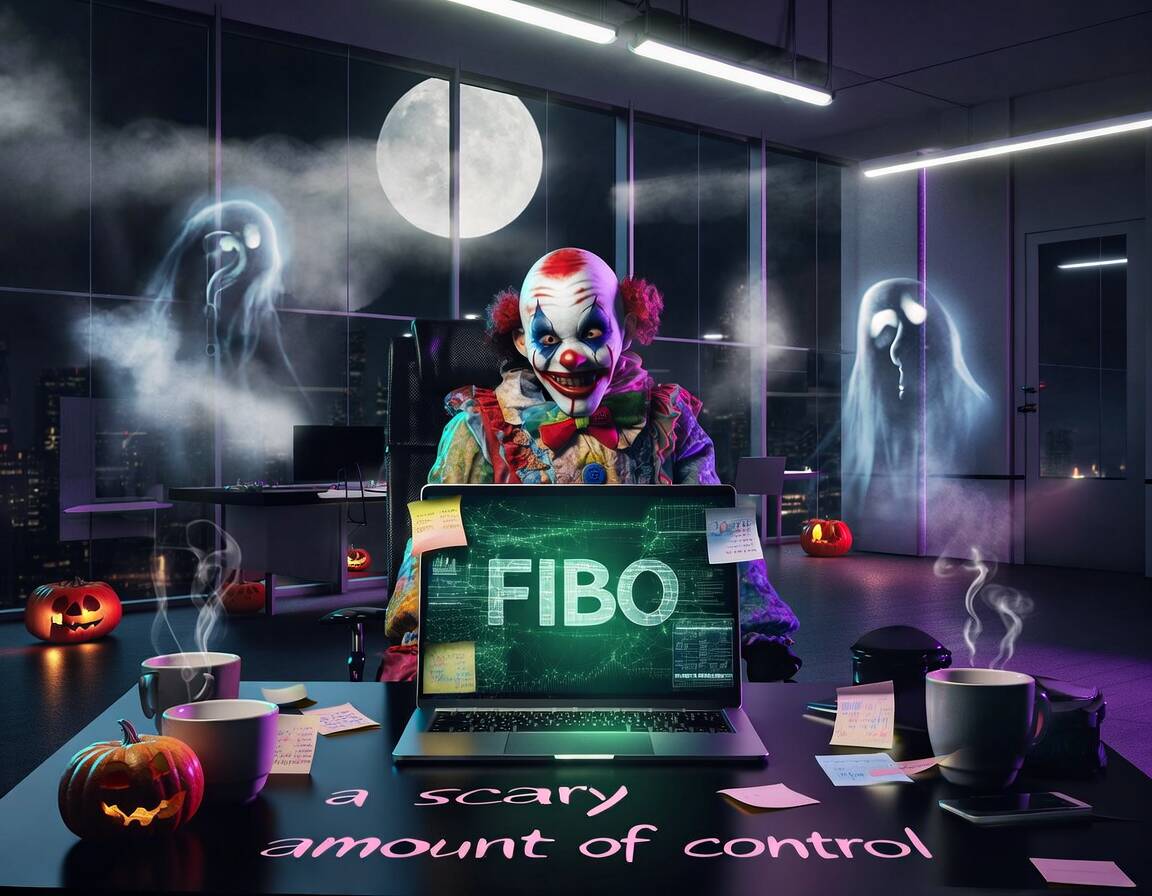}{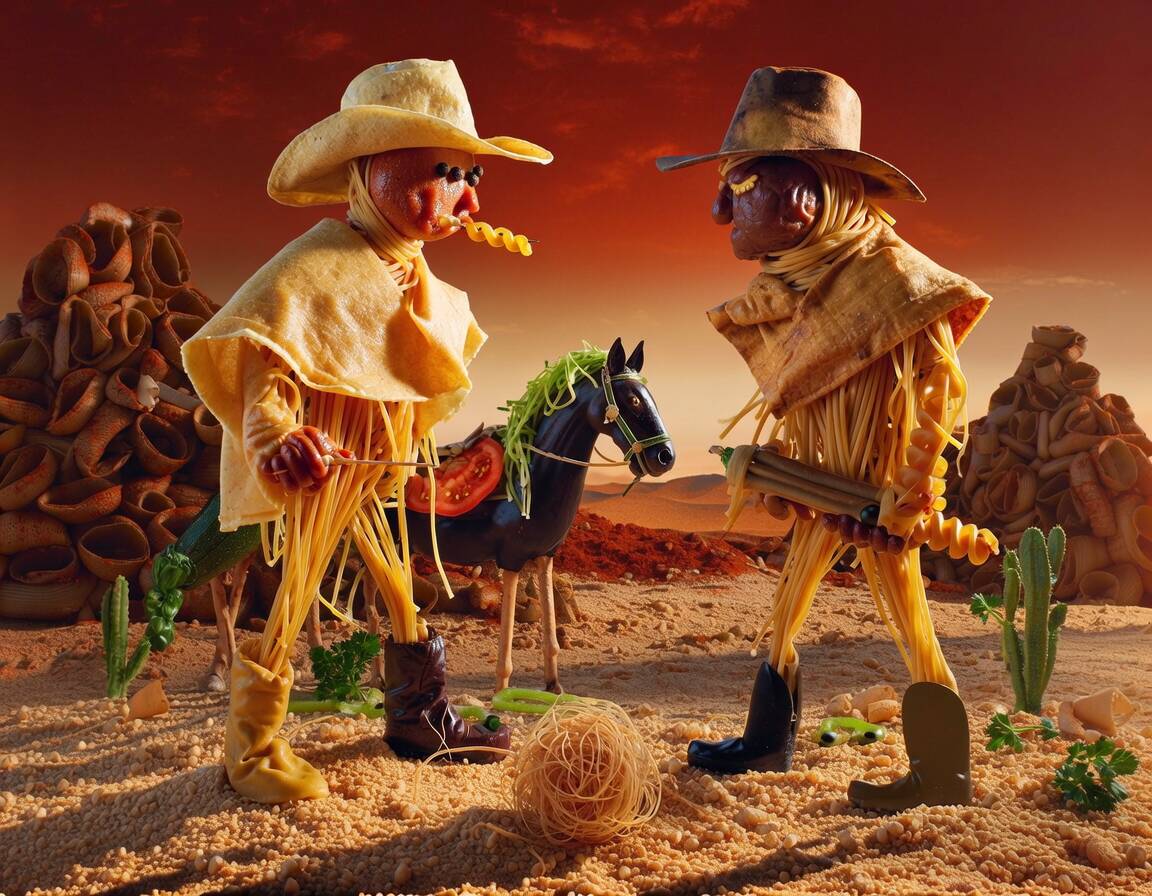}
  \\[8pt]

  \ResRow{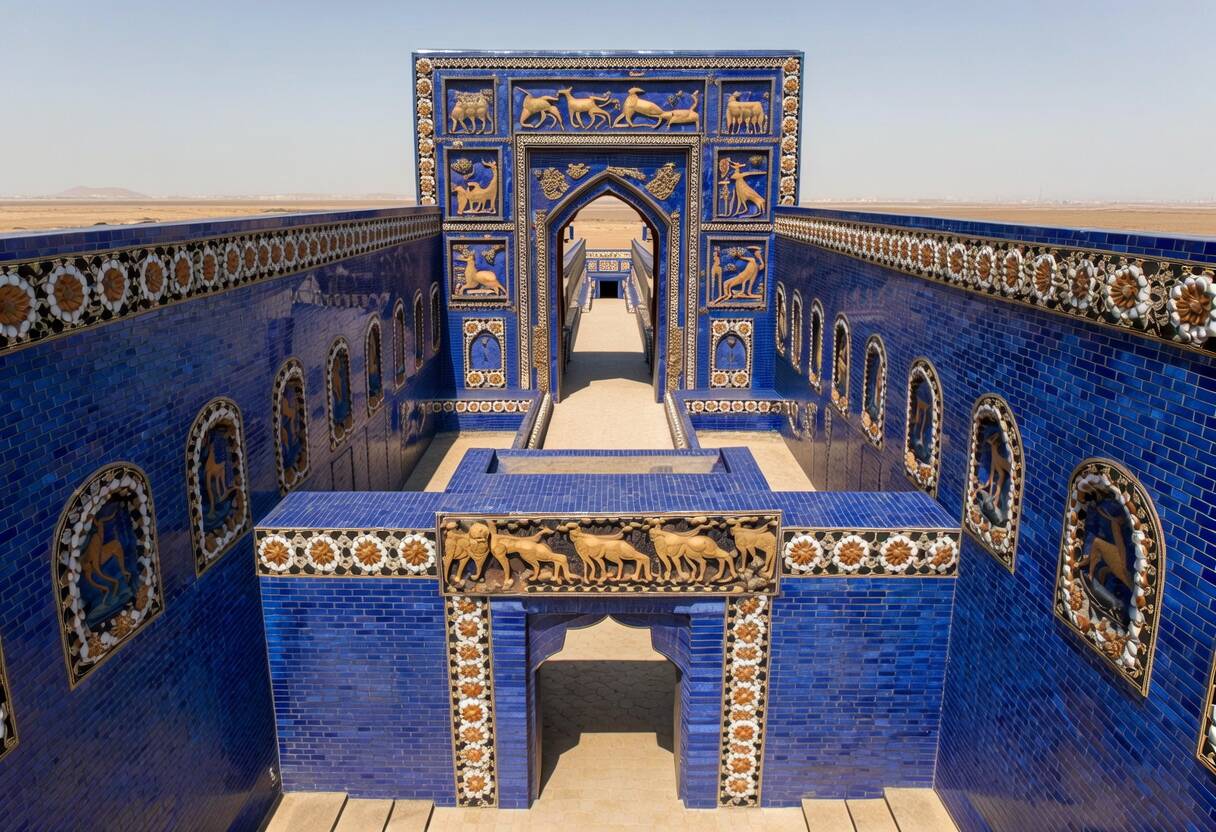}{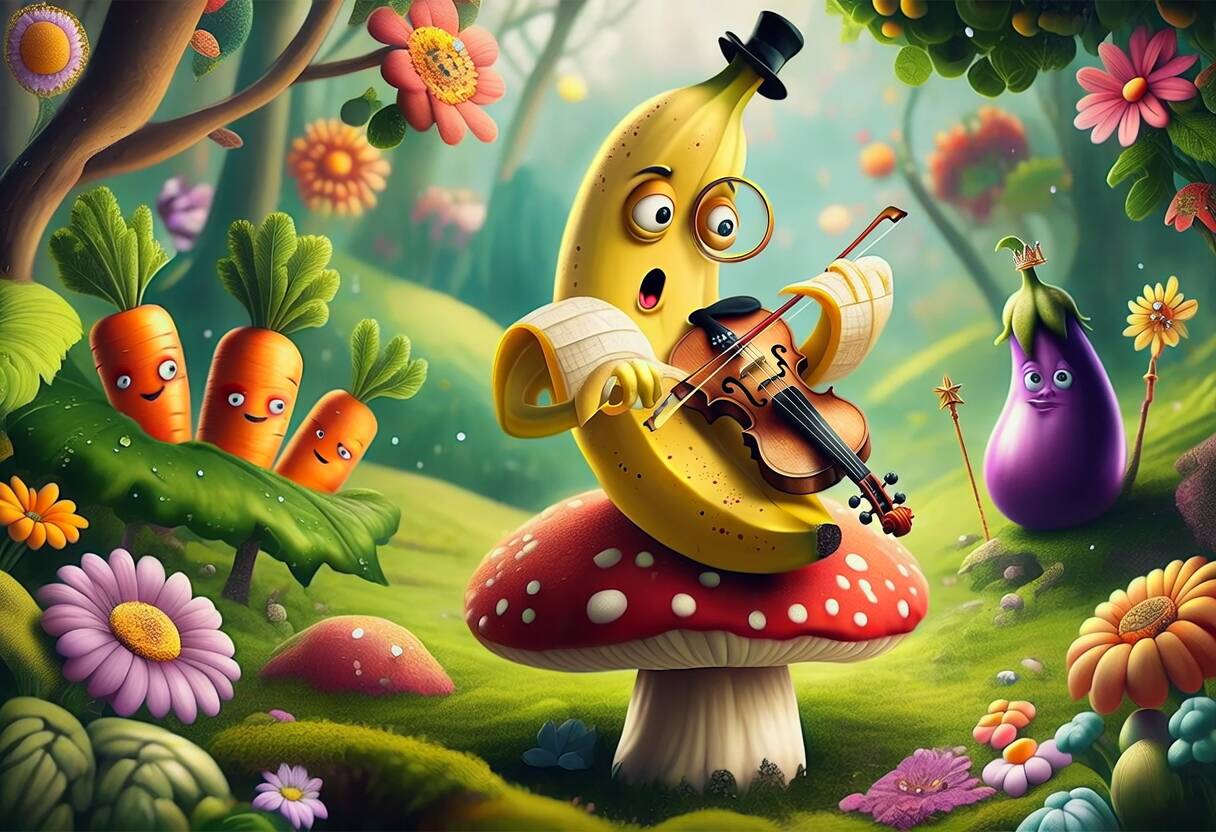}{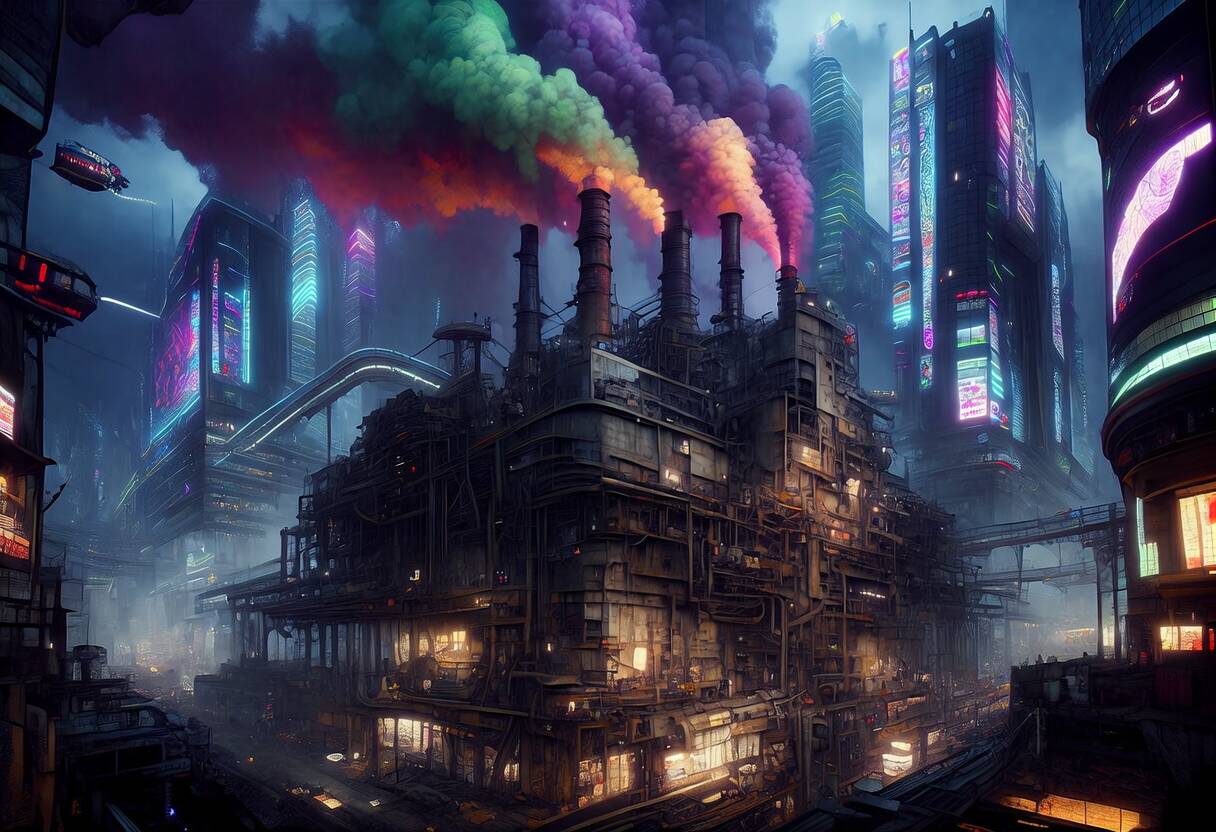}{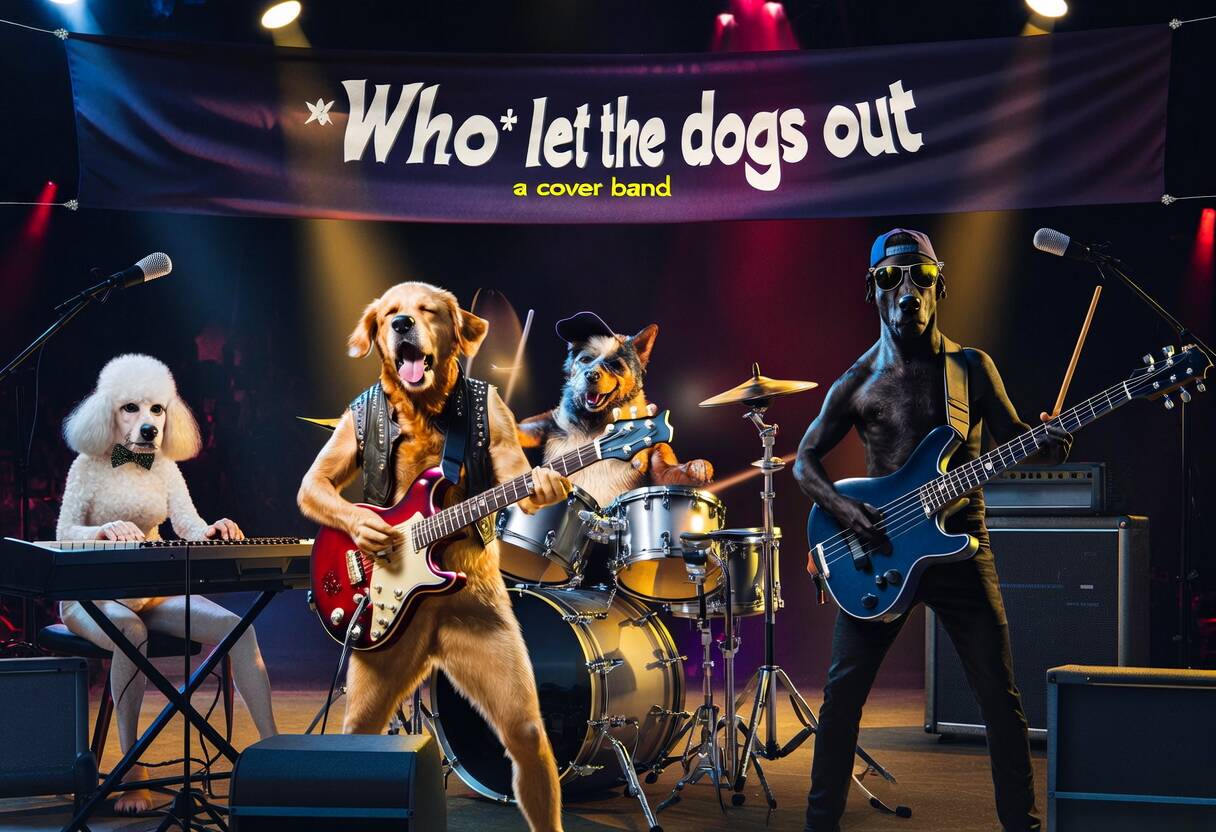}
  \\[8pt]

  \caption{Additional samples from \modelname{} in various aspect ratios.}
  \label{fig:app_samples}
\end{figure*}

%% file: Figures/disentanglement/app_dis_fig.tex
\setlength{\tabcolsep}{0.5pt}
\renewcommand{\arraystretch}{1.0}

\begin{figure*}[t]
\centering
\newcommand{\imgw}{0.24\textwidth} 
\begin{tabular}{@{}cccc@{}}

  \includegraphics[width=\imgw]{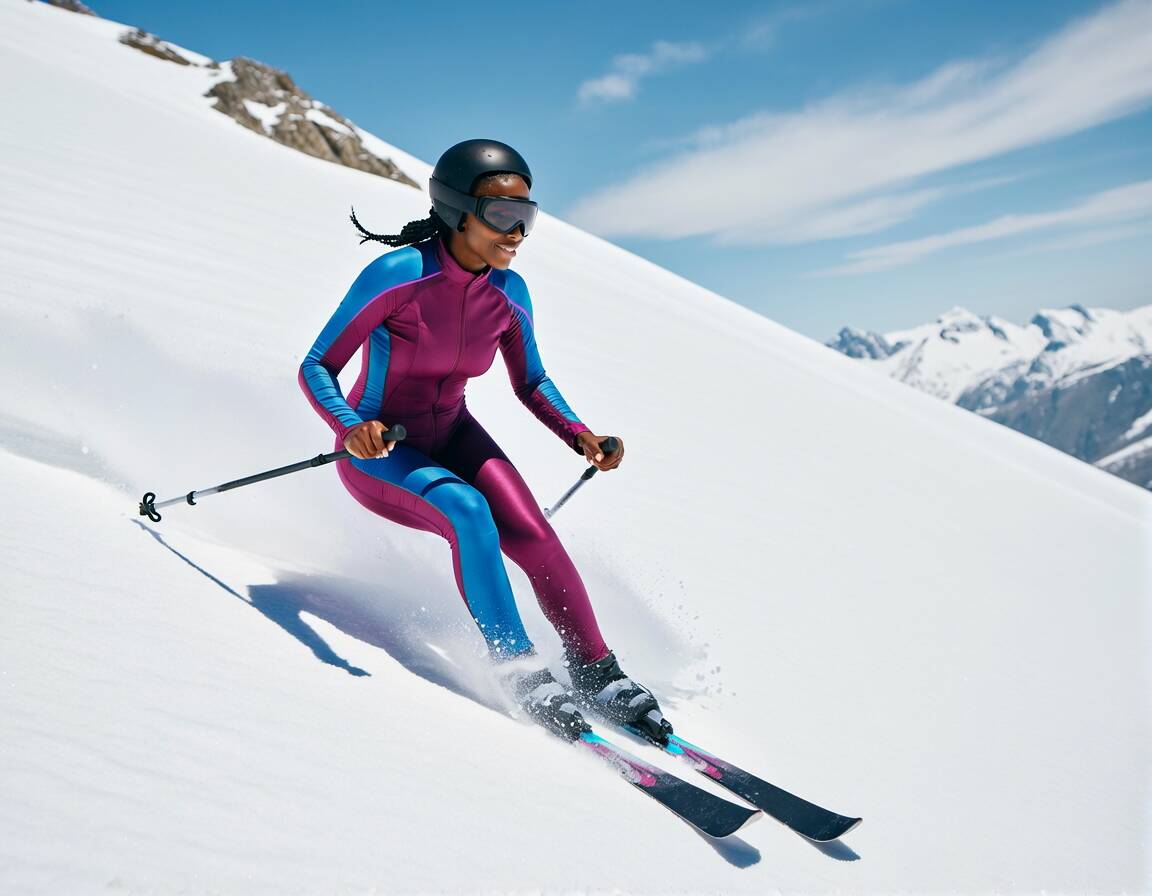} &
  \includegraphics[width=\imgw]{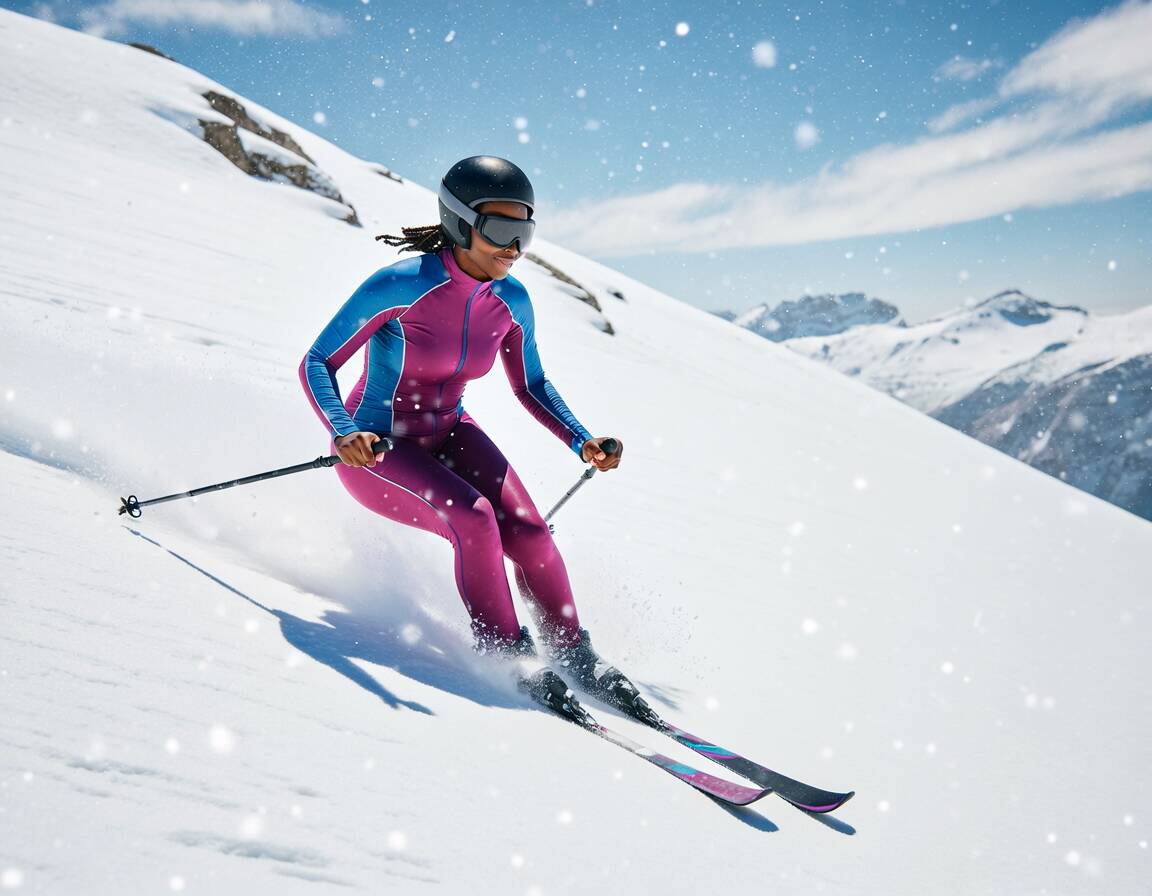} &
  \includegraphics[width=\imgw]{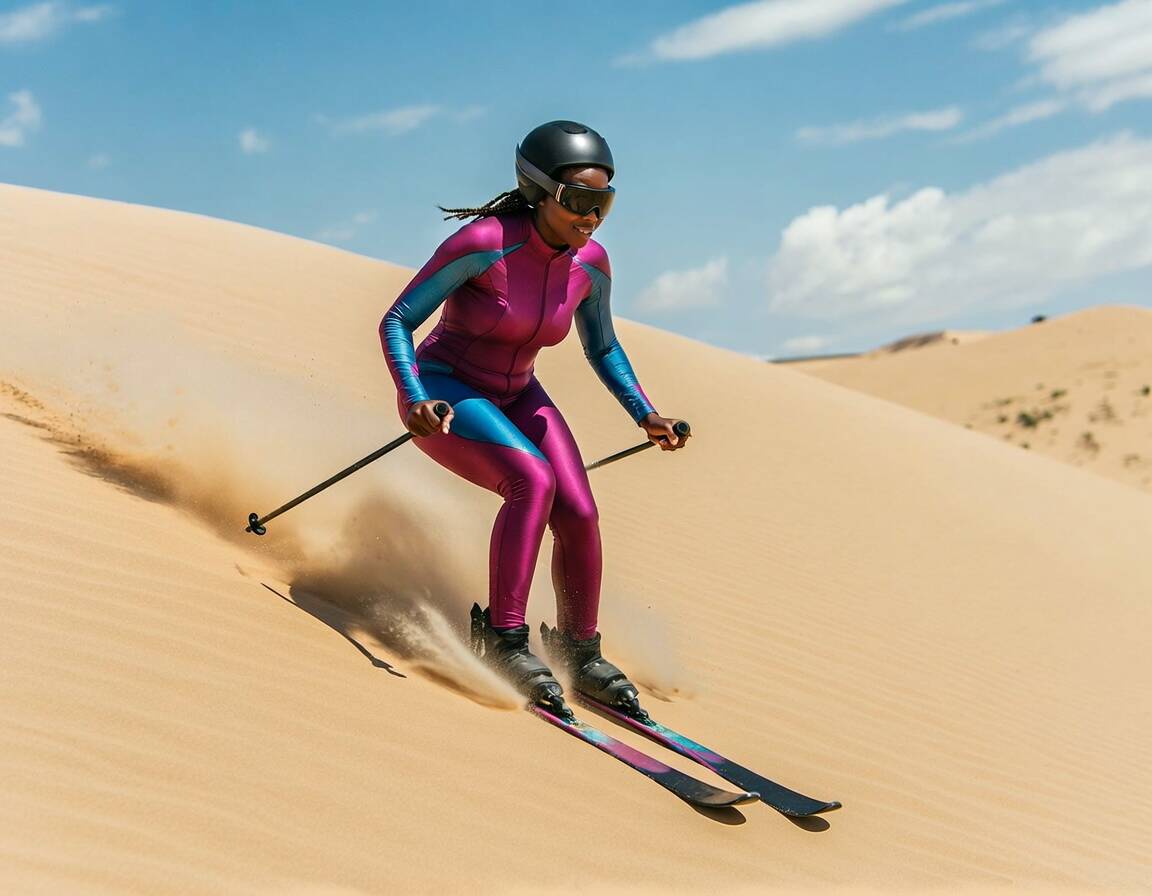} &
  \includegraphics[width=\imgw]{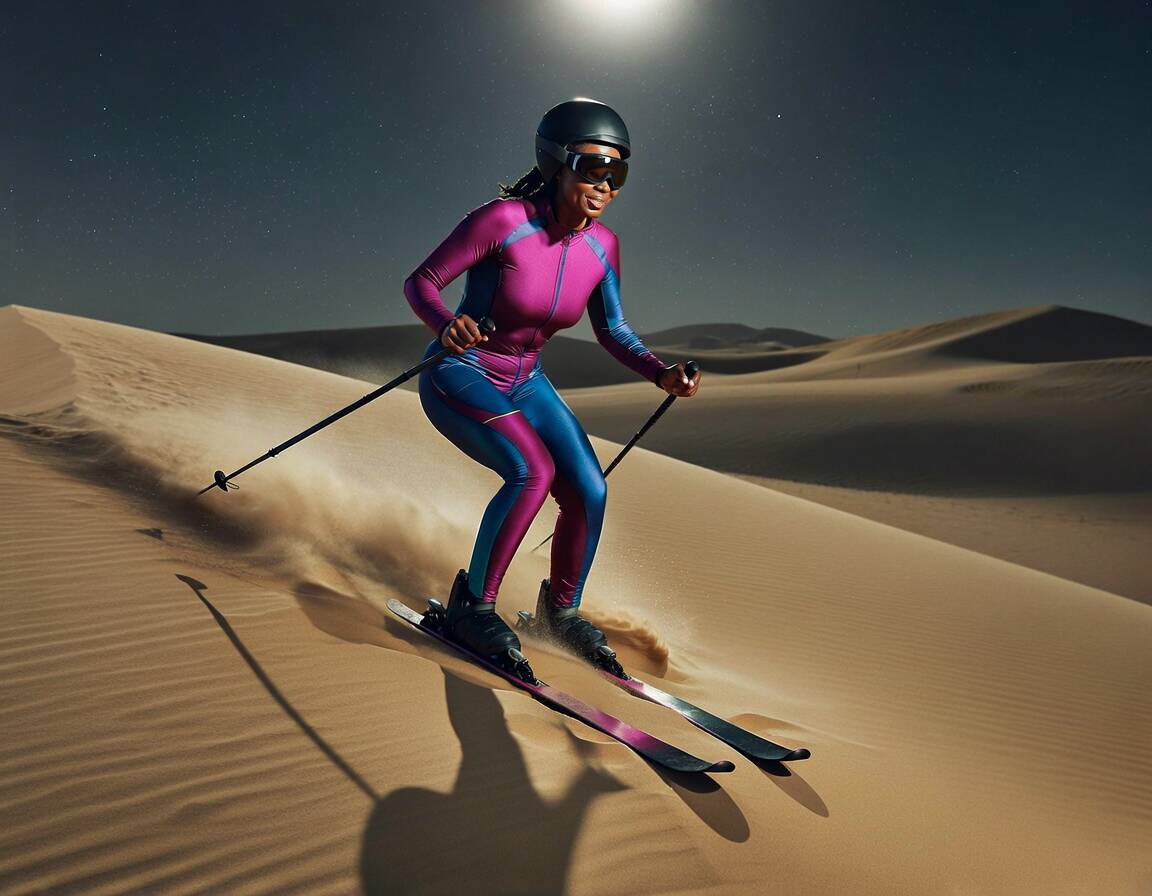}
  \\[-2pt]
  \parbox{\imgw}{\scriptsize\centering \textbf{}} &
  \parbox{\imgw}{\scriptsize\centering \textbf{make snowing}} &
  \parbox{\imgw}{\scriptsize\centering \textbf{move to desert}} &
  \parbox{\imgw}{\scriptsize\centering \textbf{make night}}
  \\[+2pt]

  \includegraphics[width=\imgw]{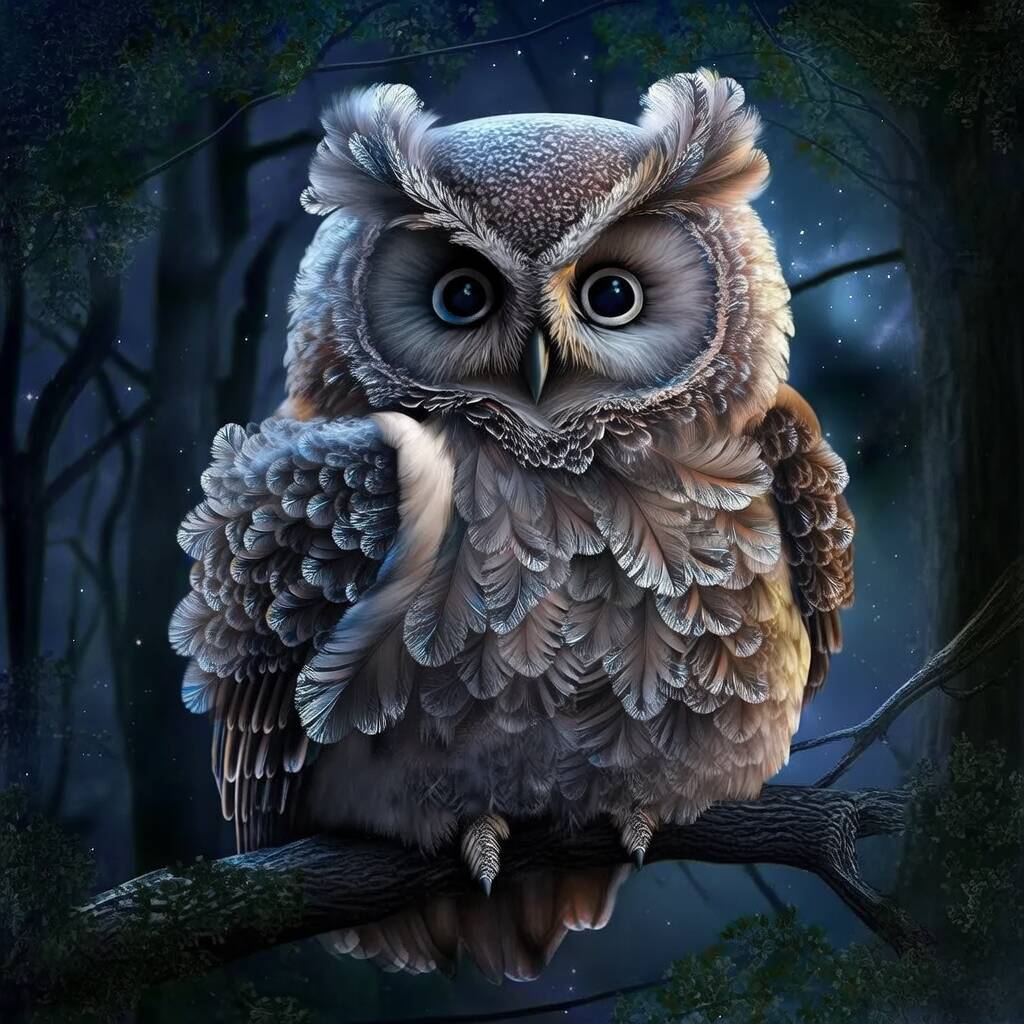} &
  \includegraphics[width=\imgw]{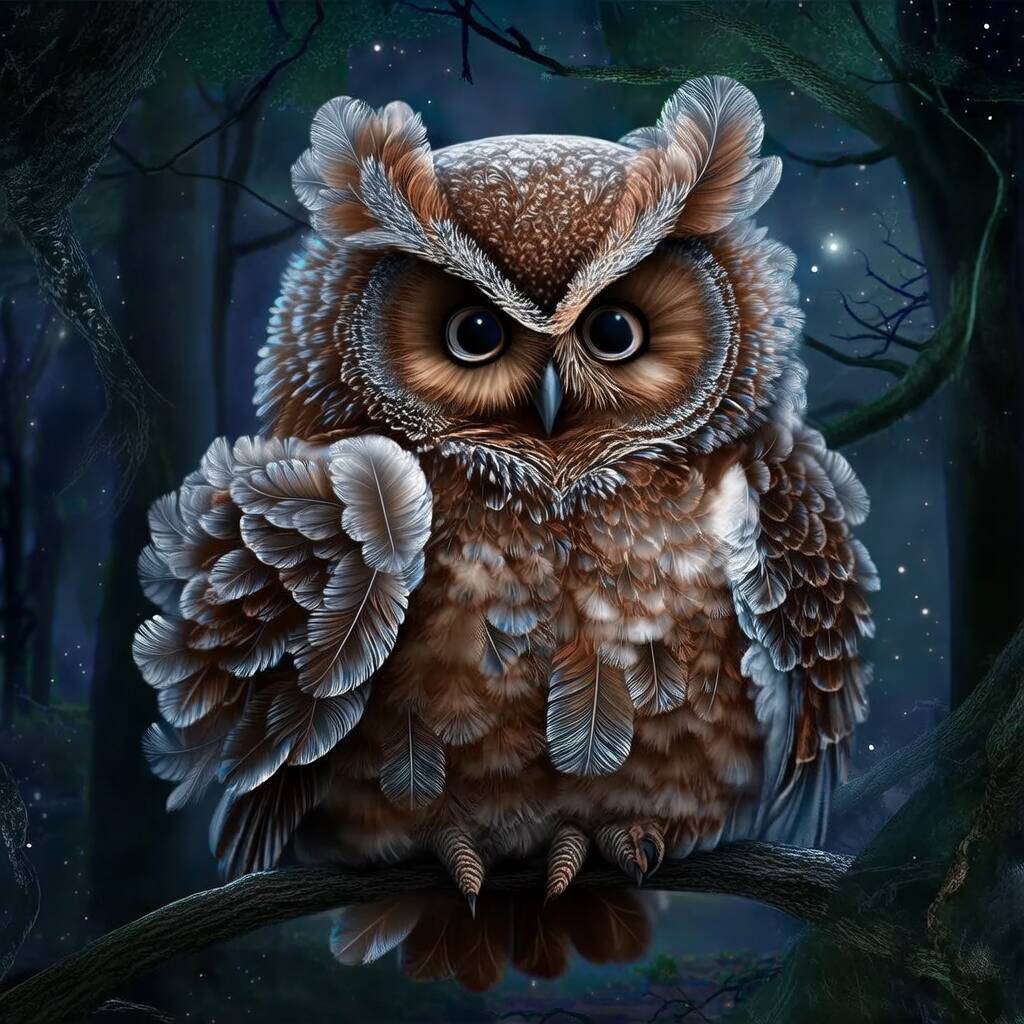} &
  \includegraphics[width=\imgw]{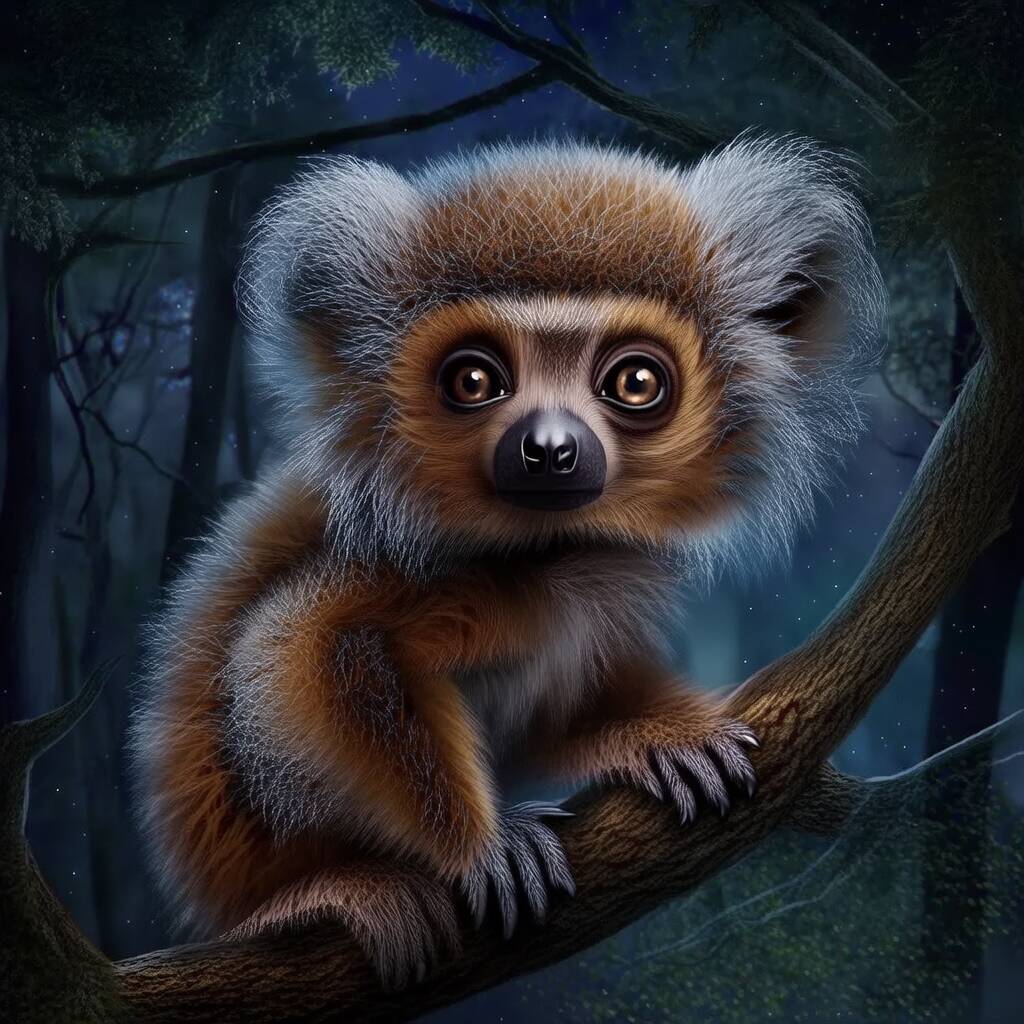} &
  \includegraphics[width=\imgw]{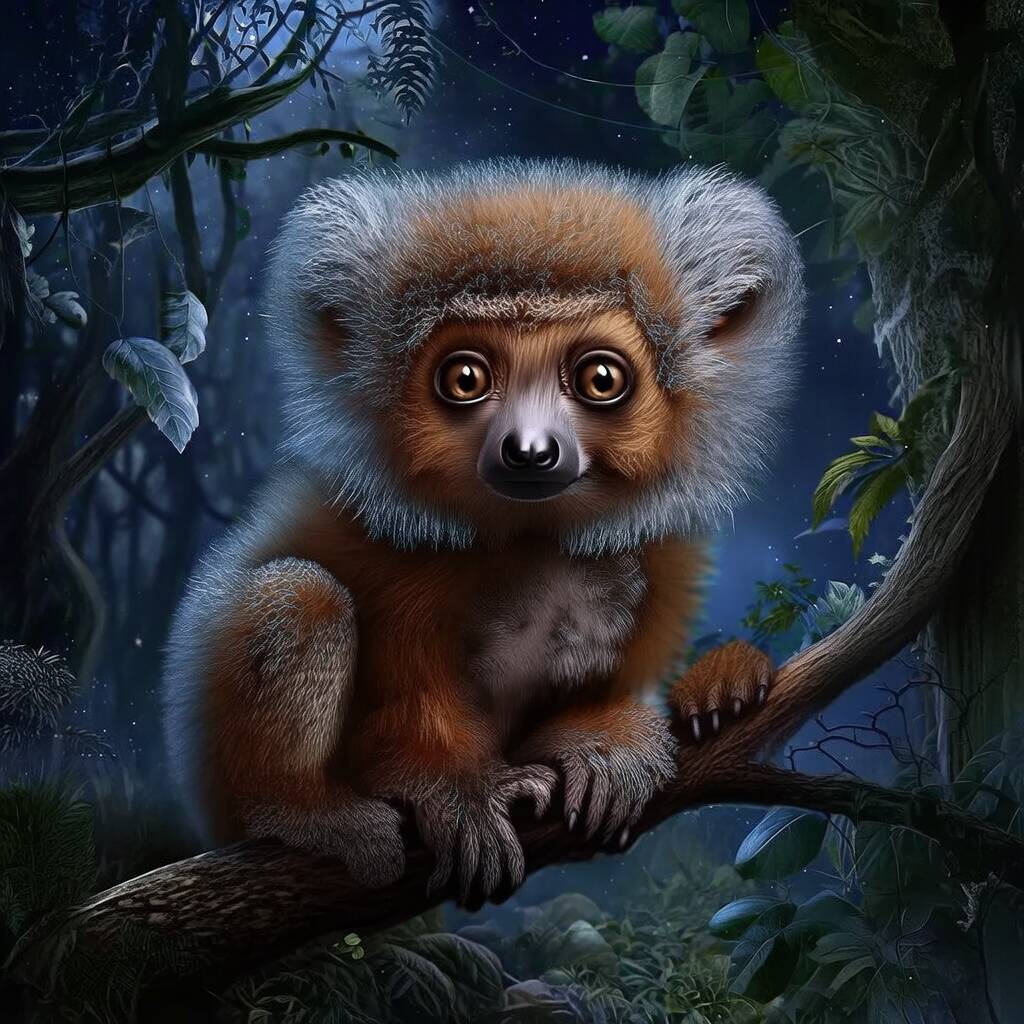}
  \\[-2pt]

  \parbox{\imgw}{\scriptsize\centering \textbf{}} &
  \parbox{\imgw}{\scriptsize\centering \textbf{make owl brown}} &
  \parbox{\imgw}{\scriptsize\centering \textbf{owl to lemur}} &
  \parbox{\imgw}{\scriptsize\centering \textbf{add jungle vegetation}}
  \\[+2pt]

    \includegraphics[width=\imgw]{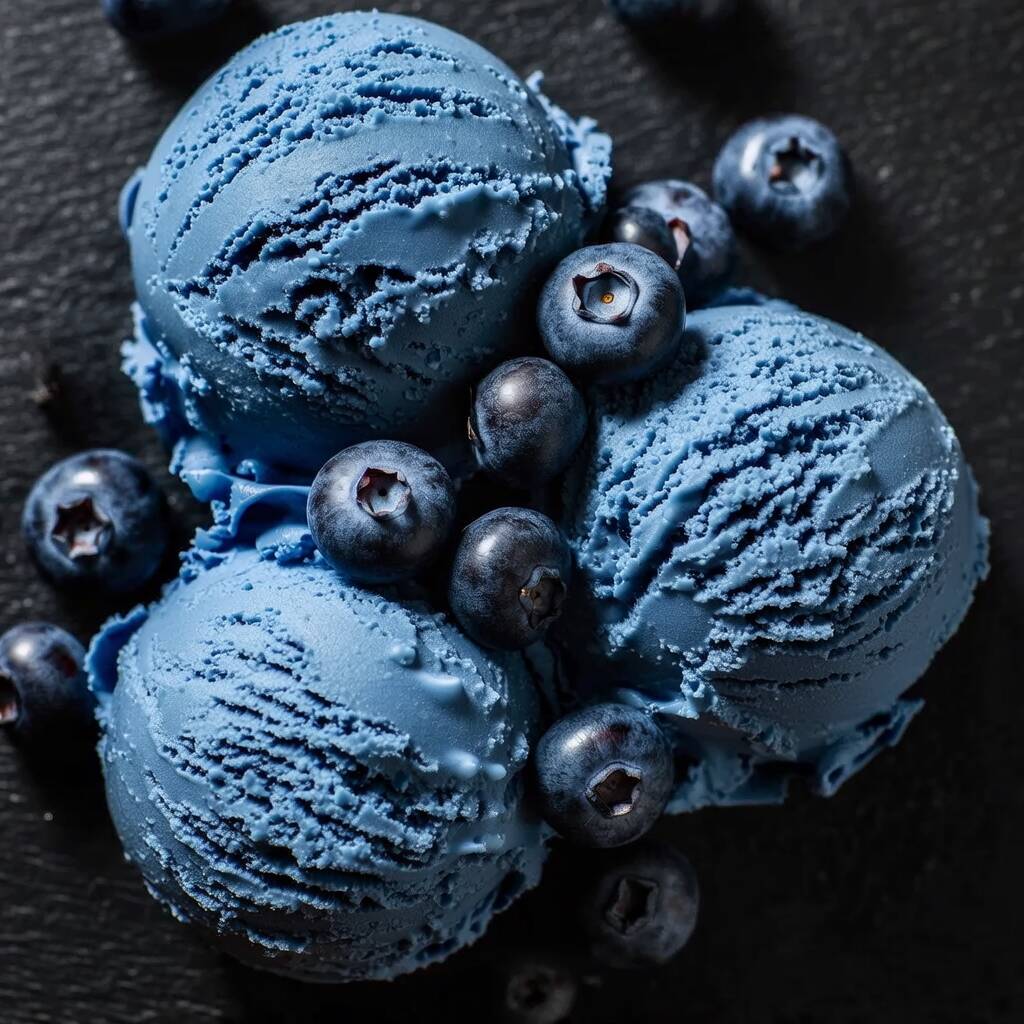} &
  \includegraphics[width=\imgw]{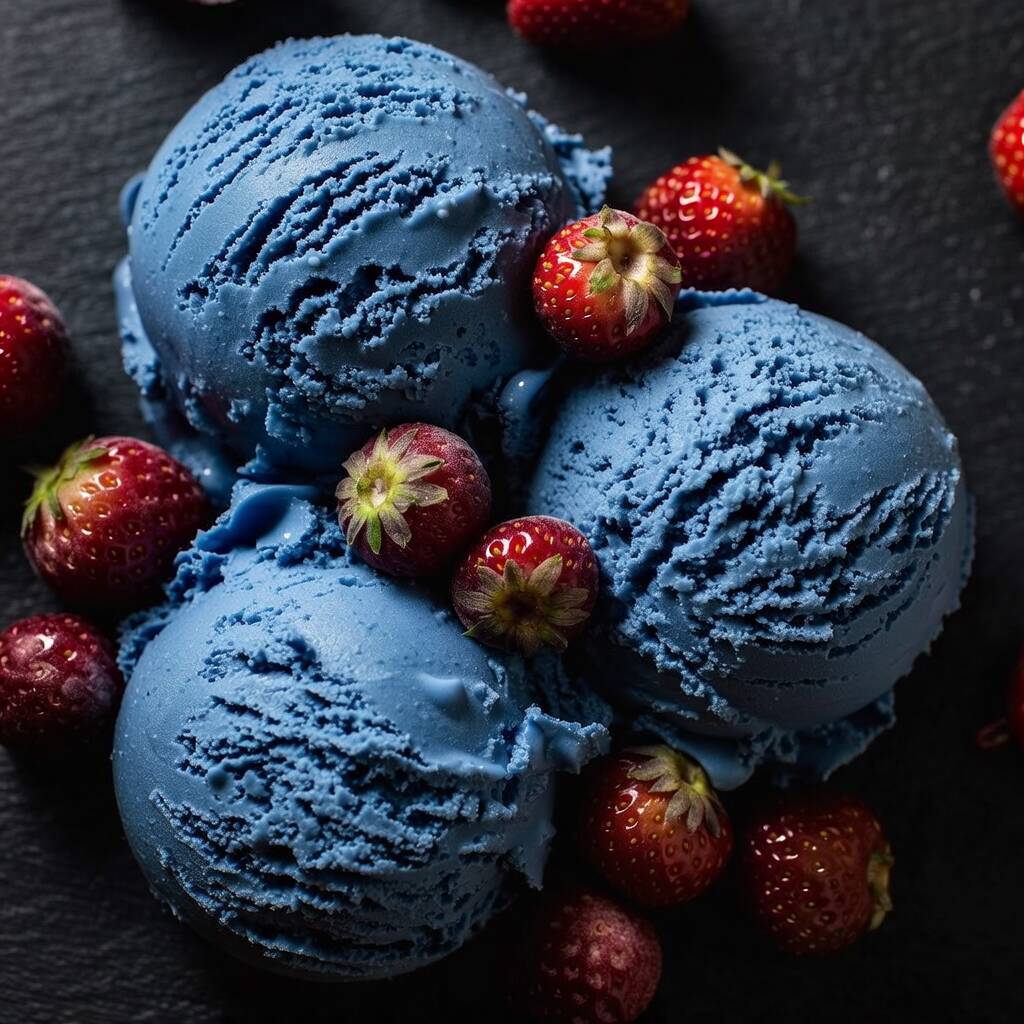} &
  \includegraphics[width=\imgw]{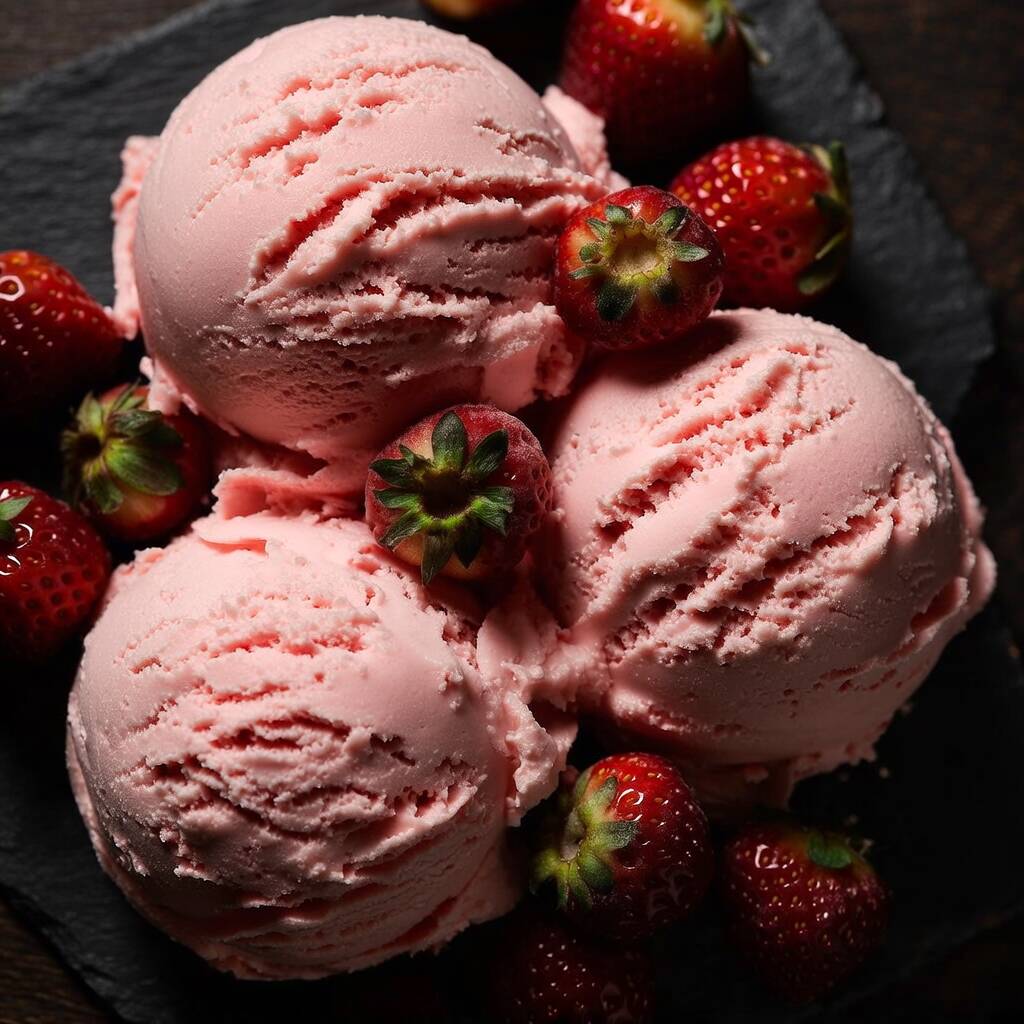} &
  \includegraphics[width=\imgw]{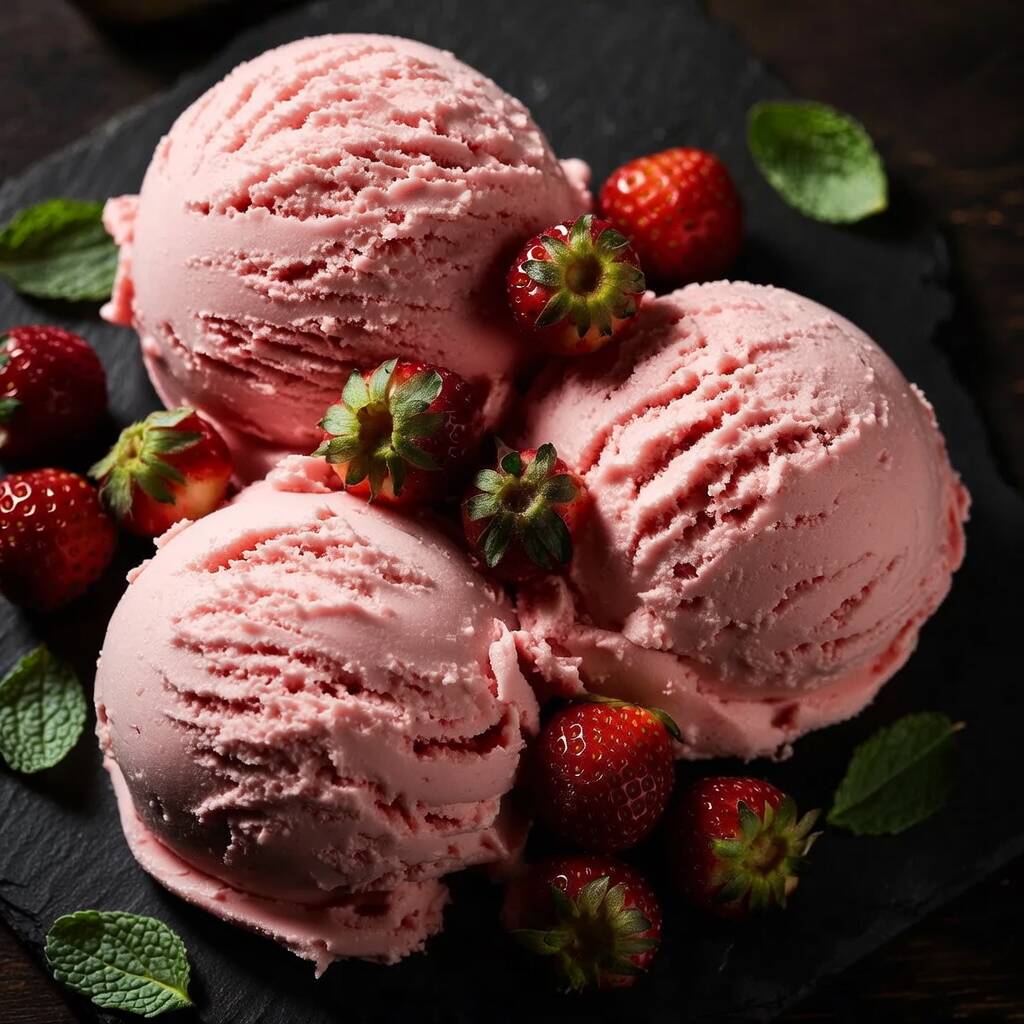}
  \\[-2pt]

  \parbox{\imgw}{\scriptsize\centering \textbf{}} &
  \parbox{\imgw}{\scriptsize\centering \textbf{blueberries to strawberries}} &
  \parbox{\imgw}{\scriptsize\centering \textbf{blue to pink}} &
  \parbox{\imgw}{\scriptsize\centering \textbf{add mint leaves}}
  \\[+2pt]

  \includegraphics[width=\imgw]{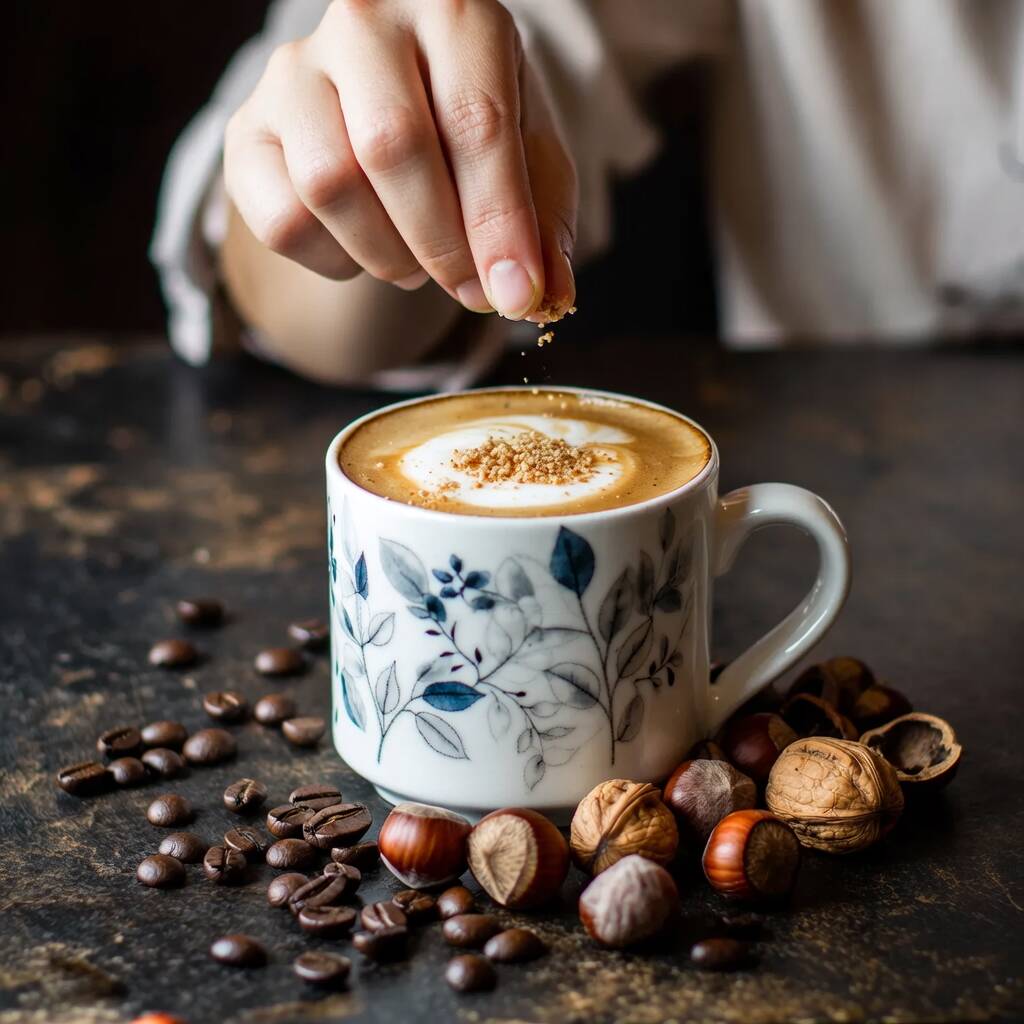} &
  \includegraphics[width=\imgw]{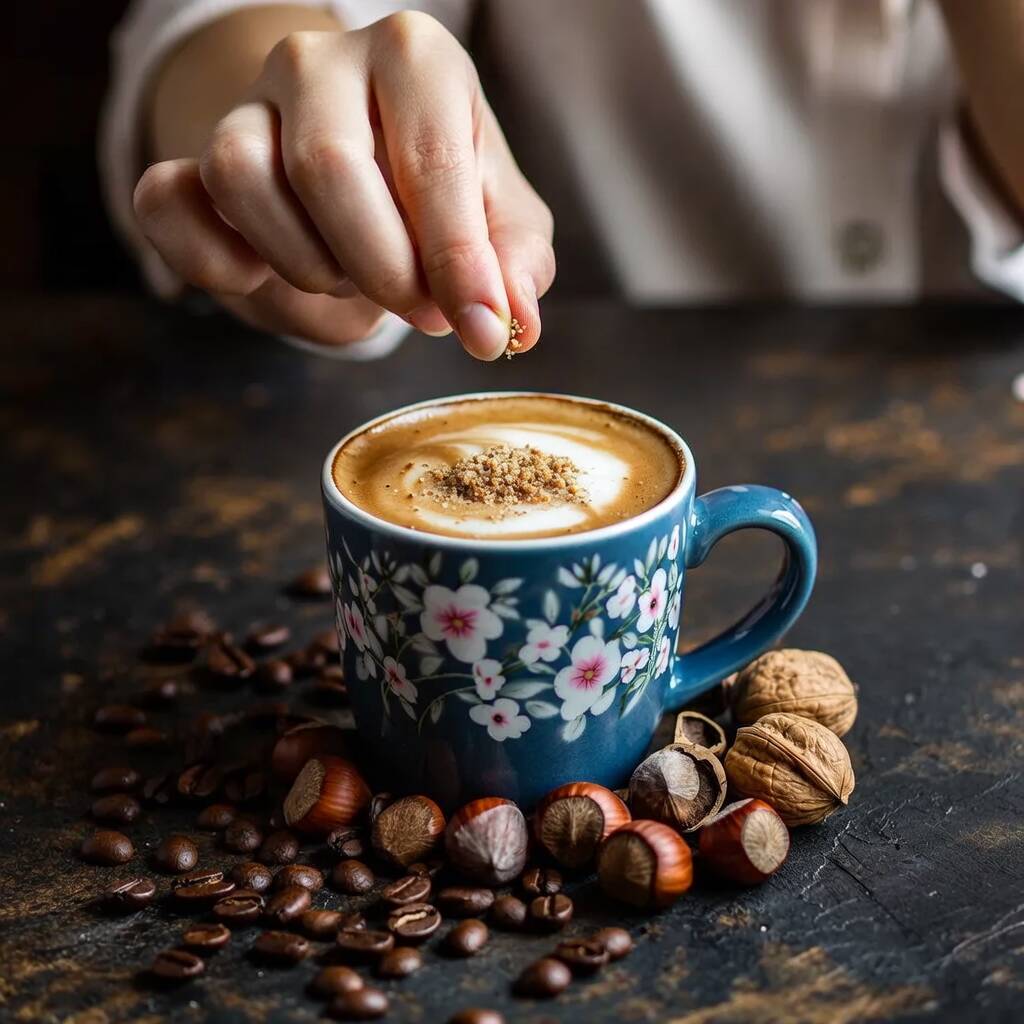} &
  \includegraphics[width=\imgw]{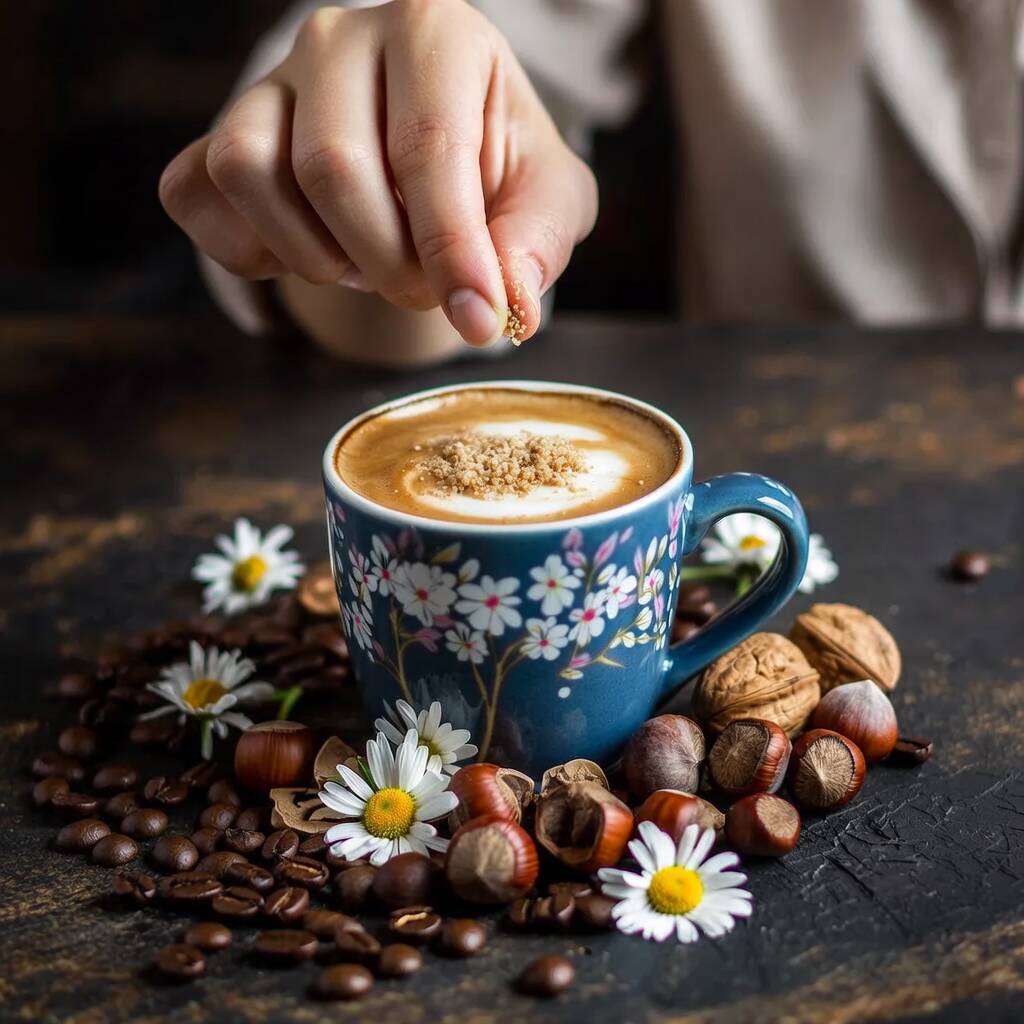} &
  \includegraphics[width=\imgw]{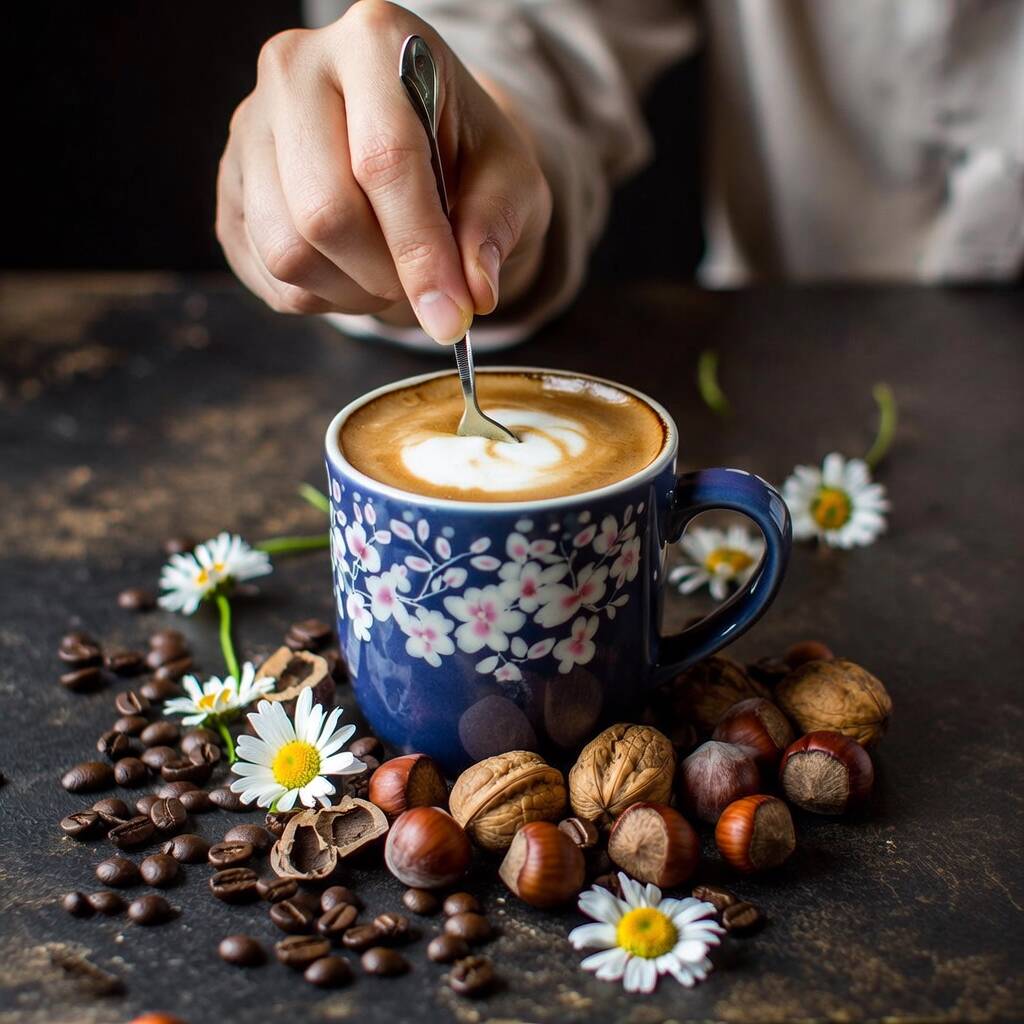}
  \\[-2pt]

  \parbox{\imgw}{\scriptsize\centering \textbf{}} &
  \parbox{\imgw}{\scriptsize\centering \textbf{blue cup with flower ornaments}} &
  \parbox{\imgw}{\scriptsize\centering \textbf{add scattered daisies}} &
  \parbox{\imgw}{\scriptsize\centering \textbf{hand steering coffee with teaspoon}}
  \\[+2pt]

    \includegraphics[width=\imgw]{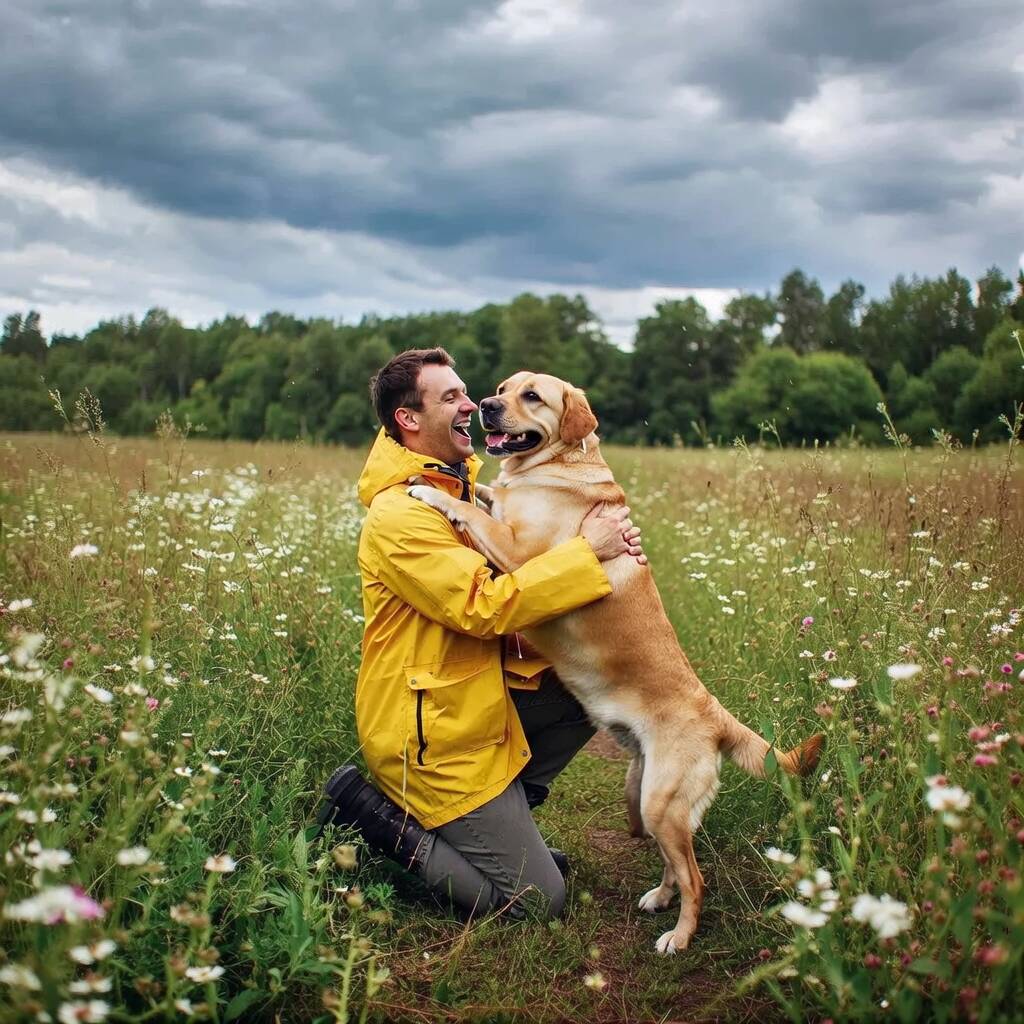} &
  \includegraphics[width=\imgw]{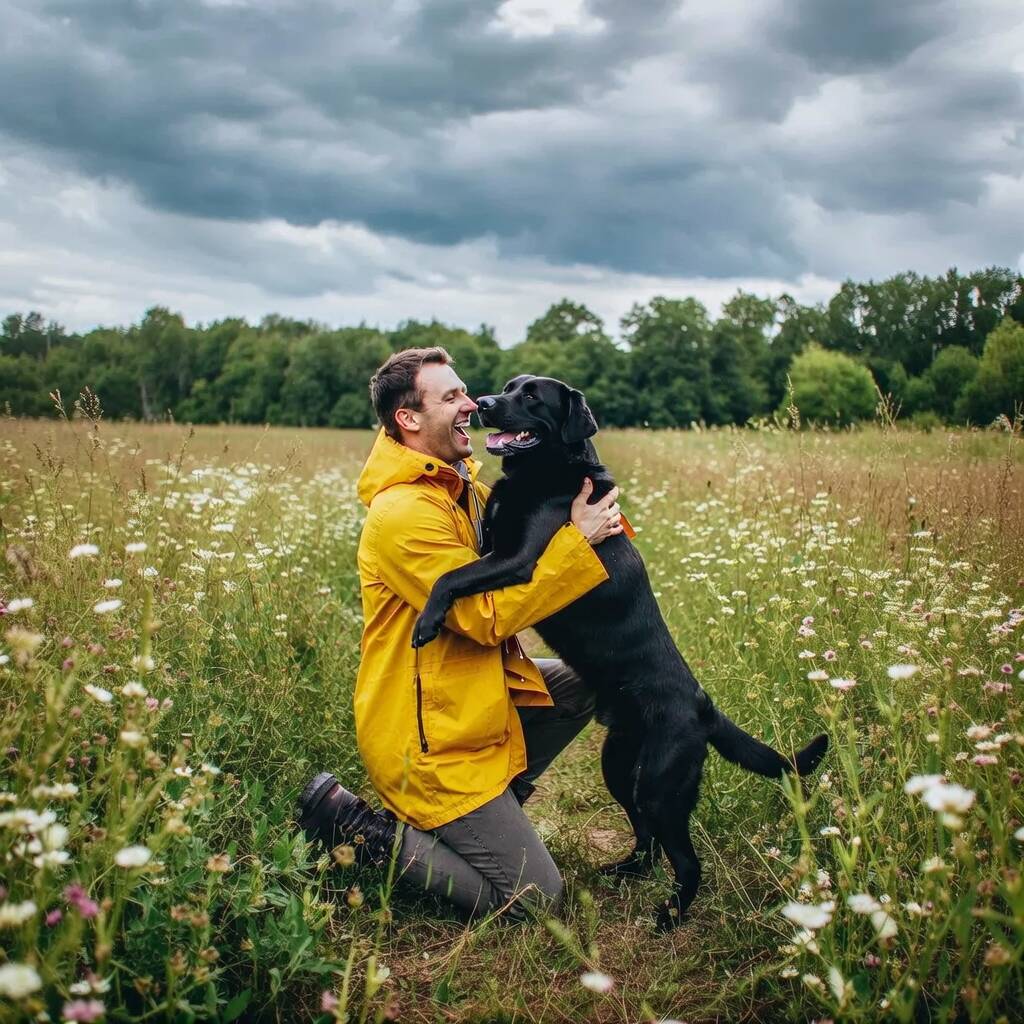} &
  \includegraphics[width=\imgw]{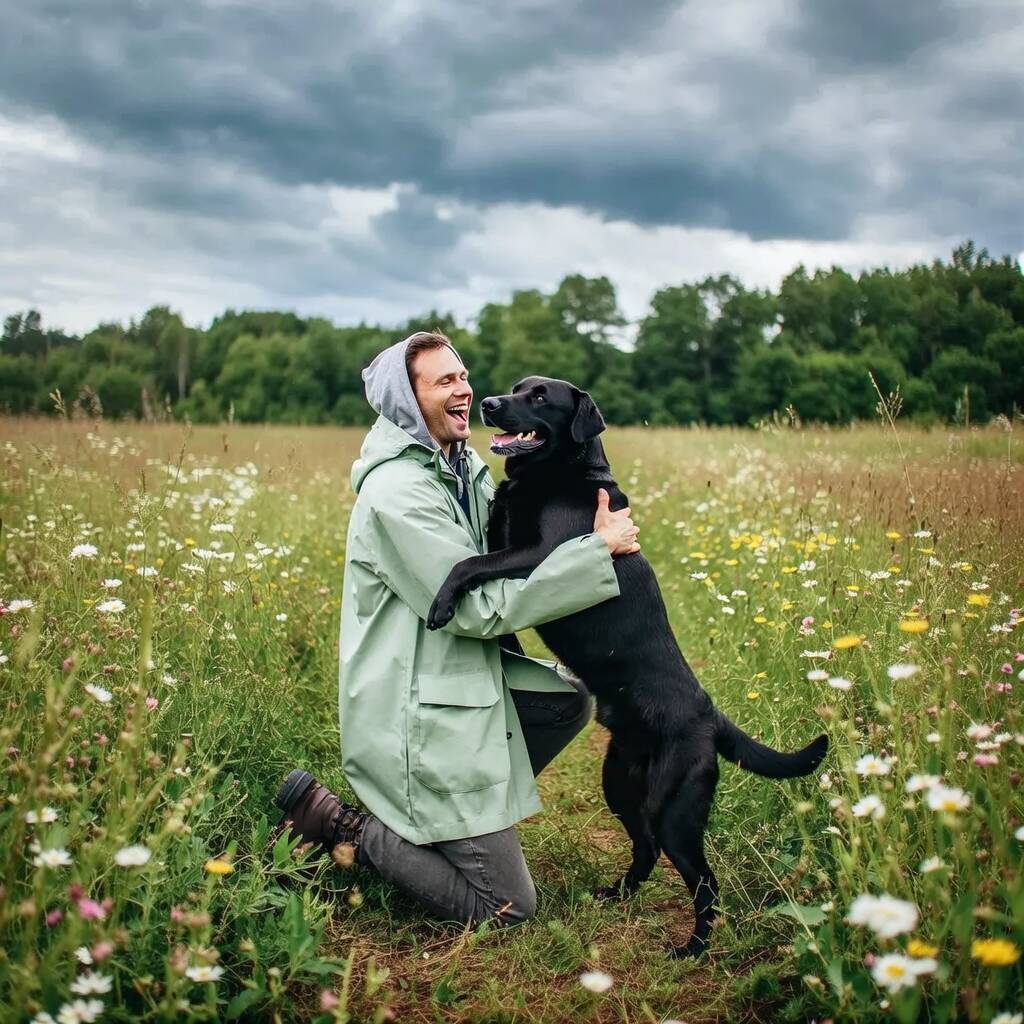} &
  \includegraphics[width=\imgw]{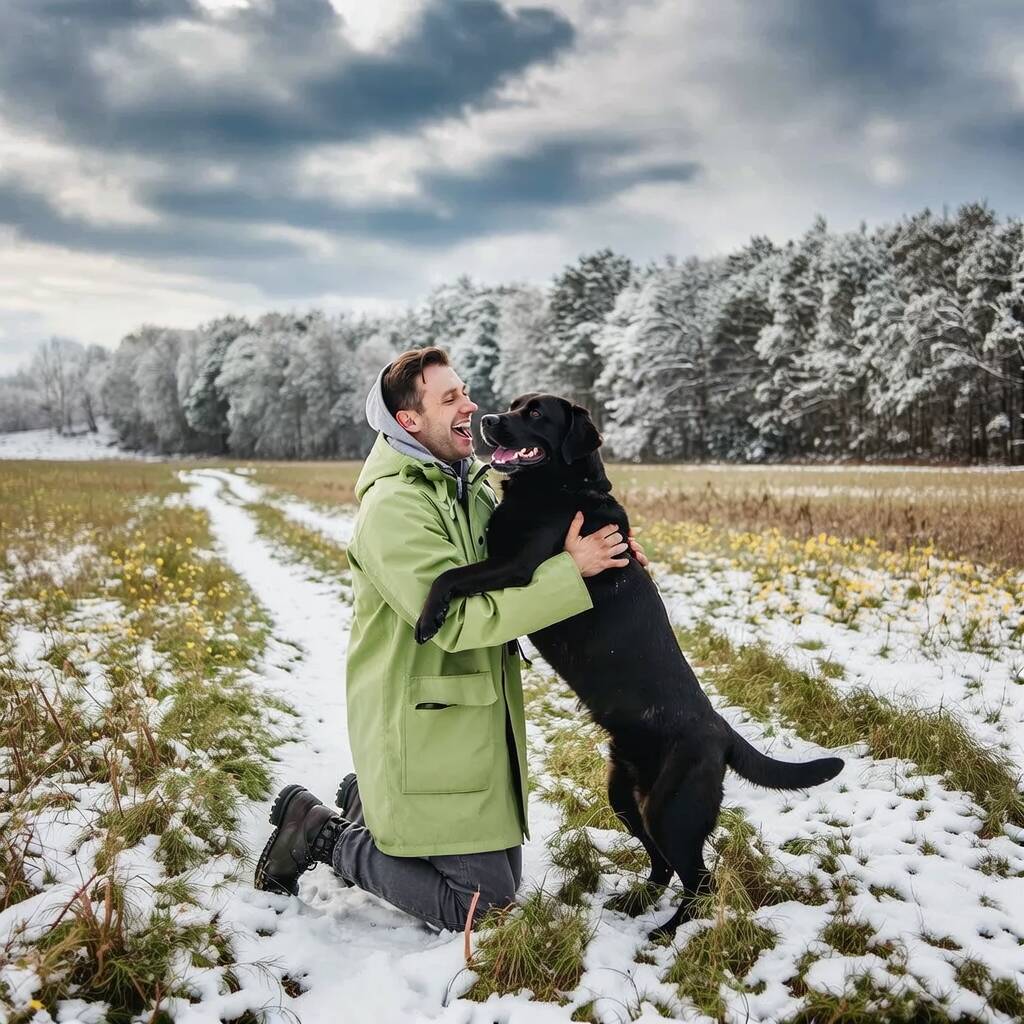}
  \\[-2pt]

  \parbox{\imgw}{\scriptsize\centering \textbf{}} &
  \parbox{\imgw}{\scriptsize\centering \textbf{black dog}} &
  \parbox{\imgw}{\scriptsize\centering \textbf{light-green coat, hoodie on}} &
  \parbox{\imgw}{\scriptsize\centering \textbf{Add snow}}

\end{tabular}

\vspace{-3pt}
\caption{Additional refinement examples.}
\label{fig:app_disentanglement}
\end{figure*}